\documentclass{article}

\usepackage{microtype}
\usepackage{graphicx}
\usepackage{booktabs} 

\usepackage{mathtools} 
\usepackage[utf8]{inputenc} 
\usepackage[T1]{fontenc}    
\usepackage{amsfonts}       
\usepackage{nicefrac}       
\usepackage{microtype}      
\usepackage{xcolor}         
\usepackage{booktabs}
\usepackage{titletoc}
\usepackage{enumitem}
\usepackage{comment}
\usepackage{amssymb}  
\newcommand{\cmark}{\checkmark}

\usepackage{url}
\usepackage{amssymb}
\usepackage{amsthm}
\usepackage{dsfont}
\usepackage{subcaption}
\usepackage{graphicx}

\usepackage{amsmath, bm}  
\usepackage{mathtools}            
\usepackage{multirow}
\usepackage{array}
\usepackage{accents}
\usepackage{makecell}
\usepackage{dsfont}
\usepackage{braket}
\usepackage{multirow}
\usepackage{ulem}
\usepackage{bbm} 
\usepackage{float}
\usepackage{placeins}

\usepackage{tikz}
\usetikzlibrary{shapes.geometric,arrows.meta,positioning, backgrounds, fit, calc, automata}

\newtheorem{theorem}{Theorem}[section]

\newtheorem{definition}[theorem]{Definition}

\newtheorem{remark}[theorem]{Remark}
\usepackage{hyperref}

\usepackage[accepted]{icml2025}

\usepackage{amsmath}
\usepackage{amssymb}
\usepackage{mathtools}
\usepackage{amsthm}
\usepackage[capitalize,noabbrev]{cleveref}

\usepackage[textsize=tiny]{todonotes}

\renewcommand{\algorithmicrequire}{\textbf{Input:}}
\renewcommand{\algorithmicensure}{\textbf{Output:}}

\DeclareMathOperator*{\argmin}{arg\,min}

\newcommand{\icmlJointSupervision}{\textsuperscript{$\dagger$}Joint supervision}
\makeatletter
\renewcommand{\@pa}[1]{%
\ifcsname the@affil#1\endcsname\else
  \ifcsname @icmlsymbol#1\endcsname\else
    \stepcounter{@affiliationcounter}%
    \newcounter{@affil#1}%
    \setcounter{@affil#1}{\value{@affiliationcounter}}%
  \fi
\fi
\ifcsname @icmlsymbol#1\endcsname
  \textsuperscript{\csname @icmlsymbol#1\endcsname\hspace{0.08em}}%
\else
  \textsuperscript{\arabic{@affil#1}\hspace{0.08em}}%
\fi
}
\makeatother

\icmltitlerunning{Stabilizing Transformer Training Through Consensus}

\begin{document}

\twocolumn[
\icmltitle{Stabilizing Transformer Training Through Consensus}



\icmlsetsymbol{equal}{*}
\icmlsetsymbol{supervised}{\ensuremath{\dagger}}

\begin{icmlauthorlist}
\icmlauthor{Shyam Venkatasubramanian}{equal,anthrogen}
\icmlauthor{Sean Moushegian}{equal,anthrogen}
\icmlauthor{Michael Lin}{equal,anthrogen}
\icmlauthor{Mir Park}{anthrogen}
\icmlauthor{Ankit Singhal}{supervised,anthrogen}
\icmlauthor{Connor Lee}{supervised,anthrogen}
\end{icmlauthorlist}

\icmlaffiliation{anthrogen}{Anthrogen PBC, San Francisco, CA, USA}

\icmlcorrespondingauthor{S. Venkatasubramanian}{shyam@anthrogen.com}
\icmlcorrespondingauthor{A. Singhal}{ankit@anthrogen.com}
\icmlcorrespondingauthor{C. Lee}{connor@anthrogen.com}

\icmlkeywords{Machine Learning, ICML}

\vskip 0.3in
]



\printAffiliationsAndNotice{\icmlEqualContribution\icmlJointSupervision} 

\begin{abstract}
Standard attention-based transformers are known to exhibit instability under learning rate overspecification during training, particularly at high learning rates. While various methods have been proposed to improve resilience to such overspecification by modifying the optimization procedure, fundamental \textit{architectural} innovations to this end remain underexplored. In this work, we illustrate that the \textit{consensus} mechanism, a drop-in replacement for attention, stabilizes transformer training across a wider effective range of learning rates. We formulate consensus as a graphical model and provide extensive empirical analysis demonstrating improved stability across learning rate sweeps on text, DNA, and protein modalities. We further propose a hybrid consensus-attention framework that preserves performance while improving stability. We provide theoretical analysis characterizing the properties of consensus.
\end{abstract}

\section{Introduction} \label{sec:introduction}

The literature on machine learning and language modeling is rich with results linking model performance to scale \cite{chinchilla}, reinforcing an impetus to increase model size. While the visible challenges of scaling are the \textit{heavy} constraints of securing compute and data, equally important are the subtler and more \textit{delicate} challenges of balancing interdependent hyperparameters within a narrow region of optimality across many orders of magnitude in model size.

In this work, we consider the latter \textit{delicate} challenge via the lens of learning rate robustness in transformer training. For large pretraining runs, sensitivity to high learning rates is a major practical concern: training instability can manifest as explicit numerical overflows or loss curves that subtly plateau above the attainable optimum, which degrades model performance \cite{porian2024resolving}. Fixing such runs can require expensive restarts from earlier checkpoints at reduced learning rates~\citep{kalra2024warmup}. To mitigate failures, practitioners often tune learning rates on smaller models and rely on heuristic methods \cite{goyal2018accuratelargeminibatchsgd} to extrapolate suitable hyperparameters to larger models. Alternatively, practitioners perform limited training runs on larger models \cite{yang2021tuning} to ablate the learning rate, though this quickly becomes expensive at scale.

Attention-based transformers have been shown to exhibit training instability at high learning rates \cite{XiongPreVsPostLN, zhang2022optopenpretrainedtransformer, kalra2024warmup}. Recent empirical studies further suggest that a key cause of this instability is concentrated in the attention mechanism itself: the spectra $\sigma(W_q^\top W_k)$ of the key and query matrices become ill-conditioned as the learning rate is increased~\citep{QiTransformerTrainingWoWU}. As the transformer backbone and attention mechanism are increasingly used beyond text---for DNA, proteins, and other modalities---developing alternative mechanisms more robust to high learning rates is of practical interest.

Much of the work on mitigating instability at high learning rates has focused on optimizers. For example, it has been shown stochastic gradient descent (SGD) performs worse for transformers than the widely used AdamW~\citep{Kingma2014AdamAM, Loshchilov2017DecoupledWD}, with evidence this gap is related to block-wise heterogeneity in the Hessian~\citep{ZhangAdamBlockWiseHessian} as well as the $\ell^\infty$ geometry of the loss~\citep{XieAdamLInfty}. Learning rate warmup using a narrow window of maximum learning rates has been shown necessary for stable large-scale training~\cite{kalra2024warmup,wortsman2024smallscale}. Regularization of the parameter norm (e.g., $\ell_2$-regularization) has been shown to stabilize training by bounding the parameters in a well-behaved neighborhood of the loss landscape~\cite{van2017l2}. Other methods, such as momentum and coordinate-wise effective step size scaling~\cite{Loshchilov2017DecoupledWD}, refine optimizers and improve convergence across learning rates.

In contrast, we shift our focus from the optimization procedure to architecture. Unbounded \textit{magnitude} of hidden embeddings can destabilize training, and adding LayerNorm to architectures can mitigate this issue \cite{kim2025perilnrevisitingnormalizationlayer}. In a similar vein, we posit that training can be destabilized by a layer exhibiting unbounded \textit{frequency}, where highly-related sequence positions exhibit wildly different representations. We hypothesize that a simple architectural regularizer, like LayerNorm, can stabilize training.

To this end, we rigorously study the \textit{consensus} mechanism, a drop-in replacement for attention proposed in \citet{odyssey1}. The inputs to consensus are a sequence of embeddings pertaining to a \textit{consensus graph}, whose edges connect related sequence positions.  Consensus interchanges information between these positions via the consensus graph, and finally outputs transformed embeddings.  In this paper, we show that this interchange is equivalent to a low-pass filter on the frequency of embeddings. Like attention, consensus admits \textit{self-consensus}, which relates an input sequence to itself, and \textit{cross-consensus}, which incorporates auxiliary context; both can be extended to multi-head formulations. Through empirical results across text, DNA, and proteins, we show that consensus tolerates a wider range of learning rates than attention, with the difference most pronounced in the high learning rate regime. Finally, we present a hybrid consensus-attention architecture demonstrating how consensus can be integrated into transformers to improve stability while preserving attention-level performance.

This paper is organized as follows. In Section~\ref{sec:background}, we provide relevant background material. In Section~\ref{sec:consensus}, we outline the consensus mechanism. In Section~\ref{sec:analysis}, we analyze relevant theoretical properties of consensus. In Section~\ref{sec:empirical_results}, we present empirical results benchmarking hybrid consensus-attention and consensus-based transformers versus attention-based transformers on text, DNA, and proteins. We conclude the paper in Section~\ref{sec:conclusion}. In the Appendix, we provide proofs and derivations, additional empirical results, architecture and training details, and sample generations.

\subsection{Graph-Based Models}
Numerous datasets comprise multiple features of distinct types, which are statistically related. Some features (e.g., adjacent pixels in an image) are strongly related, while others (e.g., opposite-corner pixels) are related weakly. Representing features as nodes and dependencies as weighted edges, one can construct a graph that captures the data's joint dependency structure. These inter-feature dependencies may arise from causality, physical proximity, or correlation under the data-generating distribution. Prior works have incorporated graphs into generative, regression, and encoding models, as well as regularization methods \cite{bp_reg}.

Bayesian networks \cite{bn_tutorial} and Markov random fields \cite{mrf_review, mrf_review_2, 910572} are widely studied probabilistic generative models \cite{maasch2025probabilisticgraphicalmodelsconcise}. Given a known and fixed feature-dependency graph, these methods learn a joint distribution over the features. In some data types, dependencies are naturally hierarchical---one feature influences others, which in turn influence additional features---in which case a directed acyclic graph is an appropriate representation. In other settings, pairwise interactions are symmetric, and undirected graphs provide a better model. Accordingly, Bayesian networks and Markov random fields parameterize distributions on directed and undirected dependency graphs, respectively, enabling sampling and inference.

Beyond generative modeling, graph-based models also have applications in nonparametric regression.  When a learned or known graph with weighted adjacency matrix $W \in \mathbb{R}^{n \times n}$ and diagonal degree matrix $D\in \mathbb{R}^{n\times n}$ marks the similarity of the data, the nonparametric regression method of Laplacian smoothing can be used for denoising. Let $X \in \mathbb{R}^{n \times d}$ be a noised version of $Y \in \mathbb{R}^{n \times d}$, where $X_i = Y_i + \epsilon_i$ with $\epsilon_i \sim \mathcal{N}(0_d, \sigma^2 I_d)$.
Observing only $X$, Laplacian smoothing \cite{shimbo2009properties, lap_smoothing} approximates the unobserved clean $Y$ by $\hat{Y}$. For sufficiently small $\rho$:
\begin{align} \label{eq:lap_smoothing}
    \hat{Y} &= \argmin_{Y} \big(\|X-Y\|^2 + \rho Y^{\top} L Y \big) \notag \\
    &= (I + \rho L)^{-1} X \approx (I -\rho L) X \notag \\
    &= (I-\rho  D)X + \rho WX,
\end{align}
where $L = D - W$ is the graph Laplacian. In this formulation, Laplacian smoothing is a \textit{consensus algorithm}, as it continuously propagates information between neighboring nodes in a graphical structure \cite{degroot1974reaching, olfati2007consensus, 1638541, xiao2004fast}.  Researchers have shown that existing mechanisms asymptotically behave as a consensus algorithm \cite{abella2025consensus}.

Graphical methods have additionally been incorporated into encoding models. For data with a known dependency structure, graph convolutional networks \cite{graph_conv_nets} extract latent representations. Given a weighted graph with adjacency matrix $W \in \mathbb{R}^{n \times n}$ and diagonal degree matrix $D \in \mathbb{R}^{n \times n}$, a graph convolutional network calculates:
\begin{equation}
    X' = \sigma\big( D^{-(1/2)} (W + I) D^{-(1/2)} X \Theta \big),
\end{equation}
where $X \in \mathbb{R}^{n \times d}$ is an observation of $\mathcal{D}$, $\Theta \in \mathbb{R}^{d \times d'}$ is a learned matrix, $X' \in \mathbb{R}^{n \times d'}$ is a latent representation of $X$, and $\sigma(\cdot)$ is an activation function. Separately, graph attention \cite{graph_attention} provides a drop-in replacement for self-attention and utilizes a known graph $\mathcal{G}$.  In its single-headed formulation, for input embedding $X_i \in \mathbb{R}^{d} : 0 \leq i < n$, graph attention calculates:
\begin{align} \label{eq:graph_attn}
    X'_{i} &=  \sigma \Big(  \sum\nolimits_{j \in \mathcal{N}(i)} \alpha_{i,j}  W X_j \Big), \quad \text{where:} \\ \notag
    \alpha_{i,j} &= \frac{\text{exp}\Big( \text{LeakyReLU}\big( \psi^{\top} [WX_i \| WX_j]\big) \Big)}{\sum\limits_{k \in \mathcal{N}(i)} \text{exp}\Big( \text{LeakyReLU}\big( \psi^{\top} [WX_i \| WX_j]\big) \Big)},
\end{align}
where $\psi \in \mathbb{R}^{2d'}$ and $W \in \mathbb{R}^{d' \times d}$ are learnable parameters, $\mathcal{N}(i)$ is the set of neighbors of node $i$, $\sigma(\cdot)$ is the sigmoid function, and $[ \cdot \| \cdot]$ denotes the concatenation operation.

\nocite{proteinmpnn}

\section{Theoretical Background} \label{sec:background}
In this section, we summarize theoretical background necessary for the analysis of the consensus algorithm in Section~\ref{sec:analysis}.

\subsection{Energy and Smoothing}
For a connected, directed, and weighted graph $\mathcal{G} = (V,E)$ of order $N$, we define the adjacency matrix $W \in \mathbb{R}^{N \times N}_{\geq 0}$ to be a matrix of edge weights where $W_{i,j} > 0$ for $(i,j) \in E$ and $W_{i,j}=0$ for $(i,j) \not\in E$.
We define the out-degree matrix as the diagonal matrix:
\begin{equation}
    D_{i,j} = \begin{cases}
        \sum_{j} W_{i,j} & \text{if} \ \ i = j \\
        0 & \text{else},
    \end{cases}
\end{equation}
and define the \textit{graph Laplacian} to be $L = D - W$.  We further let $D_\text{sym} = \tfrac12(D + D^{\top}), W_\text{sym} = \tfrac12(W + W^{\top}), L_\text{sym} = \tfrac12(L + L^{\top})$ and observe $L_\text{sym} = D_\text{sym} -W_\text{sym}$. We note self-loops in $\mathcal{G}$ do not affect the Laplacian $L$, as $W_{i,i}>0$ cancels with its own contribution to $D_{i,i}$ and hence if $\mathcal{G}'$ is the graph constructed by removing all self-loops from $\mathcal{G}$, then the graph Laplacians of $\mathcal{G}$ and $\mathcal{G}'$ are identical.

A \textit{graph signal} is a mapping from the set of vertices $V$ to scalars in $\mathbb{R}$ and can be viewed as a vector in $\mathbb{R}^N$.  Letting $u \in \mathbb{R}^N$ be a signal on graph $\mathcal{G}$, the \textit{energy} of signal $u$ is:
\begin{equation} \label{eq:energy}
    \mathcal{E}(u) = \frac{1}{2} \sum\nolimits_{i,j} W_{i,j} (u_i - u_j)^2 = u^{\top} L u.
\end{equation}
Moreover, the gradient of the energy is given by:
\begin{align}
    \nabla \mathcal{E}(u) &= \left[ \sum\nolimits_j (W_{i,j} + W_{j,i})(u_i - u_j) \right]_i \notag \\
    &= (L + L^{\top})u = 2L_\text{sym}u.
\end{align}
We note self-connections in $\mathcal{G}$ do not contribute to the energy in Eq.~\eqref{eq:energy} as $(u_i-u_j)^2=0$ when $i=j$.  We can transform the signal $u$ via a \textit{gradient update} to arrive at a lower-energy version $u'$, wherein we define $H_\text{sym} = I - 2\eta L_\text{sym}$\nocite{olfati2007consensus}:
\begin{align}
\label{eq:consensus_update}
    u' &=  u - \eta \nabla \mathcal{E}(u) 
    = \underbrace{(I - 2\eta L_\text{sym})}_{H_\text{sym}}u.
\end{align}
If $\mathcal{G}$ is undirected (whereby $L=L^{\top}$), we observe that the consensus update from Eq.~\eqref{eq:consensus_update} is identical to the Laplacian smoothing update from Eq.~\eqref{eq:lap_smoothing} with $\rho = 2\eta$.

\subsection{Low-Pass Filtering}
By the quadratic form in Eq.~\eqref{eq:energy}, the graph Laplacian $L_\text{sym}$ is symmetric positive semidefinite. Consequently, its eigenvalues are real and nonnegative, and its eigenvectors form an orthonormal basis of $\mathbb{R}^N$. Let $0=\lambda_0 < \lambda_1 \leq \cdots \leq \lambda_{N-1}$ denote the eigenvalues of $L_\text{sym}$, with associated eigenvectors $\{v_i\}_{i=0}^{N-1}$. A standard computation shows that $v_0=\mathbf{1}/\sqrt{N}$ and $\lambda_0=0$. Any signal $u \in \mathbb{R}^N$ can be decomposed\footnote{We note this can be viewed as a generalization of the Fourier transform to the graph domain. Let $I=[a,b] \subset \mathbb{R}$ with $a<b$. On $I$, the Laplacian denotes the second-derivative operator $\partial^2/\partial x^2$, with eigenfunctions given by sinusoids and eigenvalues that increase with frequency (where $\lambda_0=0$ corresponds to the constant eigenfunction $v_0(x)=\tfrac{1}{\sqrt{b - a}}$). Using the Fourier transform, any sufficiently regular function (subject to boundary conditions) can be expressed as a superposition of these modes. If many points are sampled from $I$ and edges are formed to construct a graph (with edge weights related to Euclidean proximity), then the resulting graph Laplacian can be shown \cite{belkin2008towards} to converge to the second-derivative operator (or, in higher dimensions, the trace of the Hessian) in the large-sample limit.} into the eigenbasis $\{v_i\}_{i=0}^{N-1}$ with coefficients $\alpha \in \mathbb{R}^N$:
\begin{equation} \label{eq:decompose}
    u = \sum\nolimits_{i=0}^{N-1} \alpha_i  v_i.
\end{equation}
The expression of Eq.~\eqref{eq:decompose} enables a filtering interpretation of $L_\text{sym}$ \cite{graph_conv_nets}. For the $u$ of Eq.~\eqref{eq:decompose}, we can express the filtered version ($u'$ of Eq.~\eqref{eq:consensus_update}) as:
\begin{equation}
    Hu = \sum\nolimits_{i=0}^{N-1} \alpha_i (1 - 2 \eta \lambda_i) v_i.
\end{equation}
We call this filtering \textit{non-oscillating} when $1-2\eta\lambda_{N-1} \geq 0$ (equivalently, $\eta \leq 1/(2\lambda_{N-1})$). For both graphs and Euclidean intervals, Laplacian eigenvalues $\lambda_i$ increase with the frequency of the corresponding eigenvectors: they are larger for high-frequency modes, smaller for low-frequency modes, and equal to zero for the constant eigenvector $v_0$. Consequently, high-frequency components are strongly attenuated, while low-frequency components are affected minimally. For this reason, we refer to $H_\text{sym}$ as a \textit{low-pass filter}.

\begin{figure*}[t!]
    \centering
    \includegraphics[width=0.9\linewidth]{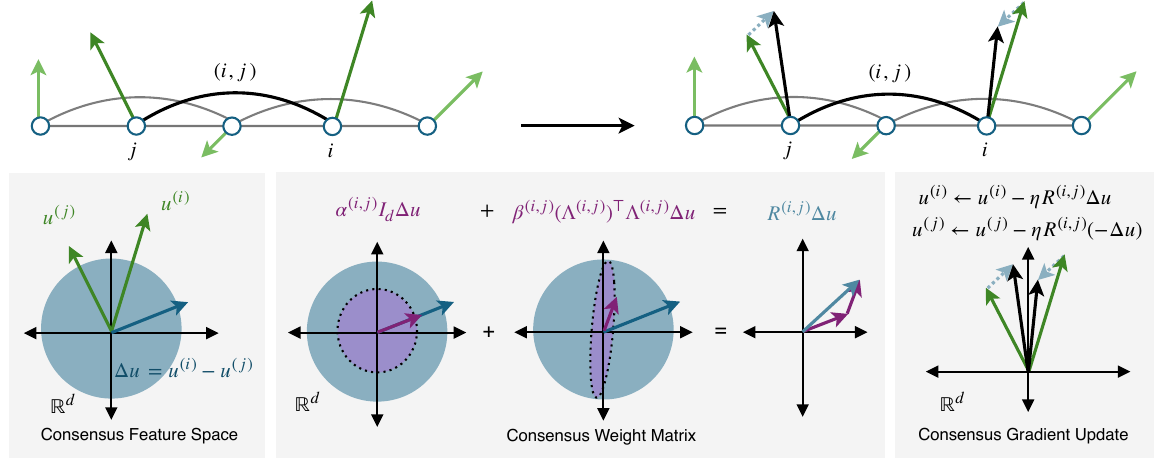}
    \caption{The consensus update step visualized between nodes $i$ and $j$. The difference (blue) between consensus features $u^{(i)}$ and $u^{(j)}$ (green) are transformed by $R^{(i,j)}$, consisting of a scalar $\alpha^{(i,j)}$ and a shear matrix $\beta^{(i,j)}(\Lambda^{(i,j)})^{\top}\Lambda^{(i,j)}$ component (the respective actions illustrated by purple ellipses).  The transformed difference scaled by $\eta$, (dotted blue) then updates $u^{(i)}, u^{(j)}$ (black).}
    \label{fig:placeholder}
\end{figure*}

\subsection{Effects of Filtering} \label{subsec:effects_filtering}
Let $\omega_i = 1-2\eta\lambda_i$. For an arbitrary eigenvector $v_i$ of $L_\text{sym}$:
\begin{align}
    H_\text{sym} v_i &= (I - 2\eta L_{\text{sym}})v_i  = v_i - 2\eta \lambda_i v_i \notag \\
    &= (1-2\eta\lambda_i) v_i = \omega_i v_i.
\end{align}
Because $H_\text{sym}v_i = \omega_i v_i$, $v_i$ is an eigenvector of $H_\text{sym}$ with accompanying eigenvalue $\omega_i$. Since the smallest eigenvalue of $L_\text{sym}$ is $\lambda_0 = 0$, the largest eigenvalue of $H_\text{sym}$ is $\omega_0 = 1$. Clearly, $\lim_{n \rightarrow \infty} \omega_0^n = 1$. Under the non-oscillation and connectedness conditions, $0 \leq \omega_{N-1} \leq ...\leq \omega_1 < 1$. It follows that $\lim_{n \rightarrow \infty} \omega_i^n = 0$ for all $1 \leq i < N$. Thus:
\begin{align}
    H_\text{sym}^n &= (Q\Omega Q^{-1})^n = Q \Omega^n Q^{-1} \notag \\
    &= Q \begin{bmatrix}
        \omega_0^n & \cdots & 0 \\
        \vdots & \ddots & \vdots \\
        0 & \cdots & \omega_{N-1}^n
    \end{bmatrix} Q^{-1},
\end{align}
where $Q$ is the orthogonal matrix whose $i$-th column is $v_i$. Let $\Gamma = \text{diag}(1,0, \dots,0)$. Then, for all signals $u$:
\begin{align} \label{eq:convergence_lpf}
    \lim_{n \rightarrow \infty} H_\text{sym}^n u &= Q \Gamma Q^{-1} u = \text{proj}_{v_0}(u) = v_0 v_0^{\top} u \notag \\
    &= (1/N) \, \mathbf{1}^{\top} \mathbf{1} u.
\end{align}
Thus, successive filtering updates $u$ towards its mean value. Let $A_n = H^n_\text{sym} - v_0v_0^{\top} = Q(\Omega^n-\Gamma)Q^{\top}$. Then:
\begin{align}
    \left\|H^n_\text{sym} u - \lim_{n\to\infty} H^n_\text{sym} u \right\|^2 =  \langle u, A^2_n u \rangle \leq \omega_1^{2n} \|u\|^2,
\end{align}
so the convergence rate in Eq.~\eqref{eq:convergence_lpf} is governed by the magnitude of the second-largest eigenvalue of $H_\text{sym}$, $\omega_1$. Hence $\lambda_1$ is proportional to the rate of information propagation under Laplacian smoothing\footnote{The second-smallest Laplacian eigenvalue $\lambda_1$ is the \textit{Fiedler value} (algebraic connectivity), measuring graph connectedness; variationally, $\lambda_1=\min_{u\perp \mathbf{1}}\langle u,Lu\rangle \,/ \, {\langle u,u\rangle}$.}. Since Eq.~\eqref{eq:consensus_update} uses a symmetrized update scheme, the relevant quantity is the Fiedler value of $L_\text{sym}$; the Fiedler value of the non-symmetric $L$ does not, in general, control the rate of information propagation.

In the regime of abundant data, scalars may be insufficient to express a particular feature. In this setting, we generalize our previous results to permit vector-valued graph signals \cite{hansen2021expansion}, mappings from the set of nodes $V$ to vectors in $\mathbb{R}^d$ which we represent as vectors in $\mathbb{R}^{Nd}$.
We replace positive edge weights $W_{i,j} > 0$ with positive definite matrices $W_{i,j} \succ 0$ where $W_{i,j} \in \mathbb{R}^{d \times d}$. From these $N^2$ $\mathbb{R}^{n \times n}$ matrices, we construct an adjacency matrix $W \in \mathbb{R}^{Nd \times Nd}$ and a block-diagonal degree matrix $D \in \mathbb{R}^{Nd \times Nd}$. We describe this procedure in more detail and discuss the resulting symmetrized Laplacian and its spectrum in Appendix~\ref{appendix:theory}.

\section{Consensus Mechanism} \label{sec:consensus}
In this section, we formalize the \textit{consensus} mechanism of \citet{odyssey1}. We provide theoretical background on graph energy and filtering, before introducing the algorithm and graph-theoretic framework underlying consensus.

The consensus mechanism is a drop-in replacement for attention in transformers that uses the update of Eq.~\eqref{eq:consensus_update}. For directed graph $(V,E)$ with $|V| = N$, consensus takes as input a vector of source embeddings $y = [y^{(0)},\dots,y^{(N-1)}]^\top \in \mathbb{R}^{N\times d}$ corresponding to nodes $V$. For each node $i \in V$, the source embedding $y^{(i)}$ is mapped to $u^{(i)}$, and for each edge $(i,j) \in E$, the pair of source embeddings $[y^{(i)}; y^{(j)}]$  is mapped to a $d$-dimensional, positive-definite, \textit{consensus weight matrix}, $R^{(i,j)}$. We perform a gradient update step to decrease the energy $\mathcal{E}(u)$, where $\mathcal{E}(u)$ is given by:
\begin{align}
  \mathcal{E}(u) = \frac{1}{2} \sum_{\{i,j\}\in E} (u^{(i)} - u^{(j)})^\top R^{(i,j)} (u^{(i)} - u^{(j)}),
\end{align}
where $R^{(i,j)}$ is the weight of edge $(i,j) \in E$.  Finally, an output projection is performed. The complete description of self-consensus is given by Algorithm~\ref{alg:self-consensus}, where $\sigma^+(\cdot)$ is the softplus activation, and RS and RN denote reshape and row-norm operations (defined in Appendix~\ref{appendix:consensus}), respectively.

Though the mechanism of Algorithms~\ref{alg:self-consensus} and \ref{alg:self-weight-network} is defined for any graph, in the case that $V$ corresponds to the positions of a 1-dimensional sequence, we often take the underlying graph to be of a form similar to that of a sliding window:
\begin{definition}[Window-Path Graph] \label{def:window_path_graph}
For window size $w$ and length $N$, the window-path graph $P^w_N$ is given by:
\begin{align}
    &P^w_N = (V_N, E_N^w), \ \ \text{where} \ \ V_N = \{0, \cdots, N-1 \} \ \ \text{and} \notag \\
    &E_N^w = \{(i,j) : 0 \leq i,j < N \land 0< |i-j| \leq w\}. \notag
\end{align}
\end{definition}
In multimodal applications, auxiliary modalities are often imbued into a primary source track~\cite{esm3}. To support this, \citet{odyssey1} proposed cross-consensus and multi-head extensions of both self and cross-consensus. We analyze these methods in Appendix~\ref{appendix:consensus}.

\section{Theoretical Analysis of Consensus} \label{sec:analysis}

For sequential data, the window-path graph of Definition~\ref{def:window_path_graph} with window size $w$ is often a useful choice. We examine the properties of the hyperparameter $w$, and runtime complexity.

We recall from Section~\ref{subsec:effects_filtering} that the second-smallest eigenvalue of $L_\text{sym}$, denoted $\lambda_1(L_\text{sym})$, quantifies the ease with which information propagates across the graph and is known as the \textit{algebraic connectivity}. Illustratively, letting $K_N$ be the complete graph on $N$ nodes, we obtain $\lambda_1(K_N)=N$, and, letting $Z_N$ be the edgeless graph on $N$ nodes, we have $\lambda_1(Z_N)=0$ (indeed, $\lambda_1=0$ for any disconnected graph).

\makeatletter
\newcommand{\linelabel}[1]{%
  \begingroup
  \edef\@currentlabel{\theALC@line}
  \phantomsection
  \label{#1}%
  \endgroup
}
\makeatother

\begin{algorithm}[t!]
\caption{Self-Consensus Mechanism} \label{alg:self-consensus}
\begin{algorithmic}[1]
\REQUIRE Embeddings $(y^{(i)})_{i=0}^{N-1}$, directed graph $\mathcal{G}=(V,E)$ 

\STATE \textbf{Hyperparameters:} step size $\eta>0$, embedding dimension $d$, rank $r$, edge hidden dimension $\xi$
\STATE \textbf{Parameters:}  $W_s \in \mathbb{R}^{d \times d}, b_s \in \mathbb{R}^d,  W_o \in \mathbb{R}^{d \times d}, b_o \in \mathbb{R}^d,$ and weight matrix constructor parameters $\mathcal{W}, \mathcal{B}, \Theta$.

\FOR{$i=0$ to $N-1$}
  \STATE \linelabel{alg:line:scInputProj} $u^{(i)} \leftarrow W_s\, y^{(i)} + b_s$
\ENDFOR

\STATE $R = \text{SCWM}(y, \mathcal{G} |  \mathcal{W}, \mathcal{B}, \Theta, d, r, \xi)$

\FOR{$i=0$ to $N-1$}
  \STATE $g^{(i)} \leftarrow \hspace{-2.5ex} \sum\limits_{j:(i,j)\in E} \hspace{-2.5ex} R^{(i,j)}(u^{(i)}-u^{(j)})
    - \hspace{-2.5ex} \sum\limits_{k:(k,i)\in E} \hspace{-2.5ex} R^{(k,i)}(u^{(k)}-u^{(i)})$
  \STATE $u'^{(i)} \leftarrow u^{(i)} - \eta\, g^{(i)}$

  \STATE $y_{\mathrm{out}}^{(i)} \leftarrow W_o u'^{(i)} + b_o$
\ENDFOR

\ENSURE Embeddings $(y_{\mathrm{out}}^{(i)})_{i=0}^{N-1}$
\end{algorithmic}
\end{algorithm}

We next present a result bounding the algebraic connectivity of powers of a cycle graph, which is the result of connecting nodes on a cycle following a sliding-window pattern:
\begin{remark} \label{remark:circulant_graph}
Consider the simple, connected, unweighted, undirected graph $C_N^w=(V,E)$ on $N$ nodes, with $w < N/2$, wherein each node $i$ is connected to nodes $i\pm 1, i\pm 2, \ldots, i\pm w$ modulo $N$ (e.g., node $0$ is connected to node $N-1$ when $w=1$). The second-smallest eigenvalue $\lambda_1$ of $L$ is:
\begin{align}
    \lambda_1 = 2w - 2\sum_{j=1}^w \cos\bigg(\frac{2 \pi}{N} j \bigg) &= 4 \sum_{j=1}^w \sin^2 \left( \frac{\pi}{N} j \right),
\end{align}
and for fixed $N$, $\lambda_1(L)$ is monotonically increasing in $w$. 
\end{remark}
\begin{proof}
The proof is provided in Appendix~\ref{appendix:proofs_and_derivations}.
\end{proof}

\begin{algorithm}[t!]
\caption{Self-Consensus Weight Matrix (SCWM)} \label{alg:self-weight-network}
\begin{algorithmic}[1]
\REQUIRE Embeddings $(y^{(i)})_{i=0}^{N-1}$, directed graph $\mathcal{G}=(V,E)$

\STATE \textbf{Hyperparameters:} embedding dimension $d$, rank $r$, edge hidden dimension $\xi$
\STATE \textbf{Parameters:} $\mathcal{W} = (W_\alpha, W_\beta, W_\Lambda) \in \mathbb{R}^\xi \times \mathbb{R}^\xi \times \mathbb{R}^{rd\times \xi}, \mathcal{B} = (b_\alpha, b_\beta, b_\Lambda) \in \mathbb{R} \times \mathbb{R} \times \mathbb{R}^{rd}$.
\STATE \textbf{Edge MLPs:} $\phi^\alpha_{\theta_\alpha}, \phi^\beta_{\theta_\beta}, \phi^\Lambda_{\theta_\Lambda}: \mathbb{R}^d \times \mathbb{R}^d \mapsto \mathbb{R}^\xi$ with parameters $\Theta = \big(\theta_\alpha, \theta_\beta, \theta_\Lambda \big)$.

\FORALL{$(i,j)\in E$}
  \STATE $\alpha^{(i,j)} \leftarrow \sigma^+\!\big(W_\alpha \phi^\alpha_{\theta_\alpha}([y^{(i)};y^{(j)}]) + b_\alpha\big)$
  \STATE $\beta^{(i,j)} \leftarrow \sigma^+\!\big(W_\beta \phi^\beta_{\theta_\beta}([y^{(i)};y^{(j)}]) + b_\beta\big)$
  \STATE $\Lambda^{(i,j)} \leftarrow \text{RN}\big(\mathrm{RS}_{r,d}(W_\Lambda \phi^\Lambda_{\theta_\Lambda}([y^{(i)};y^{(j)}]) + b_\Lambda)\big) / \sqrt{r}$
  \STATE $R^{(i,j)} \leftarrow \alpha^{(i,j)} I_d + \beta^{(i,j)} (\Lambda^{(i,j)})^\top \Lambda^{(i,j)}$
\ENDFOR

\ENSURE Edge weights $R = \{R^{(i,j)} : (i,j) \in E\}$
\end{algorithmic}
\end{algorithm}

Next, we use the result of Remark~\ref{remark:circulant_graph} to bound $\lambda_1$ on powers of a path graph, which pertains to a graph formed by connecting nodes for a sliding-window pattern on a sequence:
\begin{remark}
\label{remark:path_graph}
Consider the simple, connected, unweighted, and undirected window-path graph $P_N^w = (V, E)$ of Definition~\ref{def:window_path_graph} with $N$ nodes.  Let $w < N/2$.
Then, the Fiedler value of $P_N^w $ satisfies:
\begin{equation}
    2\sum_{j=1}^w \sin^2\bigg(\frac{ \pi}{2N} j \bigg) \leq \lambda_1 \leq  4\sum_{j=1}^w \sin^2\bigg(\frac{ \pi}{2N} j \bigg).
\end{equation}
As $\sin(x) = \Theta(x)$ for $x < \tfrac{\pi}{2}$, we have that $\lambda_1 = \Theta\left(\frac{w^3}{N^2}\right)$; thus the Fiedler value scales cubically in $w$.
\end{remark}
\begin{proof}
The proof is provided in Appendix~\ref{appendix:proofs_and_derivations}.
\end{proof}

While consensus assigns and uses edge weights for its updates, the Fiedler value of the weighted graph can be related to that of the underlying unweighted graph given global upper and lower spectral bounds of $R^{(i,j)}$. Thus, Remark~\ref{remark:path_graph} suggests increasing consensus window size $w$ increases the rate of information mixture for fixed and constant $R^{(i,j)}$.

Lastly, for a single head, we analyze the runtime complexity of the self-consensus and self-attention mechanisms. Self-attention requires forming all pairwise token interactions, yielding a quadratic cost in $N$. In contrast, self-consensus restricts interactions to a window-path graph of size $w$ and forms matrices of rank $r$ (edge hidden dimension $\xi$), yielding a linear cost in $N$ for fixed $w,r,\xi$ (see Table~\ref{tab:runtime_complexity}).

\begin{table}[h!]
\vspace{-0.75em}
\centering
\caption{Asymptotic runtime complexity for a single-head layer.}
\label{tab:runtime_complexity}
\small
\setlength{\tabcolsep}{6pt}
\renewcommand{\arraystretch}{1.1}
\begin{tabular}{lcc}
\toprule
& \textbf{Self-Attention} & \textbf{Self-Consensus}\\
\midrule
\textbf{Cost} & $\mathcal{O}(N^{2}d)$ & $\mathcal{O}\bigl(N d w r (\xi + 1)\bigr)$\\
\bottomrule
\end{tabular}
\end{table}

\begin{figure*}[t!] \centering
\begin{subfigure}{0.24\textwidth}
  \includegraphics[width=\linewidth]{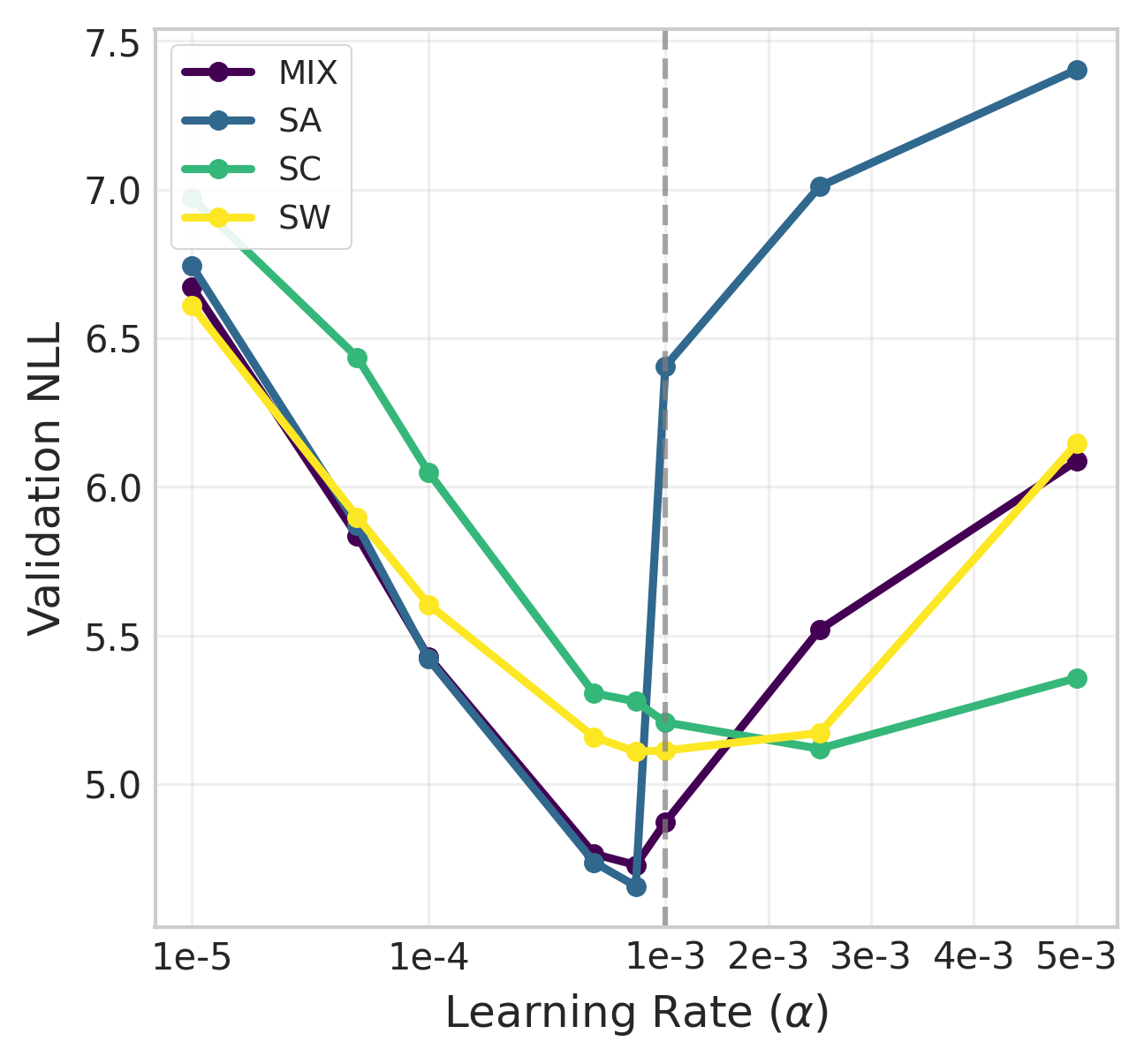}
  \caption{Text 54M val NLL}
\end{subfigure}
\begin{subfigure}{0.24\textwidth}
  \includegraphics[width=\linewidth]{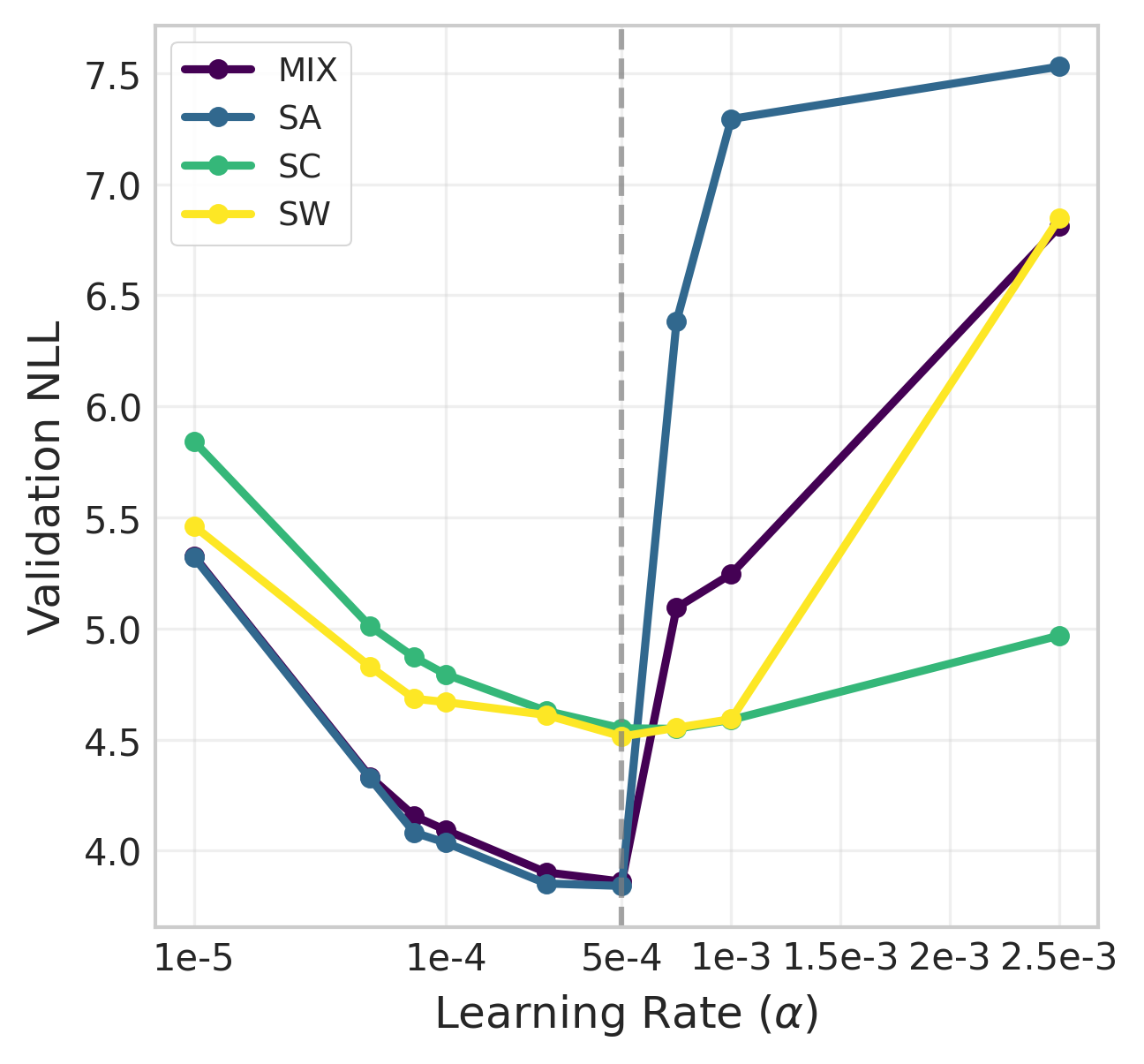}
  \caption{Text 193M val NLL}
\end{subfigure}
\begin{subfigure}{0.24\textwidth}
  \includegraphics[width=\linewidth]{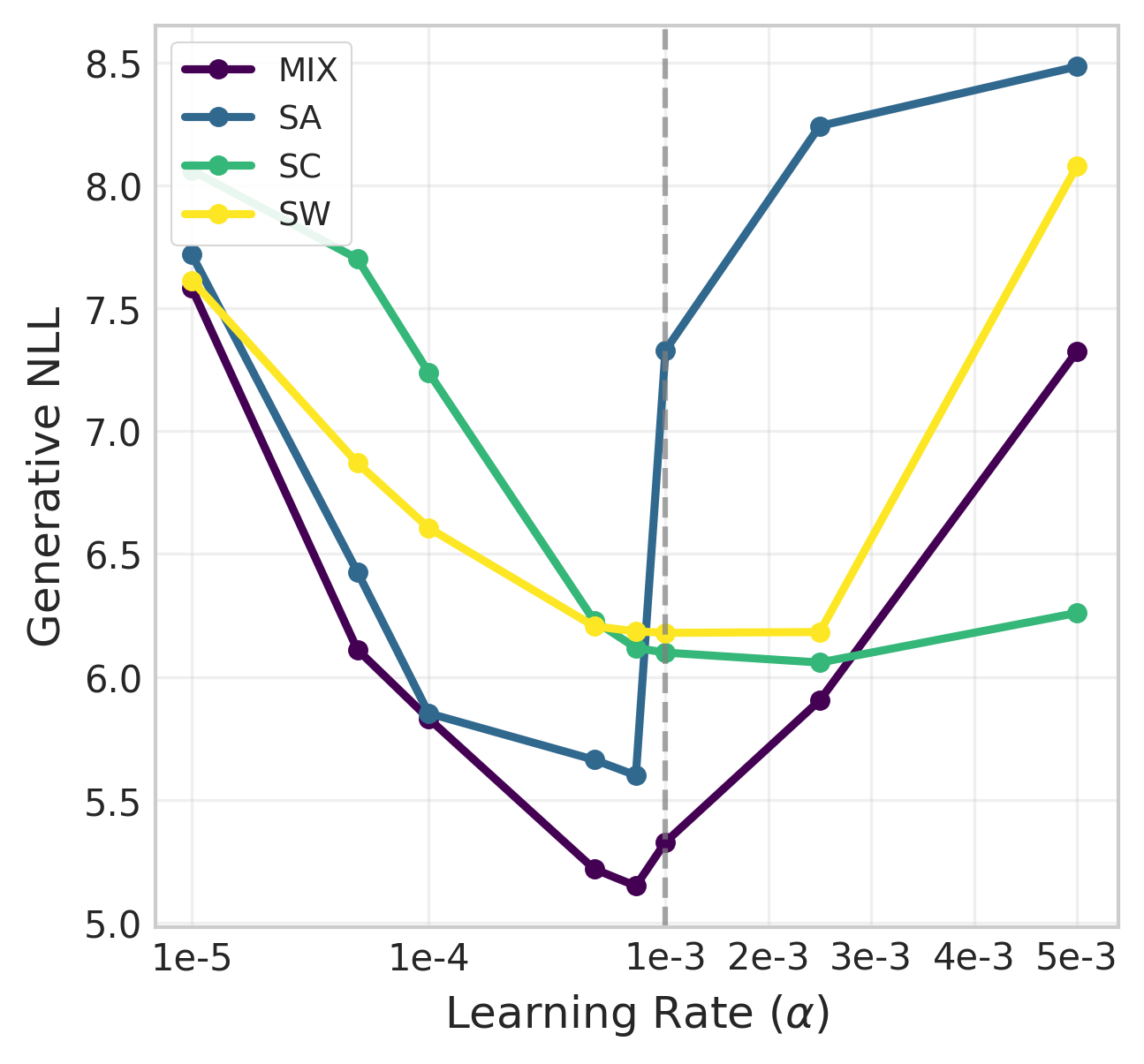}
  \caption{Text 54M gen NLL}
\end{subfigure}
\begin{subfigure}{0.24\textwidth}
  \includegraphics[width=\linewidth]{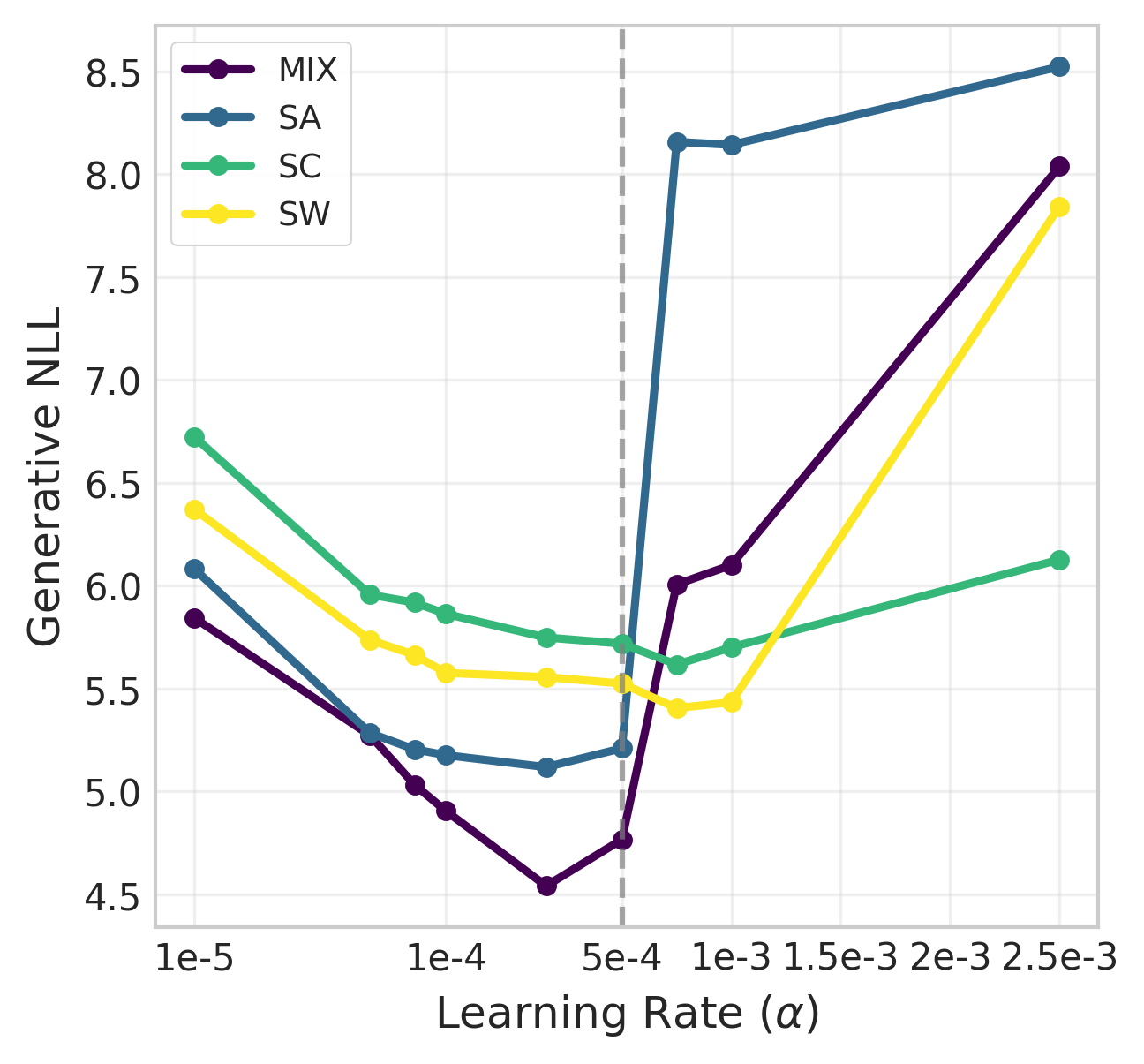}
  \caption{Text 193M gen NLL}
\end{subfigure}
\caption{Text generative NLL and validation NLL on OpenWebText through learning rate sweep. Learning-rate axis shown on log scale for small values and linear scale for a higher-learning rate zoom, where instability is observed.}
\label{fig:text_val_gen_effective_window}
\end{figure*}

\begin{figure*}[t!] \centering
\begin{subfigure}{0.24\textwidth}
  \includegraphics[width=\linewidth]{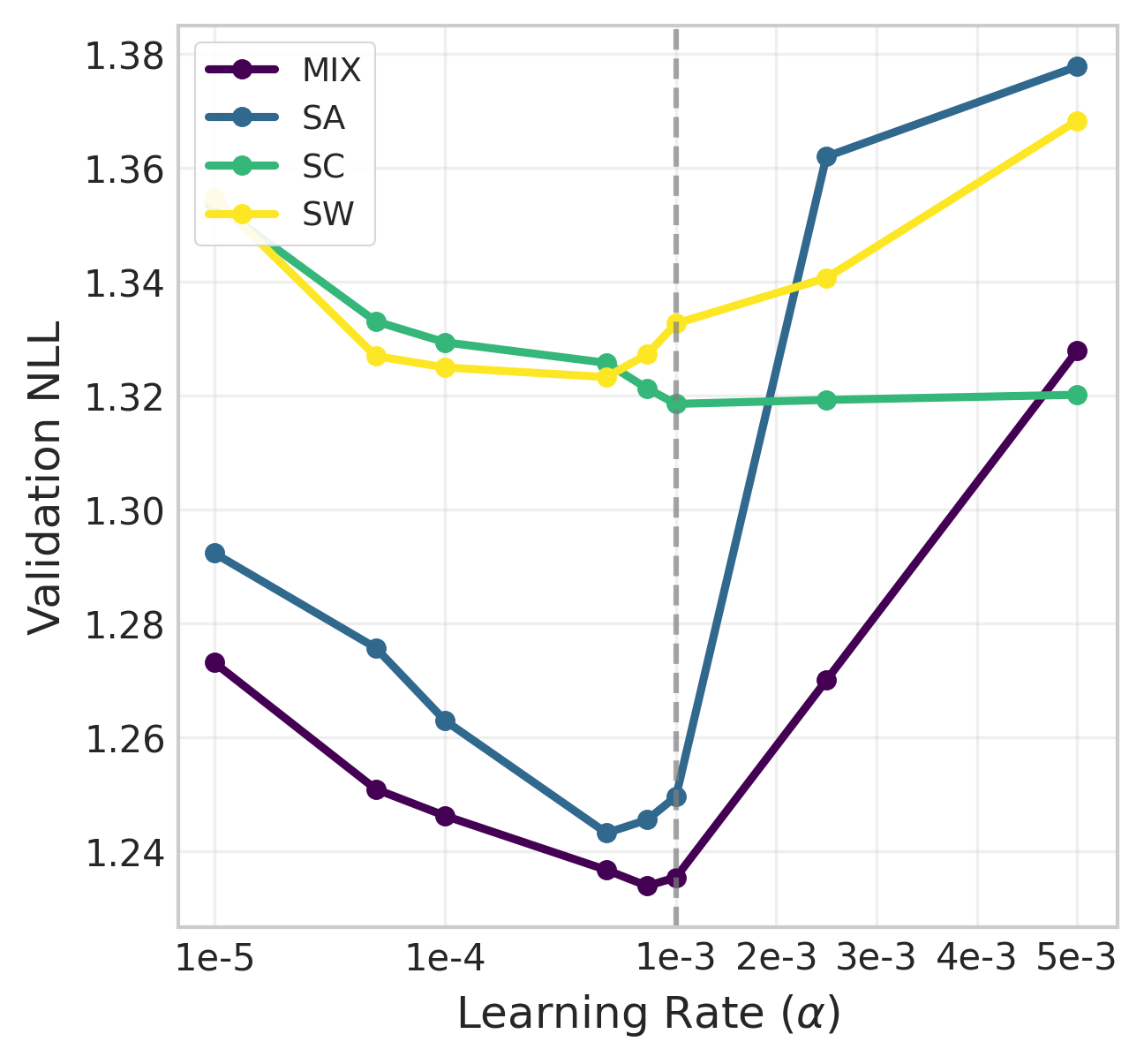}
  \caption{DNA 51M val NLL}
\end{subfigure}
\begin{subfigure}{0.24\textwidth}
  \includegraphics[width=\linewidth]{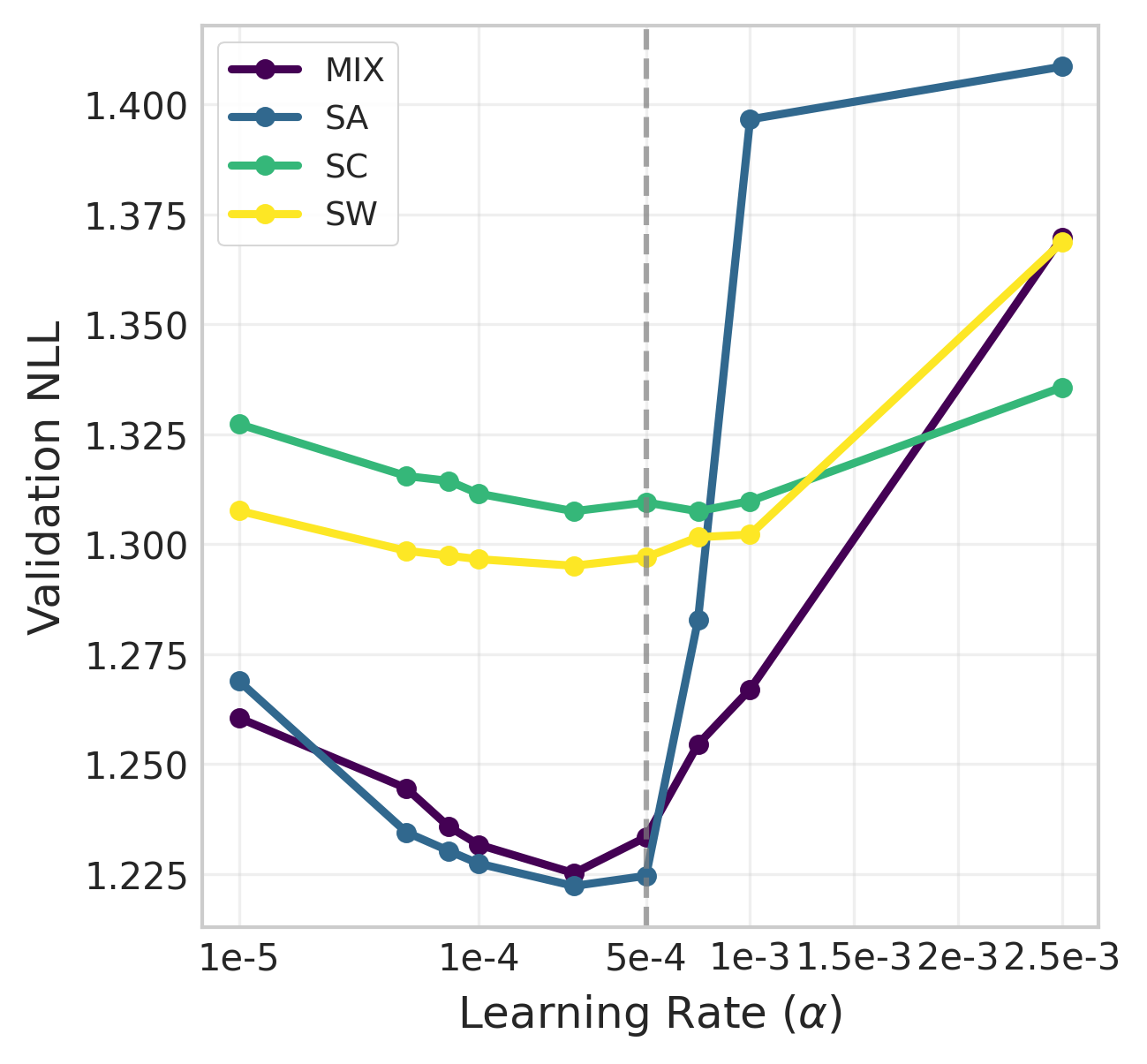}
  \caption{DNA 172M val NLL}
\end{subfigure}
\begin{subfigure}{0.24\textwidth}
  \includegraphics[width=\linewidth]{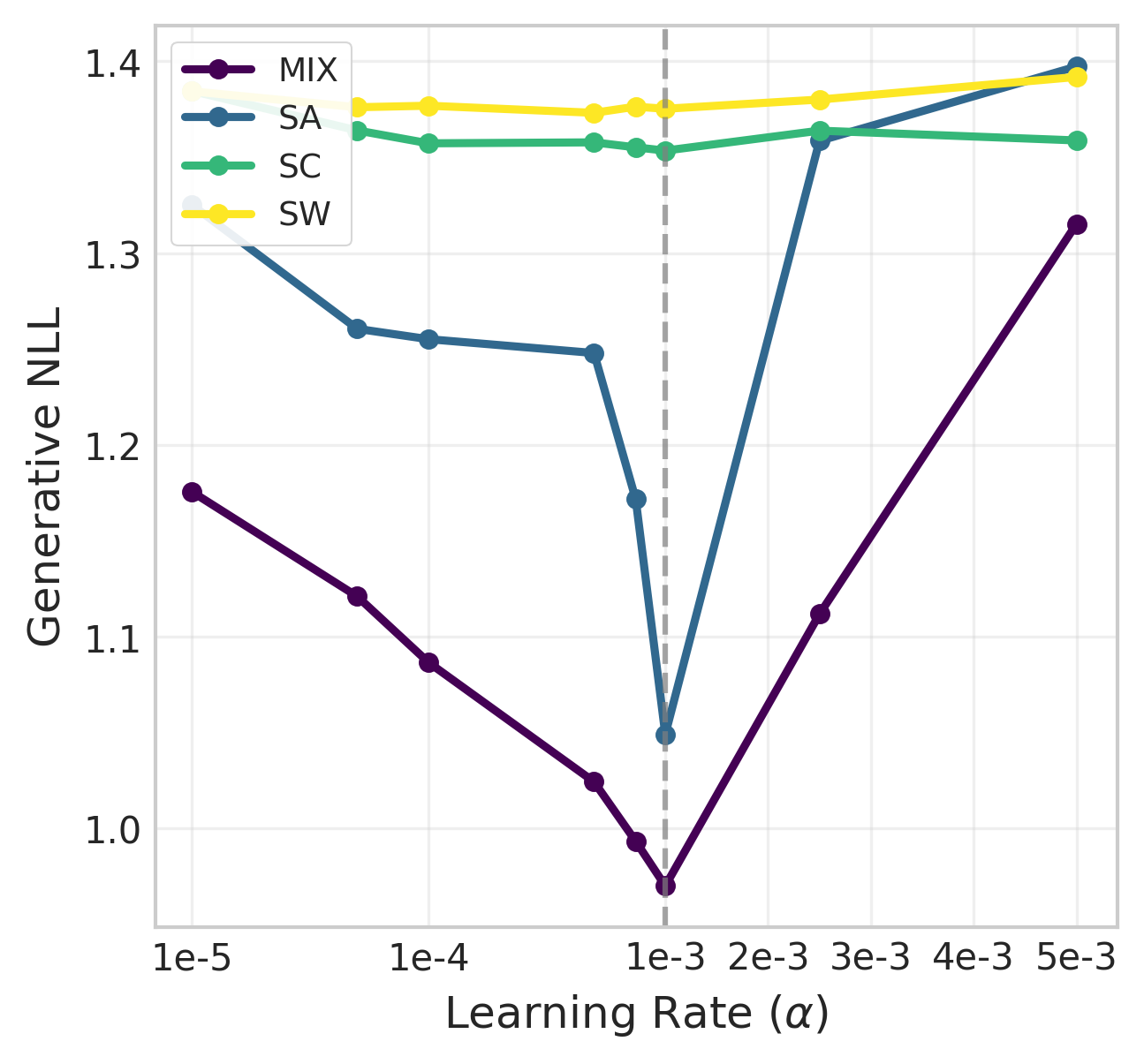}
  \caption{DNA 51M gen NLL}
\end{subfigure}
\begin{subfigure}{0.24\textwidth}
  \includegraphics[width=\linewidth]{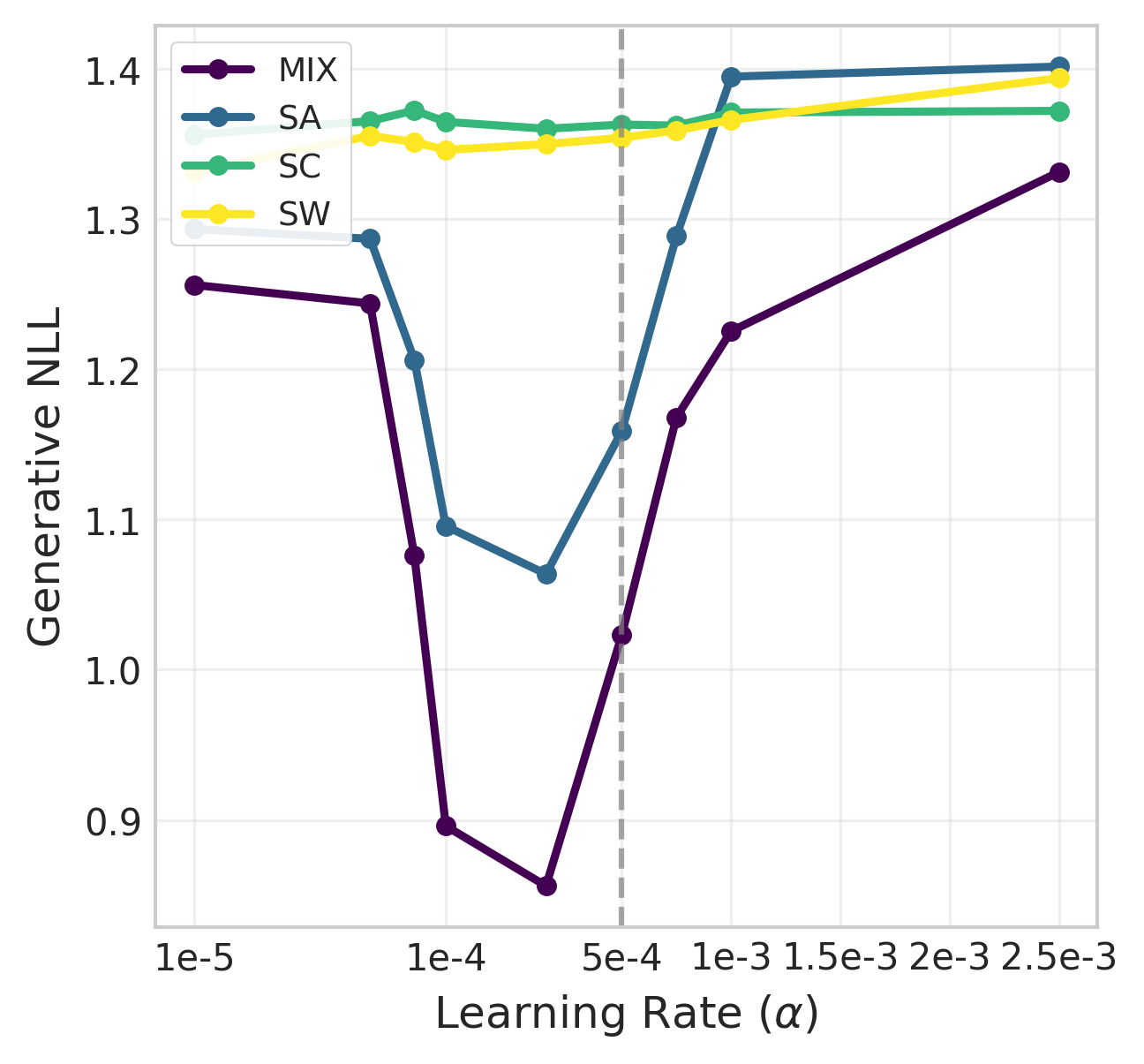}
  \caption{DNA 172M gen NLL}
\end{subfigure}
\caption{DNA generative NLL and validation NLL on OpenGenome through learning rate sweep. Learning-rate axis shown on log scale for small values and linear scale for a higher-learning-rate zoom, where instability is observed.}
\label{fig:dna_val_gen_effective_window}
\vspace{-0.4em}
\end{figure*}

\begin{table*}[t!]
\centering
\caption{Text 385M validation/generative NLLs on OpenWebText. MIX evaluated at select LRs to verify SA-level performance at scale.} \label{tab:text_385M}
\fontsize{8}{9.5}\selectfont
\setlength{\tabcolsep}{4pt}
\renewcommand{\arraystretch}{1.08}
\begin{tabular}{ccccccccccccccccc}
\toprule
\multirow{2}{*}{\textbf{Mechanism}}
& \multicolumn{8}{c}{\textbf{Validation NLL}  {\scriptsize (LR sweep)}} & \multicolumn{8}{c}{\textbf{Generative NLL}  {\scriptsize (LR sweep)}} \\
\cmidrule(lr){2-9}\cmidrule(lr){10-17}
& 1e-5 & 5e-5 & 7.5e-5 & 1e-4 & 2.5e-4 & 5e-4 & 7.5e-4 & 1e-3
& 1e-5 & 5e-5 & 7.5e-5 & 1e-4 & 2.5e-4 & 5e-4 & 7.5e-4 & 1e-3 \\
\midrule
MIX & \textemdash & \textemdash & 3.969 & 3.913 & 3.772 & 6.457 & \textemdash & \textemdash
    & \textemdash & \textemdash & 5.109 & 5.105 & 4.835 & 7.062 & \textemdash & \textemdash \\
SA  & 5.098 & 4.105 & 3.986 & 3.863 & 3.770 & 7.176 & 7.255 & 7.201
    & 5.508 & 5.352 & 5.304 & 5.100 & 4.842 & 8.178 & 8.131 & 8.197 \\
SC  & 5.536 & 4.713 & 4.589 & 4.553 & 4.484 & 4.428 & 4.426 & 4.496
    & 6.549 & 5.636 & 5.597 & 5.568 & 5.532 & 5.566 & 5.556 & 5.723 \\
\bottomrule
\end{tabular}
\vspace{-1.25em}
\end{table*}

\begin{table}[t!]
\centering
\caption{DNA 331M validation/generative NLLs on OpenGenome.} \label{tab:dna_385M}
\fontsize{8}{9.5}\selectfont
\setlength{\tabcolsep}{3pt}
\renewcommand{\arraystretch}{1.05}
\begin{tabular}{ccccccccc}
\toprule
\multirow{2}{*}{\textbf{Mech.}}
& \multicolumn{4}{c}{\textbf{Validation NLL} {\scriptsize (LR sweep)}} 
& \multicolumn{4}{c}{\textbf{Generative NLL} {\scriptsize (LR sweep)}} \\
\cmidrule(lr){2-5}\cmidrule(lr){6-9}
& 1e-4 & 2.5e-4 & 5e-4 & 7.5e-4 & 1e-4 & 2.5e-4 & 5e-4 & 7.5e-4 \\
\midrule
MIX & 1.230 & 1.228 & 1.225 & 1.273 & 1.060 & 1.064 & 1.063 & 1.066 \\
SA  & 1.226 & 1.222 & 1.221 & 1.277 & 1.214 & 1.220 & 1.249 & 1.100 \\
\bottomrule
\end{tabular}
\vspace{-1.25em}
\end{table}

\section{Empirical Results} \label{sec:empirical_results}
We evaluate whether integrating consensus into transformers improves robustness to learning rate overspecification during training. Our empirical study spans multiple modalities and model scales---we report results across 54M, 193M, and 385M parameters for text, 51M, 172M, and 331M parameters for DNA, and 35M, 133M, and 320M parameters for proteins (multimodal sequence and structure modeling). Validation NLLs are computed on a held-out $1\%$ split for each dataset, and generative NLLs are computed by scoring model-generated samples under a larger oracle model. Using $8$ NVIDIA A40 GPUs with an effective batch size of $512$, we train text 54M and DNA models for $5$K global steps, text 193M/385M models for $20$K global steps, and protein models for $3$K global steps, and we evaluate terminal checkpoints to compute validation and generative NLLs. To connect observed instability to local curvature, we also estimate a directional maximum stable learning rate from the loss Hessian along optimizer update directions using terminal checkpoints. Complete hyperparameter and architectural details, alongside additional text experiments ablating consensus hyperparameters, are provided in Appendix~\ref{appendix:detail_empirical}.

\begin{figure*}[t!] \centering
\begin{subfigure}{0.24\textwidth}
  \includegraphics[width=\linewidth]{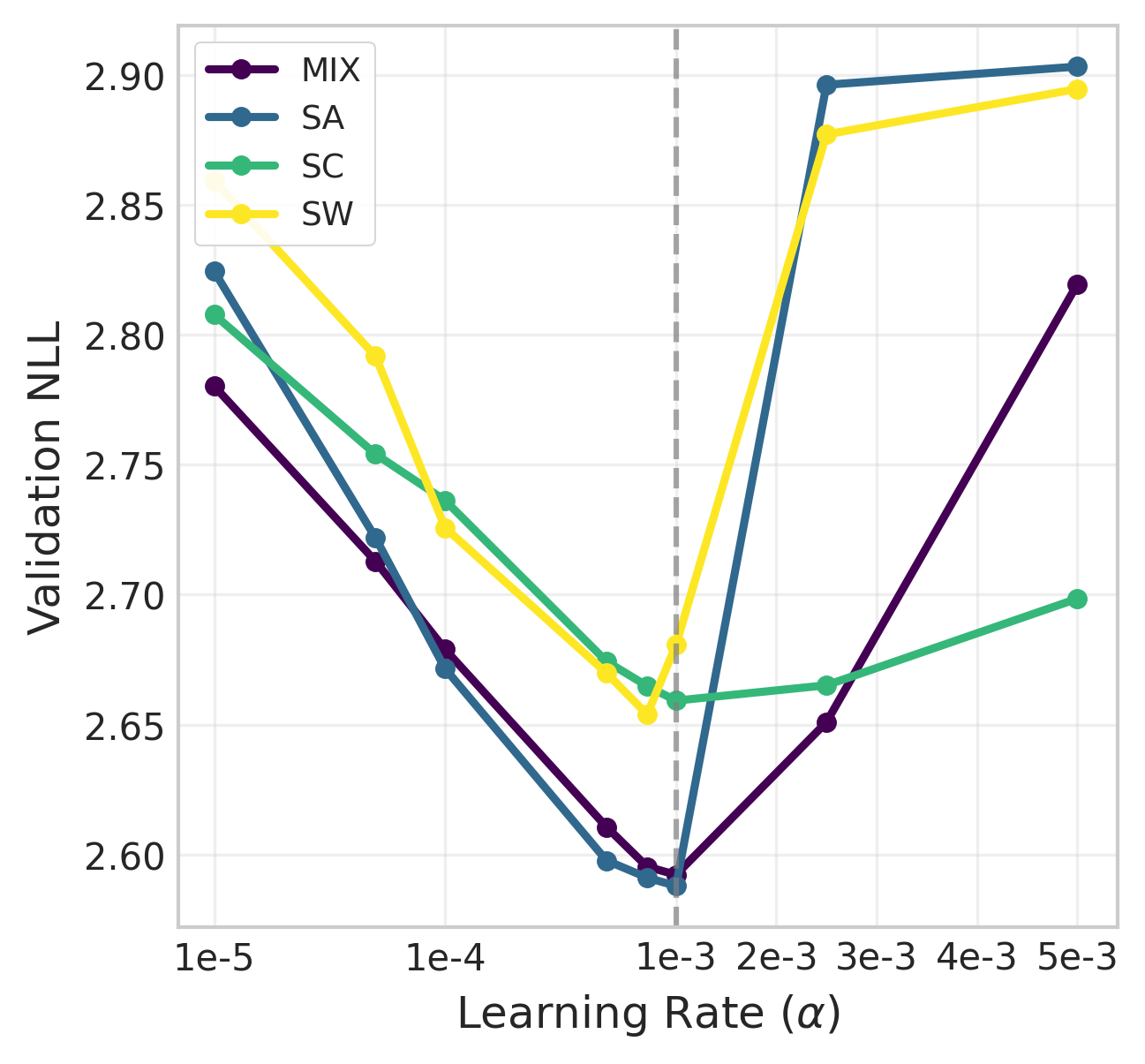}
  \caption{Sequence 35M val NLL}
\end{subfigure}
\begin{subfigure}{0.24\textwidth}
  \includegraphics[width=\linewidth]{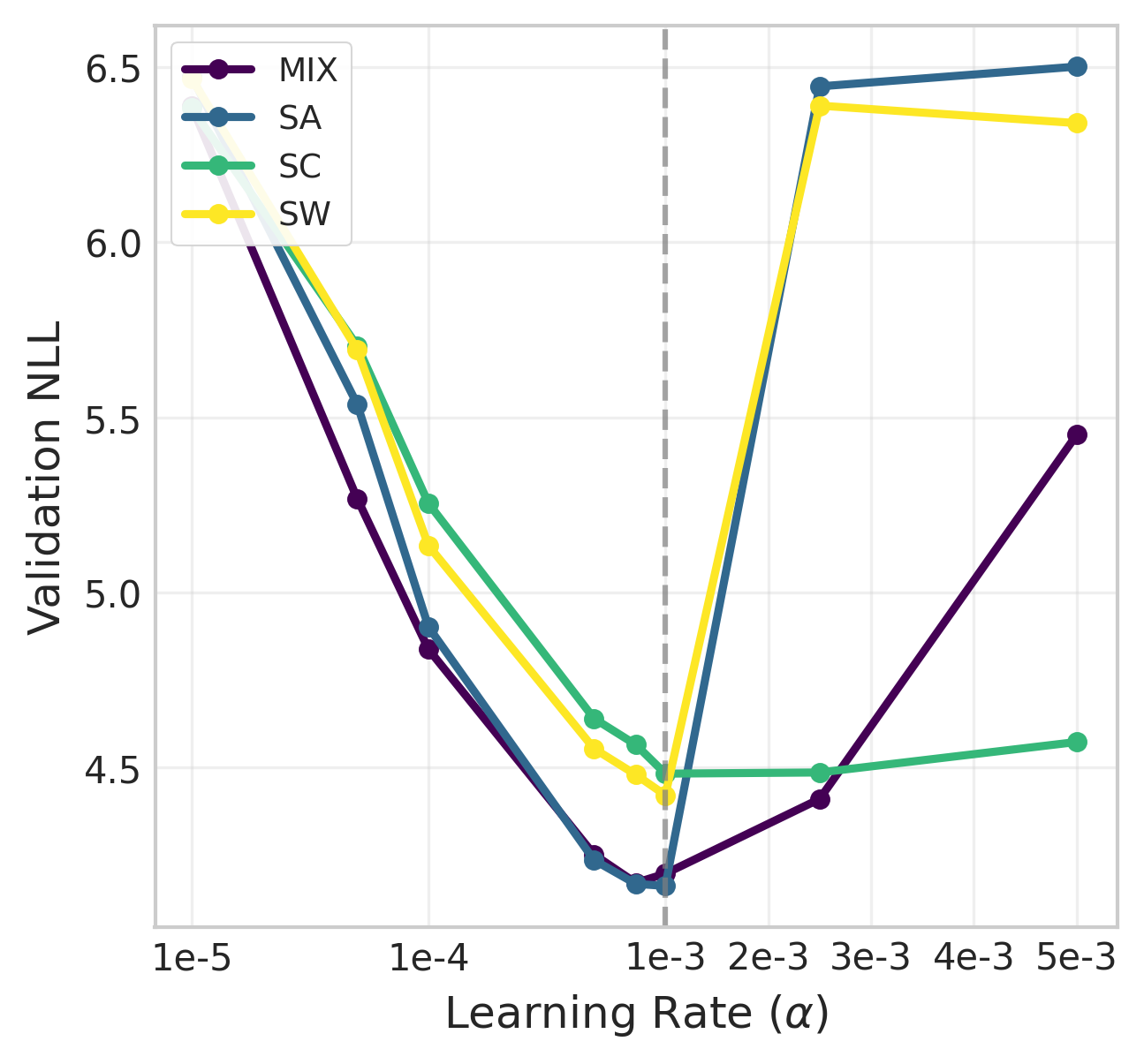}
  \caption{Structure 35M val NLL}
\end{subfigure}
\begin{subfigure}{0.24\textwidth}
  \includegraphics[width=\linewidth]{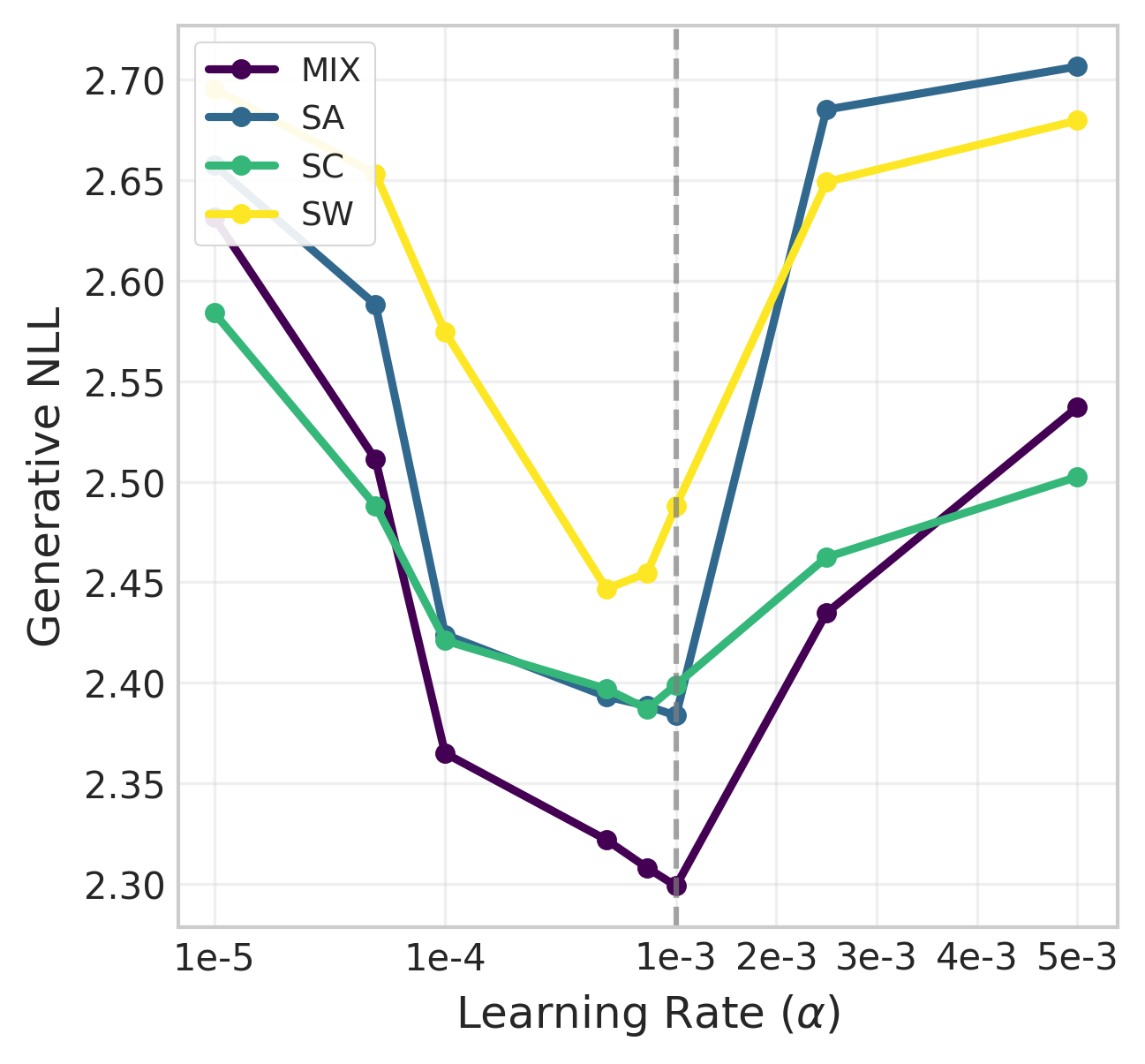}
  \caption{Sequence 35M gen NLL}
\end{subfigure}
\begin{subfigure}{0.24\textwidth}
  \includegraphics[width=\linewidth]{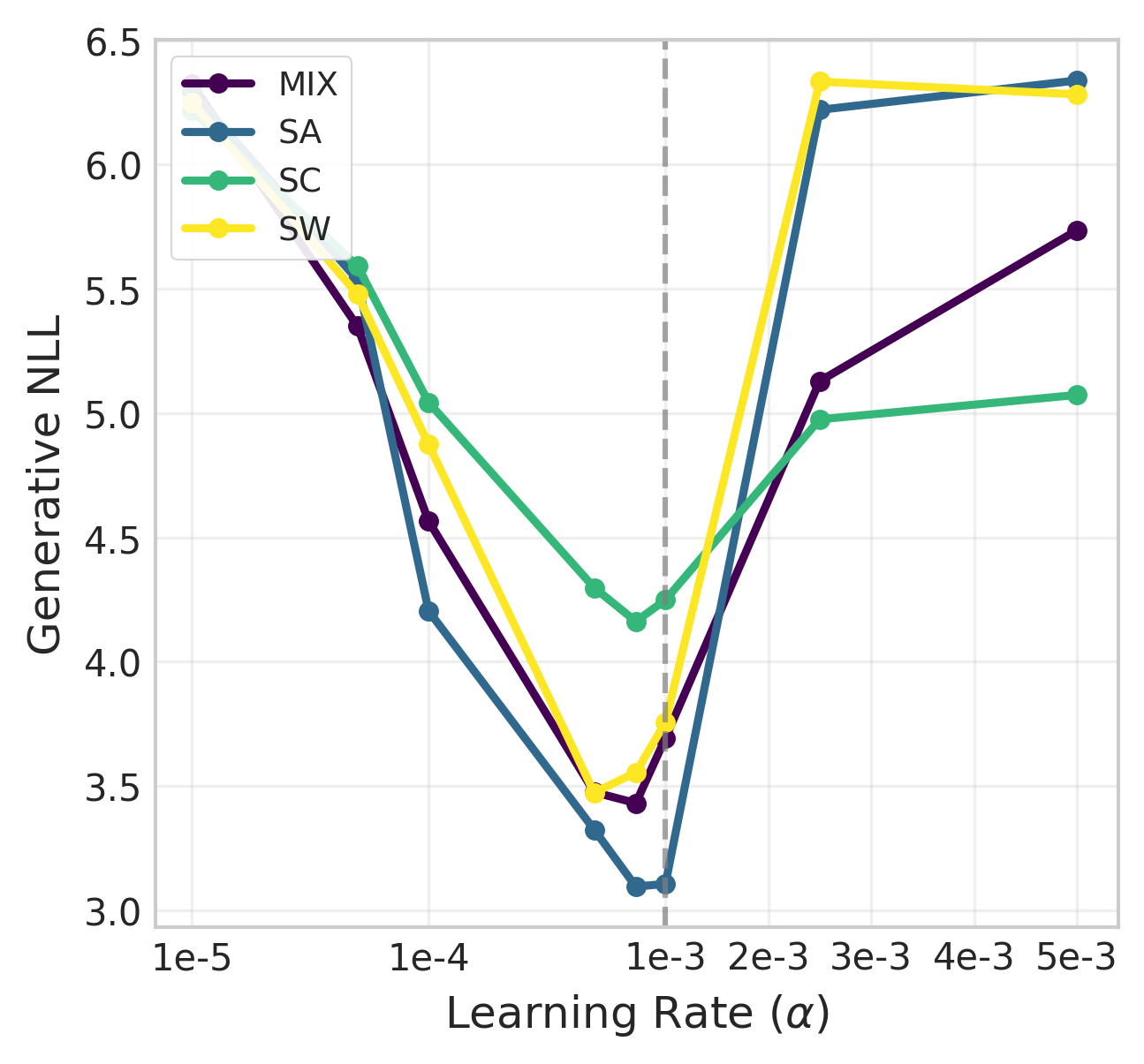}
  \caption{Structure 35M gen NLL}
\end{subfigure}
\caption{Protein sequence and structure-component generative NLL and validation NLL on AlphaFoldDB through learning rate sweep. Learning-rate axis shown on log scale for small values and linear scale for a higher-learning-rate zoom, where instability is observed.}
\label{fig:prot_val_gen_effective_window}
\vspace{-0.75em}
\end{figure*}

\begin{table*}[t!]
\centering
\caption{Protein 320M total, sequence-component, structure-component validation/generative NLLs on AlphaFoldDB.} \label{tab:prot_385M}
\fontsize{8}{9.5}\selectfont
\setlength{\tabcolsep}{7pt}
\renewcommand{\arraystretch}{1.05}
\begin{tabular}{cccccccccc}
\toprule
\multirow{2}{*}{\textbf{Component}} & \multirow{2}{*}{\textbf{Mechanism}}
& \multicolumn{4}{c}{\textbf{Validation NLL}  {\scriptsize (LR sweep)}} & \multicolumn{4}{c}{\textbf{Generative NLL}  {\scriptsize (LR sweep)}} \\
\cmidrule(lr){3-6}\cmidrule(lr){7-10}
& & 1e-4 & 2.5e-4 & 5e-4 & 7.5e-4 & 1e-4 & 2.5e-4 & 5e-4 & 7.5e-4 \\
\midrule
\textbf{Sequence}    & MIX & 2.585 & 2.563 & 2.519 & 2.534 & 2.278 & 2.254 & 2.226 & 2.330 \\
       & SA  & 2.551 & 2.541 & 2.498 & 2.610 & 2.382 & 2.400 & 2.364 & 2.426 \\
\midrule
\textbf{Structure} & MIX & 4.100 & 3.970 & 3.754 & 3.860 & 2.968 & 2.950 & 3.173 & 4.521 \\
       & SA  & 4.008 & 3.917 & 3.711 & 4.253 & 3.396 & 3.277 & 3.257 & 4.492 \\
\bottomrule
\end{tabular}
\vspace{-1.25em}
\end{table*}

\subsection{Learning Rate Sweep Across Modalities and Scales} \label{sec:lr_sweeps}
To isolate learning rate sensitivity, we sweep across learning rate using the AdamW optimizer with default weight decay $1\text{e-}2$, comparing four mechanisms: \textbf{(i)} self-attention (SA), \textbf{(ii)} self-consensus (SC), \textbf{(iii)} sliding-window-attention (SW) \cite{beltagy2020longformerlongdocumenttransformer}, and \textbf{(iv)} a hybrid consensus-attention framework (MIX) where the first half of layers use SA and the remaining layers use SC. For each modality and scale, we train each transformer across a wide range of learning rates and report terminal validation and generative NLLs.

\paragraph{Text.} \label{sec:empirical_text}
We tokenize text using the GPT-2 tokenizer and train on OpenWebText~\cite{Gokaslan2019OpenWeb} via the geometric discrete diffusion noise schedule described in~\citet{odyssey1}. Validation NLLs are computed on a $1\%$ split and generative NLLs are computed using GPT-2-Large as the oracle scorer~\cite{radford2019language}. We report 54M and 193M results in Fig.~\ref{fig:text_val_gen_effective_window} and 385M results in Table~\ref{tab:text_385M}.

\paragraph{DNA.}
We tokenize nucleic-acid sequences by mapping each IUPAC nucleotide code to a unique integer and train on OpenGenome~\cite{OpenGenome2024} using the betalinear30 masked language modeling noise schedule from~\citet{esm3}. Validation NLLs are computed on a $1\%$ split and generative NLLs are computed using Evo2-1B as the oracle scorer~\cite{evo2}. We report 51M and 172M results in Fig.~\ref{fig:dna_val_gen_effective_window} and 331M results in Table~\ref{tab:dna_385M}.

\paragraph{Proteins.} \label{sec:empirical_proteins}
Using the tokenization scheme and pretraining dataset described in~\citet{odyssey1}, we train utilizing masked language modeling on sequence/structure (with the respective betalinear30/cosine noise schedules from~\citet{esm3}). Context secondary structure, SASA, pLDDT, domains, semantic descriptions, and orthologous groups are corrupted with the square-root noise schedule of~\citet{esm3}. Validation NLLs are computed on a $1\%$ split of the AlphaFoldDB~\cite{AlphaFoldDB} subset of~\citet{odyssey1} and generative NLLs are computed using Odyssey-1.2B as the oracle scorer~\cite{odyssey1}. We report 35M results in Fig.~\ref{fig:prot_val_gen_effective_window} and 320M results in Table~\ref{tab:prot_385M}.

Across all three modalities, SC exhibits a wider \textit{stable} learning rate window than SA: as the learning rate is increased beyond the narrow band used in practice, SA and SW incur sharp degradation in terminal NLL, whereas SC degrades more gradually. We note that MIX is designed to leverage the stability margin afforded by SC without ceding attention-level performance. Across all modalities and scales, MIX mirrors SA at the best learning rate while reducing sensitivity to overspecification, yielding a wider effective training window. The results in Tables~\ref{tab:text_385M}, \ref{tab:dna_385M}, and \ref{tab:prot_385M} serve as a scaling case study, demonstrating that the stability and performance benefits of the MIX design persist as model size increases.

\subsection{Estimation of Max Learning Rate From the Hessian}
Using the terminal checkpoints from Section \ref{sec:lr_sweeps}, we estimate a maximum stable learning rate from the loss Hessian encountered during the optimizer updates. For training step $t$, let $\theta_t$ denote the model parameters, $u_t$ be the optimizer step direction at step $t$, and $\alpha$ be the learning rate, so that $\theta_{t+1} = \theta_t + \alpha u_t$. A second-order Taylor expansion for the batch loss $L_t$ around $\theta_t$ gives us:
\begin{align}
\label{eq:loss_ineq}
    &L_t(\theta_{t+1}) - L_t(\theta_t) \leq \\ 
    &\notag \qquad \alpha \langle \nabla L_t, u_t \rangle +\alpha^2 \frac{ \langle u_t, \nabla^2 L_t u_t\rangle}{2} + \alpha^3 O(\|u_t\|^3).
\end{align}
When the step direction is a descent direction for the batch loss, we have that $\langle \nabla L_t, u_t \rangle < 0$, and if the local curvature along the step is positive, then $\langle u_t, \nabla^2 L_t u_t \rangle > 0$. If these conditions hold, we may define:
\begin{align}
    \alpha_{\max}(t) &= -2 \langle \nabla L_t, u_t \rangle \, / \, \langle u_t, \nabla^2 L_t u_t \rangle,
\end{align}
where Eq.~\eqref{eq:loss_ineq} has negative RHS, implying batch loss drops up to second order when $\alpha < \alpha_{\max}(t)$. If $\langle \nabla L_t, u_t \rangle > 0$, we define $\alpha_{\max} =0$. Otherwise, if $\langle u_t, \nabla^2 L_t u_t \rangle < 0$, then $\alpha_{\max} = \infty$. In the case of SGD $u_t=-\nabla L_t(\theta_t)$, we have: 
\begin{equation}
    \alpha_{\max}(t)= \big[ 2\|\nabla L_t\|^2 \big] \, / \,\big[\nabla L_t^\top \nabla^2 L_t\, \nabla L_t \big],
\end{equation}
which implies $\alpha_{\max}(t)\ge 2/\|\nabla^2 L_t\|_2$. Therefore the standard \cite{arora2022understandinggradientdescentedge} SGD stability bound of $\|\nabla^2L_t\|_2 \leq C$ and $\alpha<2/C$ implies $\alpha<\alpha_{\max}(t)$.

\begin{table}[t!]
\caption{Text median $\alpha_{\max}$ and \% stable steps. \textbf{Bold} indicates median $\alpha_{\max} >$ training LR (left) or highest \% Stable (right).} \label{tab:text_hessian}
\centering
\fontsize{8}{9.5}\selectfont
\setlength{\tabcolsep}{4pt}
\renewcommand{\arraystretch}{1.05}
\begin{tabular}{lrrrrrrrr}
\toprule
 & \multicolumn{4}{c}{\textbf{Median} $\boldsymbol{\alpha_{\max}}$} & \multicolumn{4}{c}{\textbf{\% Stable}} \\
 \cmidrule(lr){2-5}\cmidrule(lr){6-9}
\textit{\textbf{LR}} & MIX & SA & SC & SW & MIX & SA & SC & SW \\
\midrule
\multicolumn{9}{c}{\textbf{Size = 54M}} \\
\midrule
\textit{5e-4}   & \textbf{2.2e-3} & \textbf{2.6e-3} & \textbf{2.9e-3} & \textbf{2.2e-3} & \textbf{100} & \textbf{100} & \textbf{100} & \textbf{100} \\
\textit{1e-3}   & \textbf{4.3e-3} & 1.5e-6 & \textbf{6.7e-3} & \textbf{4.1e-3} & \textbf{100} & 0 & \textbf{100} & \textbf{100} \\
\textit{5e-3}   & 1.4e-5 & \textemdash & \textbf{5.5e-3} & 1.1e-5 & 0 & 0 & \textbf{67} & 0 \\
\midrule
\multicolumn{9}{c}{\textbf{Size = 193M}} \\
\midrule
\textit{5e-4}   & \textbf{0.01} & \textbf{4.5e-3} & \textbf{0.02} & \textbf{0.01} & \textbf{100} & 92 & \textbf{100} & 64 \\
\textit{7.5e-4} & 2.1e-6 & 6.4e-7 & \textbf{8.4e-3} & \textbf{0.02} & 0 & 0 & \textbf{100} & 56 \\
\textit{1e-3}   & 9.6e-7 & 3.5e-7 & \textbf{4.9e-3} & \textbf{0.01} & 0 & 0 & \textbf{92} & 56 \\
\bottomrule
\end{tabular}
\vspace{-0.75em}
\end{table}

For AdamW, the step direction $u_t$ depends on the optimizer state $(m_t,v_t)$ rather than on $-\nabla L_t$ alone, so classical SGD step size bounds based on $\|\nabla^2 L_t\|_2$ do not directly apply. In this setting, $\alpha_{\max}(t)$ is a directional and state-dependent stability threshold for scaling the current update $u_t$.

Previous works in optimization have leveraged global curvature (e.g., $\|\nabla^2 L\|_2$ or alternative norms like $\|\nabla^2 L\|_{1,1}$) to quantify roughness of the loss landscape. These quantities either appear as global smoothness assumptions for proving convergence rates, or only apply to SGD. In contrast, $\langle u_t,\nabla^2 L_t u_t\rangle$ measures curvature in the update direction, yielding a local estimate of the maximal stable step size at any point in training, given an optimizer and model state.

\subsubsection{Hessian Experiment Details and results} \label{sec:hessian_experiments}
We perform Hessian analysis for the 54M/51M/35M and 193M/172M/133M terminal checkpoints across all modalities and mechanisms. For all computations, we clip the $\ell_2$-norm of the gradient to $1$ as in training. For each checkpoint, we run a warmup over $5$ batches without stepping to initialize AdamW statistics: $m$ is the average gradient over the $5$ batches, and $v$ the average squared gradient. Due to computational constraints, we use a batch size of $512$ for 54M/51M/35M and $64$ for 193M/172M/133M. We then run $5$ AdamW steps under default hyperparameters (with bias correction using step count starting at $6$) for further warmup, and record $\alpha_{\max}(t)$ using a Hessian-vector product via PyTorch AutoGrad \cite{paszke2019pytorch} (cost $\mathcal{O}(\text{dim}(\theta))$. We take $25$ AdamW steps after warmup, recording the $\alpha_{\max}$ of each step. For SW, our band-masked implementation yields numerically unstable AutoGrad Hessian-vector products, so we approximate $Hv$ via central finite difference ($\epsilon=1e\text{-}4$):
\begin{align}
    Hv \;\approx\; \big[\nabla L(\theta+\epsilon v)-\nabla L(\theta-\epsilon v)\big] \, / \, (2\epsilon).
\end{align}
We validate this estimator on the other mechanisms (where AutoGrad HVPs are stable). For runs that achieve a suboptimal loss, the median non-infinite $\alpha_{\max}(t)$ typically falls below the training learning rate. Across all modalities and scales, SC attains the largest median $\alpha_{\max}(t)$ and the highest fraction of stable steps, per Tables~\ref{tab:text_hessian}, \ref{tab:dna_hesssian}, and \ref{tab:prot_hesssian}.

\begin{table}[t!]
\caption{DNA median $\alpha_{\max}$ and \% stable steps. \textbf{Bold} indicates median $\alpha_{\max} >$ training LR (left) or highest \% Stable (right).}  \label{tab:dna_hesssian}
\centering
\fontsize{8}{9.5}\selectfont
\setlength{\tabcolsep}{4pt}
\renewcommand{\arraystretch}{1.05}
\begin{tabular}{lrrrrrrrr}
\toprule
 & \multicolumn{4}{c}{\textbf{Median} $\boldsymbol{\alpha_{\max}}$} & \multicolumn{4}{c}{\textbf{\% Stable}} \\
\cmidrule(lr){2-5}\cmidrule(lr){6-9}
\textit{\textbf{LR}} & MIX & SA & SC & SW & MIX & SA & SC & SW \\
\midrule
\multicolumn{9}{c}{\textbf{Size = 51M}} \\
\midrule
\textit{1e-3}   & \textbf{3.0e-3} & \textbf{6.2e-3} & \textbf{0.01} & \textbf{0.02} & \textbf{100} & 96 & 96 & 52 \\
\textit{2.5e-3} & 9.6e-4          & 1.4e-7          & \textbf{0.08} & 8.7e-5        & 20           & 0  & \textbf{96} & 0 \\
\textit{5e-3}   & 9.0e-6          & 4.1e-6          & \textbf{0.06} & 2.7e-6        & 0            & 0  & \textbf{96} & 0 \\
\midrule
\multicolumn{9}{c}{\textbf{Size = 172M}} \\
\midrule
\textit{5e-4}   & \textbf{2.5e-3} & \textbf{2.3e-3} & \textbf{1.1e-3} & \textbf{1.5e-3} & 84 & \textbf{96} & 80 & 76 \\
\textit{7.5e-4} & 5.3e-4          & 7.5e-5          & \textbf{1.8e-3} & \textbf{2.1e-3} & 28 & 4  & \textbf{92} & 76 \\
\textit{1e-3}   & 3.7e-5          & 1.7e-5          & \textbf{2.5e-3} & 4.4e-4          & 0  & 0  & \textbf{84} & 36 \\
\bottomrule
\end{tabular}
\vspace{-0.75em}
\end{table}

\begin{table}[t!]
\caption{Protein median $\alpha_{\max}$ and \% stable steps. \textbf{Bold} indicates median $\alpha_{\max} >$ training LR (left) or highest \% Stable (right).} \label{tab:prot_hesssian}
\centering
\fontsize{8}{9.5}\selectfont
\setlength{\tabcolsep}{4pt}
\renewcommand{\arraystretch}{1.05}
\begin{tabular}{lrrrrrrrr}
\toprule
 & \multicolumn{4}{c}{\textbf{Median} $\boldsymbol{\alpha_{\max}}$} & \multicolumn{4}{c}{\textbf{\% Stable}} \\
\cmidrule(lr){2-5}\cmidrule(lr){6-9}
\textit{\textbf{LR}} & MIX & SA & SC & SW & MIX & SA & SC & SW \\
\midrule
\multicolumn{9}{c}{\textbf{Size = 35M}} \\
\midrule
\textit{1e-3}   & \textbf{3.7e-3} & \textbf{7.2e-3} & \textbf{4.0e-3} & \textbf{4.8e-3} & \textbf{100} & \textbf{100} & \textbf{100} & \textbf{100} \\
\textit{2.5e-3} & \textbf{0.01}   & 3.8e-8          & \textbf{6.8e-3} & 2.5e-5          & 88            & 0             & \textbf{96}    & 0 \\
\textit{5e-3}   & 2.1e-4          & 8.5e-7          & 1.8e-3          & 8.5e-5          & 4             & 0             & \textbf{12}    & 4 \\
\midrule
\multicolumn{9}{c}{\textbf{Size = 133M}} \\
\midrule
\textit{5e-4}   & \textbf{5.6e-3} & \textbf{7.6e-3} & \textbf{2.8e-3} & \textbf{2.7e-3} & 96            & \textbf{100}  & \textbf{100}   & 80 \\
\textit{7.5e-4} & \textbf{6.7e-3} & 1.3e-4          & \textbf{3.8e-3} & \textbf{4.2e-3} & \textbf{100}  & 12            & \textbf{100}   & 60 \\
\textit{1e-3}   & \textbf{7.4e-3} & 1.2e-6          & \textbf{6.8e-3} & 3.1e-5          & 84            & 0             & \textbf{100}   & 0 \\
\bottomrule
\end{tabular}
\vspace{-0.75em}
\end{table}

\section{Conclusion} \label{sec:conclusion}
In this paper, we reformulated consensus using graph spectral theory, relating information propagation to algebraic connectivity and deriving bounds for graph classes relevant to sequential data. Across text, DNA, and proteins, we found that consensus was more robust to learning rate overspecification than attention, and that a hybrid consensus-attention architecture preserved attention-level performance while inheriting the stability benefits of consensus at scale. We further connected these empirical findings to effective step sizes by estimating a directional maximum stable learning rate from the loss Hessian along optimizer update directions. We note that consensus requires a choice of graph connectivity that is not always available \textit{a priori}, motivating future work on adaptive or learned graph construction.

\section*{Impact Statement}
This paper improves the training stability of transformers by increasing robustness to learning-rate overspecification. By reducing the need for extensive hyperparameter tuning and mitigating training instabilities, the method may lower the practical cost and expertise required to train large generative models, which can benefit engineering applications but may also broaden access to capable generative modeling in ways that raise familiar dual-use concerns.

\bibliography{main}
\bibliographystyle{icml2025}

\newpage

\appendix
\onecolumn
\section{Additional Theoretical Background} \label{appendix:theory}
We extend the scalar-valued graph-signal formulation in Section~\ref{subsec:effects_filtering} of the main text to the vector-valued setting used by consensus. For completeness, we summarize the corresponding block-matrix construction and notation below.

\subsection{Vector-Valued Signals}
We construct a block adjacency matrix $W\in \mathbb{R}^{Nd\times Nd}$ as:
\begin{equation}
    W= \begin{bmatrix}
        W_{0,0} &  \cdots & W_{0, N-1}\\
        \vdots & \ddots & \vdots \\
        W_{N-1,0} & \cdots & W_{N-1,N-1}
    \end{bmatrix},
\end{equation}
where $W_{i,j}\in\mathbb{R}^{d\times d}$ is positive definite ($W_{i,j}\succ 0$) when nodes $i$ and $j$ share an edge, and $W_{i,j}=0_{d\times d}$ otherwise. We define the \textit{per-node out-degree} matrices:
\begin{equation}
    D_i \;=\; \sum\nolimits_{j\neq i} W_{i,j},
\end{equation}
and assemble the (block-diagonal) degree matrix $D\in \mathbb{R}^{Nd\times Nd}$ as:
\begin{equation}
    D = \begin{bmatrix}
        D_0 & \cdots & 0 \\
        \vdots & \ddots & \vdots \\
        0 & \cdots & D_{N-1}
    \end{bmatrix}.
\end{equation}
We then construct the (generally non-symmetric) Laplacian $L = D - W \in \mathbb{R}^{Nd \times Nd}$ and symmetrize it as $L_\text{sym} = \tfrac{1}{2}(L + L^{\top})$. Analogous to the scalar-valued case, $L_\text{sym}$ has $d$ zero eigenvalues, $\lambda_0=\lambda_1=\cdots=\lambda_{d-1}=0$, with corresponding eigenvectors given by block-permutations of:
\begin{equation}
    v_0 = \mathrm{cat}(1_d, 0_d, \cdots, 0_d)\in\mathbb{R}^{Nd}.
\end{equation}
The consensus update is given by Eq.~\eqref{eq:consensus_update}, where indexing and summation are interpreted over the $d\times d$ block structure.

\section{Proofs and Derivations} \label{appendix:proofs_and_derivations}
We present proofs and derivations for the remarks outlined in Section \ref{sec:analysis} of the main text.

\begin{proof}[Proof of Remark~\ref{remark:circulant_graph}]
The resulting graph Laplacian $L$ is a matrix whose columns are populated by the cyclic permutations of the vector, $\Pi_N^w \in \mathbb{R}^N$, where $\Pi_N^w$ is given by:
\begin{align*}
    \Pi_N^w = \begin{bmatrix}
        2w & 
        \smash{\underbrace{\begin{matrix}-1 & \cdots & -1\end{matrix}}_{w}} &
        \smash{\underbrace{\begin{matrix}0 & \cdots & 0 \end{matrix}}_{N - 2w - 1}} &
        \smash{\underbrace{\begin{matrix}-1 & \cdots & -1\end{matrix}}_{w}} 
    \end{bmatrix}^{\top}.\\
\end{align*}
Following \cite{salahub2022approximating}, we calculate the eigenvalues of this symmetric circulant matrix.  Of note, we see that for any $k_0 \in \mathbb{R}$, the following is an eigenvector for $\lambda_1$:
\begin{align}\label{circulantEigenvector}
    \phi_k^{N, k_0} = \cos\left(\frac{2\pi (k+k_0)}{N}\right), \quad \text{for:} \quad 0 \leq k < N.
\end{align}
\end{proof}

\begin{proof}[Proof of Remark~\ref{remark:path_graph}]
We leverage a comparison to the Laplacian of $C^{w}_{2N} = (V,E)$. Let $L$ and $\overline{L}$ be the graph Laplacians of $P^w_N$ and $C^w_{2N}$ respectively. Let $V_1 = \{0,...,N-1\}$ and $V_2 = \{N,...,2N-1\}$ such that $V = V_1 \sqcup V_2$. Let $G_1$ and $G_2$ be the induced subgraphs of $V_1$ and $V_2$ respectively with edge sets $E_1$ and $E_2$, and note that $G_1$ and $G_2$ are isomorphic to $P^w_N$. Define the following 2 to 1 projection $\pi : V \to V_1$:
\begin{align}
    \pi(k) = \begin{cases}
        k & \text{if}\ 0 \leq k \leq N-1 \\
        2N-1-k & \text{if}\ N \leq k \leq 2N -1.
    \end{cases}
\end{align}
For a function $v$ on $V_1$, let $\overline{v}$ be the following extension of $v$ on all of $V$ given by $\overline{v}_k = v_{\pi(k)}$. Thus we have $\|\overline{v}\|^2 = 2\|v\|^2$, and $\overline{v} \perp \mathbf{1}_{2N}$ if and only if $v \perp \mathbf{1}_N$. Subsequently, we have that:
\begin{align}
    \langle \overline v, \overline{L} \overline{v}\rangle &=  \sum\nolimits_{\{i,j\} \in E} (\overline{v}_i - \overline{v}_j)^2 \\
    &= \sum\nolimits_{\{i,j\} \in E_1} (\overline{v}_i - \overline{v}_j)^2 + \sum\nolimits_{\{i,j\} \in E_2} (\overline{v}_i - \overline{v}_j)^2 +\sum\nolimits_{\{i,j\} \in E \setminus (E_1 \cup E_2)} (\overline{v}_i - \overline{v}_j)^2 \\ \label{eq:path_graph_power_expanded}
     &= 2 \sum\nolimits_{\{i,j\} \in E_1} (v_i - v_j)^2 +\sum\nolimits_{\{i,j\} \in E \setminus (E_1 \cup E_2)} (v_{\pi(i)} - v_{\pi(j)})^2.
\end{align}
To upper bound the second term, we rely upon the assumption that $w < N/2$ and note that the projection $\pi$ satisfies $\{\pi(i), \pi(j)\} \in E_1$ whenever $\pi(i)\neq \pi(j)$ and $\{i,j\} \in E$. We also note that the second sum is indexed by edges that cross $V_1$ and $V_2$, i.e., in each term, exactly one of $i, j$ is in $V_1$ and the other index in $V_2$. Given $k,l \in E_1$, there's at most 2 cross-edges that project to to $\{k,l\}$ under $\pi$. Hence, after this double counting, we obtain:
\begin{align}
    \langle \overline v, \overline{L} \overline{v}\rangle &\leq 2 \sum\nolimits_{\{i,j\} \in E_1} (v_i - v_j)^2 + 2 \sum\nolimits_{\{i,j\} \in E_1} (v_i - v_j)^2 \leq 4 \langle v, L v\rangle.
\end{align}
Next, since the second term of Eq.~\eqref{eq:path_graph_power_expanded} is non-negative, we automatically get $2 \langle v, Lv \rangle \leq \langle \overline{v} , \overline{L} \overline{v}\rangle$ and thus:
\begin{equation}
\label{eq:ineq_summary}
    2 \braket{v, Lv } \leq \braket{\overline{v}, \overline{L} \overline{v}} \leq 4 \braket{v, Lv}.
\end{equation}
Using Eq.~\eqref{eq:ineq_summary} and the fact that $\|\overline{v}\|^2 = 2\|v\|^2$, we have that:
\begin{align}
    \lambda_1(C^{R}_{2N}) \leq \frac{\langle\overline{v}, \overline{L} \overline{v} \rangle}{\|\overline{v}\|^2} \leq 2 \frac{\langle v, Lv \rangle}{\|v\|^2},
\end{align}
by the variational characterization of $\lambda_1$. Minimizing the RHS over all $v \perp \mathbf{1}$ yields:
\begin{align}
\label{eq:rayleigh}
     \frac{1}{2}\lambda_1(C^{R}_{2N}) \leq \lambda_1(P^w_N).
\end{align}
Consider the eigenfunction $\phi = \phi^{2N, \tfrac{1}{2}}$ of $\bar{L}$ defined in Eq.~\eqref{circulantEigenvector} corresponding to eigenvalue $\lambda_1(C_{2N}^w)$ on $C_{2N}^w$. A simple computation shows that $\phi_k = \phi_{\pi(k)}$ and hence $\phi = \bar{\gamma}$ for some function (not necessarily an eigenfunction) $\gamma$ on $G_1$. Thus:
\begin{align}
    \lambda_1(P^w_N) \leq \frac{\langle \gamma, L\gamma \rangle}{\|\gamma\|^2} \leq \frac12 \frac{\langle \phi, \overline{L} \phi \rangle}{\|\phi \|^2/2} = \lambda_1(C^w_{2N}).
\end{align}
Finally, we have that:
\begin{equation}
    \frac12 \lambda_1(C_{2N}^w) \leq \lambda_1(P_N^w) \leq\lambda_1(C^w_{2N}),
\end{equation}
and the remark follows from the values of $\lambda_1(C_{2N}^w)$ provided in Remark~\ref{remark:circulant_graph}.
\end{proof}

\section{Variants of the Consensus Mechanism} \label{appendix:consensus}
In this section, we formalize the cross-consensus, multi-headed self-consensus, and multi-headed cross-consensus mechanisms described in the main text. We further outline an adaptation of rotary positional embeddings (RoPE) for consensus.

Across all algorithms, we define the following functions.  For $\Psi \in \mathbb{R}^{\Omega \times \omega}$,  we define $\text{RS}_{\Omega, \omega} : \mathbb{R}^{\Omega \omega} \mapsto \mathbb{R}^{\Omega \times \omega}$ (reshape) and $\text{RN} : \mathbb{R}^{\Omega \times \omega} \mapsto \mathbb{R}^{\Omega \times \omega}$ (row-norm) by:
\begin{align} \label{eq:reshape}
    \left[\text{RS}_{\Omega, \omega}(\Psi)\right]_{i,j} &= \Psi_{i \omega + j},  
\quad\quad\quad
    \left[\text{RN}(\Psi)\right]_{i,j} =  \begin{cases}
        {\Psi_{i,j}}/{ \| \Psi_{i,:} \| } & \text{ if } \| \Psi_{i,:} \| > 0, \\
        \Psi_{i,j} & \text{ else}. 
    \end{cases} 
\end{align}
We define $\sigma^+(\cdot)$ to be the softplus activation function, and denote SCWM and CCWM as the Self-Consensus Weight Matrix (Algorithm~\ref{alg:self-weight-network}) and the Cross-Consensus Weight Matrix (Algorithm~\ref{alg:cross-weight-network}), respectively. We note that $b_s$ cancels in the consensus update; as such, practitioners may omit it from Algorithm~\ref{alg:self-consensus} without affecting computation.

\subsection{Cross-Consensus}
In multimodal applications (e.g., the protein setting in Section~\ref{sec:empirical_results}), we seek to imbue information from an auxiliary \textit{context} modality into the \textit{source} modality. Let the source embeddings be $y = [y^{(0)},\dots,y^{(N-1)}]^\top \in \mathbb{R}^{N\times d}$, and let the context embeddings be $c = [c^{(0)},\dots,c^{(J-1)}]^\top \in \mathbb{R}^{J\times d}$. We outline cross-consensus, a drop-in replacement for cross-attention, which performs a consensus-style update on a bipartite graph.

Let $V$ and $V'$ be node sets with $|V|=N$ and $|V'|=J$, indexing the source and context nodes, respectively. We define a bipartite directed graph $\mathcal{G}_+ = (V \cup V', E_+)$ with edge set $E_+ \subset \{ (i,j) \in V \times V'\}$. Given projected embeddings $(u^{(i)})_{i\in V}$ and $(v^{(j)})_{j\in V'}$, cross-consensus attempts to decrease the bipartite energy:
\begin{align}
  \mathcal{E}(u, v) = \frac{1}{2} \sum_{(i,j)\in E_+} \big(u^{(i)} - v^{(j)} \big)^\top R^{(i,j)} \big(u^{(i)} - v^{(j)} \big),
\end{align}
where $R^{(i,j)}\in\mathbb{R}^{d\times d}$ is the positive-definite consensus weight matrix associated with edge $(i,j)\in E_+$, connecting source node $i$ to context node $j$. We first outline an algorithm to compute these consensus weight matrices for the edges of $\mathcal{G}_+$.

\begin{algorithm}[h!]
\caption{Cross-Consensus Weight Matrix (CCWM)} \label{alg:cross-weight-network}
\begin{algorithmic}[1]
\REQUIRE Source embeddings $(y^{(i)})_{i=0}^{N-1}$, context embeddings $(c^{(j)})_{j=0}^{J-1}$, bipartite directed graph $\mathcal{G}_+=(V\cup V',E_+)$
\STATE \textbf{Hyperparameters:} embedding dimension $d$, rank $r$, edge hidden dimension $\xi$
\STATE \textbf{Parameters:} $\mathcal{W} = (W_\alpha, W_\beta, W_\Lambda) \in \mathbb{R}^\xi \times \mathbb{R}^\xi \times \mathbb{R}^{rd\times \xi}$,
$\mathcal{B} = (b_\alpha, b_\beta, b_\Lambda) \in \mathbb{R} \times \mathbb{R} \times \mathbb{R}^{rd}$
\STATE \textbf{Edge MLPs:} $\phi^\alpha_{\theta_\alpha}, \phi^\beta_{\theta_\beta}, \phi^\Lambda_{\theta_\Lambda}: \mathbb{R}^d \times \mathbb{R}^d \mapsto \mathbb{R}^\xi$
with parameters $\Theta = (\theta_\alpha, \theta_\beta, \theta_\Lambda)$
\FORALL{$(i,j)\in E_+$}
  \STATE $\alpha^{(i,j)} \leftarrow \sigma^+\!\big(W_\alpha \phi^\alpha_{\theta_\alpha}([y^{(i)};c^{(j)}]) + b_\alpha\big)$
  \STATE $\beta^{(i,j)} \leftarrow \sigma^+\!\big(W_\beta \phi^\beta_{\theta_\beta}([y^{(i)};c^{(j)}]) + b_\beta\big)$
  \STATE $\Lambda^{(i,j)} \leftarrow \text{RN}\!\Big(\text{RS}_{r,d}\!\big(W_\Lambda \phi^\Lambda_{\theta_\Lambda}([y^{(i)};c^{(j)}]) + b_\Lambda\big)\Big)/\sqrt{r}$
  \STATE $R^{(i,j)} \leftarrow \alpha^{(i,j)} I_d + \beta^{(i,j)} (\Lambda^{(i,j)})^\top \Lambda^{(i,j)}$
\ENDFOR
\ENSURE Edge weights $R = \{R^{(i,j)} : (i,j) \in E_+\}$
\end{algorithmic}
\end{algorithm}

We next outline cross-consensus, which performs a source-track update using the edge weights computed by Algorithm~\ref{alg:cross-weight-network}.

\begin{algorithm}[h!]
\caption{Cross-Consensus Mechanism}
\label{alg:cross-consensus}
\begin{algorithmic}[1]
\REQUIRE Source embeddings $\{y^{(i)}\}_{i=0}^{N-1}$, context embeddings $\{c^{(j)}\}_{j=0}^{J-1}$, bipartite directed graph $\mathcal{G}_+=(V\cup V',E_+)$

\STATE \textbf{Hyperparameters:} step size $\eta>0$, embedding dimension $d$, rank $r$, edge hidden dimension $\xi$
\STATE \textbf{Parameters:} $W_s \in \mathbb{R}^{d \times d}, b_s \in \mathbb{R}^d, W_c \in \mathbb{R}^{d \times d}, b_c \in \mathbb{R}^d,$
$W_o \in \mathbb{R}^{d \times d}, b_o \in \mathbb{R}^d,$ and CCWM parameters $\mathcal{W}, \mathcal{B}, \Theta$

\FOR{$i=0$ to $N-1$}
  \STATE $u^{(i)} \leftarrow W_s\, y^{(i)} + b_s$ \linelabel{alg:line:ccSourceProj}
\ENDFOR
\FOR{$j=0$ to $J-1$}
  \STATE $v^{(j)} \leftarrow W_c\, c^{(j)} + b_c$ \linelabel{alg:line:ccContextProj}
\ENDFOR

\STATE $R \leftarrow \text{CCWM}(y, c, G_+ \mid \mathcal{W}, \mathcal{B}, \Theta, d, r, \xi)$

\FOR{$i=0$ to $N-1$}
  \STATE $g^{(i)} \leftarrow \sum_{j:(i,j)\in E_+} R^{(i,j)}\big(u^{(i)}-v^{(j)}\big)$
  \STATE $u'^{(i)} \leftarrow u^{(i)} - \eta\, g^{(i)}$
  \STATE $y_{\mathrm{out}}^{(i)} \leftarrow W_o u'^{(i)} + b_o$
\ENDFOR

\ENSURE Embeddings $(y_{\mathrm{out}}^{(i)})_{i=0}^{N-1}$
\end{algorithmic}
\end{algorithm}

If the source and context embeddings correspond to a common sequence and thus have a common length ($J=N$), one may optionally choose a windowed bipartite edge set:
\begin{equation}
    E_N^w = \{(i,j) : 0 \leq i,j < N \ \land\ |i-j| \leq w \}.
\end{equation}

\subsection{Multi-Head Consensus}
As an analog of multi-head attention~\cite{attention_all_you_need}, consensus admits a multi-head generalization in which $H$ heads operate in parallel on disjoint feature subspaces. We set $H\in\mathbb{N}$ and assume $d_H \coloneqq d/H \in \mathbb{Z}$. For each node $i$, we view the projected state $u^{(i)}\in\mathbb{R}^{d}$ as a concatenation of $H$ head sub-vectors:
\begin{equation}
    u^{(i)} = \big[ u^{(i)}_0;\, \cdots;\, u^{(i)}_{H-1} \big], \quad \text{where:} \quad u^{(i)}_h \in \mathbb{R}^{d_H}.
\end{equation}
In multi-headed consensus, each head $h$ performs its own consensus update on $\{u_h^{(i)}\}_{i=0}^{N-1}$ using head-specific edge weights $\{R_h^{(i,j)}\}_{(i,j)\in E}$ computed by a head-specific weight network. The heads are processed independently and are concatenated only at the output projection stage.

\paragraph{Multi-Head Self-Consensus.}
Algorithm~\ref{alg:multi-head-self-consensus} implements the multi-head self-consensus mechanism on a directed graph $\mathcal{G}=(V,E)$ with $|V|=N$. The mechanism first applies a shared source projection, then splits $u^{(i)}$ into $\{u_h^{(i)}\}_{h=0}^{H-1}$. For each head $h$, we produce positive-definite edge weights $R_h^{(i,j)}\in\mathbb{R}^{d_H\times d_H}$. The head-wise update produces $\{u_h'^{(i)}\}$, which are concatenated to form $u'^{(i)}$ and mapped to the output embedding.

\begin{algorithm}[h!]
\caption{Multi-Head Self-Consensus Mechanism}
\label{alg:multi-head-self-consensus}
\begin{algorithmic}[1]
\REQUIRE Embeddings $\{y^{(i)}\}_{i=0}^{N-1}$, directed graph $\mathcal{G}=(V,E)$ with $|V|=N$

\STATE \textbf{Hyperparameters:} step size $\eta>0$, embedding dimension $d$, rank $r$, edge hidden dimension $\xi$, number of heads $H$, and $d_H=d/H$
\STATE \textbf{Parameters:} $W_s \in \mathbb{R}^{d \times d}$, $b_s \in \mathbb{R}^{d}$, $W_o \in \mathbb{R}^{d \times d}$, $b_o \in \mathbb{R}^{d}$
\STATE \hspace{1.7em} Head-specific SCWM parameters $\{(\mathcal{W}_h,\mathcal{B}_h,\Theta_h)\}_{h=0}^{H-1}$ with
\STATE \hspace{1.7em} $\mathcal{W}_h=(W_{\alpha,h},W_{\beta,h},W_{\Lambda,h})$, $\mathcal{B}_h=(b_{\alpha,h},b_{\beta,h},b_{\Lambda,h})$, $\Theta_h=(\theta_{\alpha,h},\theta_{\beta,h},\theta_{\Lambda,h})$

\FOR{$i=0$ to $N-1$}
  \STATE $u^{(i)} \leftarrow W_s\, y^{(i)} + b_s$
  \STATE $\big[u^{(i)}_0;\cdots;u^{(i)}_{H-1}\big] \leftarrow u^{(i)}$ \COMMENT{$u_h^{(i)}\in\mathbb{R}^{d_H}$}
\ENDFOR

\FOR{$h=0$ to $H-1$}
  \STATE $R_h \leftarrow \text{SCWM}(y,\mathcal{G}\mid \mathcal{W}_h,\mathcal{B}_h,\Theta_h,d_H,r,\xi)$
  \FOR{$i=0$ to $N-1$}
    \STATE $g_h^{(i)} \leftarrow 
      \sum_{j:(i,j)\in E} R_h^{(i,j)}\big(u_h^{(i)}-u_h^{(j)}\big)
      - \sum_{k:(k,i)\in E} R_h^{(k,i)}\big(u_h^{(k)}-u_h^{(i)}\big)$
    \STATE $u_h'^{(i)} \leftarrow u_h^{(i)} - \eta\, g_h^{(i)}$
  \ENDFOR
\ENDFOR

\FOR{$i=0$ to $N-1$}
  \STATE $u'^{(i)} \leftarrow \big[u_0'^{(i)};\cdots;u_{H-1}'^{(i)}\big]$
  \STATE $y_{\mathrm{out}}^{(i)} \leftarrow W_o\, u'^{(i)} + b_o$
\ENDFOR

\ENSURE Output embeddings $\{y_{\mathrm{out}}^{(i)}\}_{i=0}^{N-1}$
\end{algorithmic}
\end{algorithm}

\paragraph{Multi-Head Cross-Consensus.}
Algorithm~\ref{alg:multi-head-cross-consensus} outlines the multi-head cross-consensus mechanism on a directed bipartite graph $\mathcal{G}^+=(V\cup V',E^+)$ with $|V|=N$ and $|V'|=J$. The mechanism first applies shared source and context projections, then splits $u^{(i)}$ and $v^{(j)}$ into $u_h^{(i)}, v_h^{(j)} \in \mathbb{R}^{d_H}$. For each head $h$, we produce positive-definite edge weights $R_h^{(i,j)}\in\mathbb{R}^{d_H\times d_H}$ for $(i,j)\in E^+$. The head-wise update produces $\{u_h'^{(i)}\}$ by aggregating discrepancies to $\{v_h^{(j)}\}$. The $\{u_h'^{(i)}\}$ are concatenated to form $u'^{(i)}$ and mapped to the output embedding. We note that in multi-head cross-consensus, we similarly expand: 
\begin{equation}
    v^{(j)} = \big[ v^{(j)}_0, \cdots, v^{(j)}_{H-1} \big], \quad \text{where:} \quad v^{(j)}_h \in \mathbb{R}^{d_H}.
\end{equation}

\begin{algorithm}[t!]
\caption{Multi-Head Cross-Consensus Mechanism}
\label{alg:multi-head-cross-consensus}
\begin{algorithmic}[1]
\REQUIRE Source embeddings $\{y^{(i)}\}_{i=0}^{N-1}$, context embeddings $\{c^{(j)}\}_{j=0}^{J-1}$, bipartite directed graph $\mathcal{G}^+=(V\cup V',E^+)$ with $|V|=N$, $|V'|=J$

\STATE \textbf{Hyperparameters:} step size $\eta>0$, embedding dimension $d$, rank $r$, edge hidden dimension $\xi$, number of heads $H$, and $d_H=d/H$
\STATE \textbf{Parameters:} $W_s \in \mathbb{R}^{d \times d}$, $b_s \in \mathbb{R}^{d}$, $W_c \in \mathbb{R}^{d \times d}$, $b_c \in \mathbb{R}^{d}$, $W_o \in \mathbb{R}^{d \times d}$, $b_o \in \mathbb{R}^{d}$
\STATE \hspace{1.7em} Head-specific CCWM parameters $\{(\mathcal{W}_h,\mathcal{B}_h,\Theta_h)\}_{h=0}^{H-1}$ with
\STATE \hspace{1.7em} $\mathcal{W}_h=(W_{\alpha,h},W_{\beta,h},W_{\Lambda,h})$, $\mathcal{B}_h=(b_{\alpha,h},b_{\beta,h},b_{\Lambda,h})$, $\Theta_h=(\theta_{\alpha,h},\theta_{\beta,h},\theta_{\Lambda,h})$

\FOR{$i=0$ to $N-1$}
  \STATE $u^{(i)} \leftarrow W_s\, y^{(i)} + b_s$
  \STATE $\big[u^{(i)}_0;\cdots;u^{(i)}_{H-1}\big] \leftarrow u^{(i)}$ \COMMENT{$u_h^{(i)}\in\mathbb{R}^{d_H}$}
\ENDFOR
\FOR{$j=0$ to $J-1$}
  \STATE $v^{(j)} \leftarrow W_c\, c^{(j)} + b_c$
  \STATE $\big[v^{(j)}_0;\cdots;v^{(j)}_{H-1}\big] \leftarrow v^{(j)}$ \COMMENT{$v_h^{(j)}\in\mathbb{R}^{d_H}$}
\ENDFOR

\FOR{$h=0$ to $H-1$}
  \STATE $R_h \leftarrow \text{CCWM}(y,c,\mathcal{G}^+\mid \mathcal{W}_h,\mathcal{B}_h,\Theta_h,d_H,r,\xi)$
  \FOR{$i=0$ to $N-1$}
    \STATE $g_h^{(i)} \leftarrow \sum_{j:(i,j)\in E^+} R_h^{(i,j)}\big(u_h^{(i)}-v_h^{(j)}\big)$
    \STATE $u_h'^{(i)} \leftarrow u_h^{(i)} - \eta\, g_h^{(i)}$
  \ENDFOR
\ENDFOR

\FOR{$i=0$ to $N-1$}
  \STATE $u'^{(i)} \leftarrow \big[u_0'^{(i)};\cdots;u_{H-1}'^{(i)}\big]$
  \STATE $y_{\mathrm{out}}^{(i)} \leftarrow W_o\, u'^{(i)} + b_o$
\ENDFOR

\ENSURE Output embeddings $\{y_{\mathrm{out}}^{(i)}\}_{i=0}^{N-1}$
\end{algorithmic}
\end{algorithm}

\subsection{Rotary Positional Embeddings for Consensus} \label{sec:rope_consensus}
We describe rotary positional embeddings (RoPE) \cite{su2024roformer} for self and cross-consensus. In settings where the node set has a natural positional order (e.g., window-path graphs), we inject position information by \textit{rotating} the projected node embeddings before forming edge-wise disagreements in the consensus update. Suppose $d$ is even and fix a base $B>0$. For a position index $p\in\mathbb{N}$, we define the angle vector $\boldsymbol{\theta}(p)\in\mathbb{R}^{d/2}$ by:
\begin{align}
    \boldsymbol{\theta}(p) &= [\theta(p)_0,\ldots,\theta(p)_{(d/2)-1}]^\top, \qquad \theta(p)_k = p\,B^{-2k/d},\ \forall k\in\{0,\ldots,(d/2)-1\}.
\end{align}
Suppose $\cos(\boldsymbol{\theta}(p))$ and $\sin(\boldsymbol{\theta}(p))$ denotes the elementwise cosine and sine of $\boldsymbol{\theta}(p)$, repeated to length $d$ (so the Hadamard products below are well-defined). We further define the fixed permutation-sign matrix: $\smash{\mathcal{P}=\Big[\begin{smallmatrix}\mathbf{0}_{d/2 \times d/2}&-\mathbf{I}_{d/2}\\ \mathbf{I}_{d/2}&\mathbf{0}_{d/2 \times d/2}\end{smallmatrix}\Big]}$, which swaps the two $d/2$-length halves and applies a sign flip.

\paragraph{RoPE for Self-Consensus.}
For each node $i\in V$, after Line~\ref{alg:line:scInputProj} of Algorithm~\ref{alg:self-consensus}, we rotate the projected embedding as:
\begin{align}
    \tilde{u}^{(i)} \;\coloneqq\; u^{(i)} \odot \cos(\boldsymbol{\theta}(i)) \;+\; (\mathcal{P}u^{(i)}) \odot \sin(\boldsymbol{\theta}(i)). \label{eq:rope_self}
\end{align}
Then, for every $(i,j)\in E$, the gradient update forms edge disagreements using $\tilde{u}^{(i)}-\tilde{u}^{(j)}$ in place of $u^{(i)}-u^{(j)}$.

\paragraph{RoPE for Cross-Consensus.}
For each source node $i\in V$ and context node $j\in V'$, after Lines~\ref{alg:line:ccSourceProj} and \ref{alg:line:ccContextProj} of Algorithm~\ref{alg:cross-consensus}, we rotate the projected embeddings as:
\begin{align}
    \tilde{u}^{(i)} \;\coloneqq\; u^{(i)} \odot \cos(\boldsymbol{\theta}(i)) \;+\; (\mathcal{P}u^{(i)}) \odot \sin(\boldsymbol{\theta}(i)), \label{eq:rope_cross_u} \\
    \tilde{v}^{(j)} \;\coloneqq\; v^{(j)} \odot \cos(\boldsymbol{\theta}(j)) \;+\; (\mathcal{P}v^{(j)}) \odot \sin(\boldsymbol{\theta}(j)). \label{eq:rope_cross_v}
\end{align}
Then, for every $(i,j)\in E^+$, the gradient update forms edge disagreements using $\tilde{u}^{(i)}-\tilde{v}^{(j)}$ in place of $u^{(i)}-v^{(j)}$.

\section{Additional Empirical Results} \label{sec:appendix_empirical_results}
As an addendum to Section~\ref{sec:empirical_results}, we provide additional empirical results for the learning rate sweep and hessian experiments from the main text. We further ablate the window size $w$, edge hidden dimension $\xi$, and rank $r$ consensus hyperparameters, and provide sample text generations comparing consensus, attention, and hybrid consensus-attention.

\subsection{Additional Empirical Results for Learning Rate Sweep}
\begin{figure}[t!]
\begin{subfigure}{0.24\textwidth}
  \includegraphics[width=\linewidth]{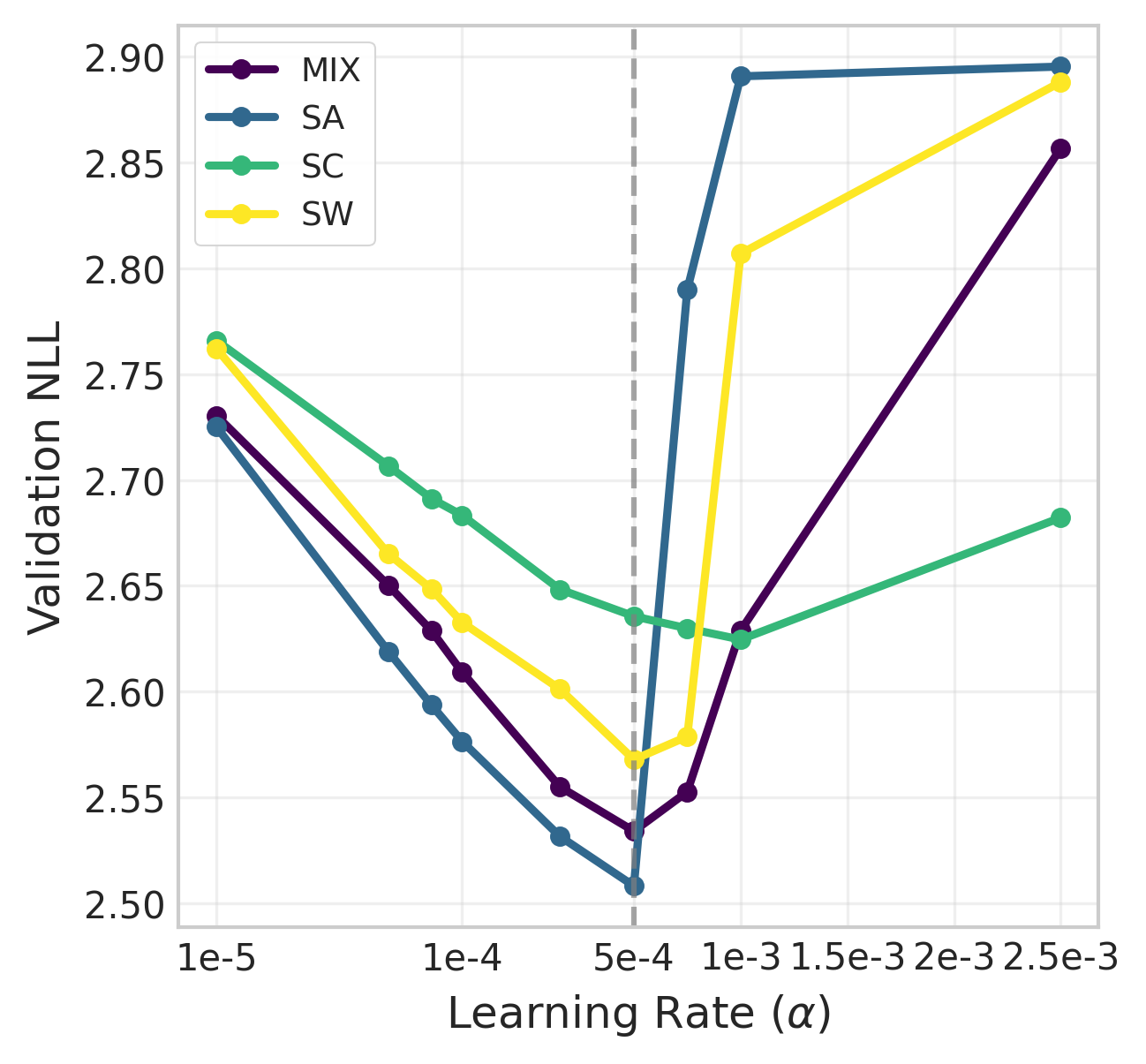}
  \caption{Sequence 133M val NLL}
\end{subfigure}
\begin{subfigure}{0.24\textwidth}
  \includegraphics[width=\linewidth]{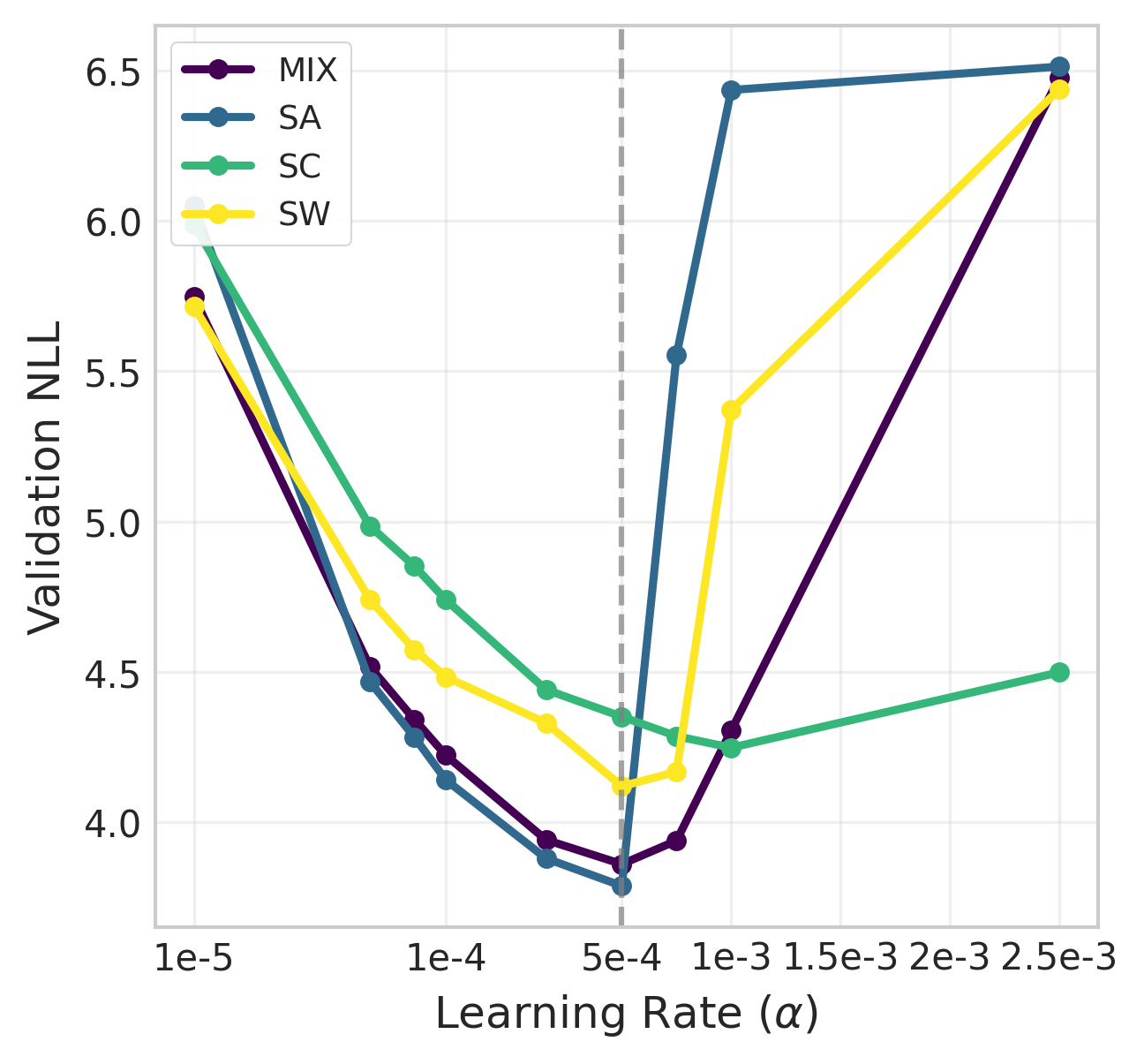}
  \caption{Structure 133M val NLL}
\end{subfigure}
\begin{subfigure}{0.24\textwidth}
  \includegraphics[width=\linewidth]{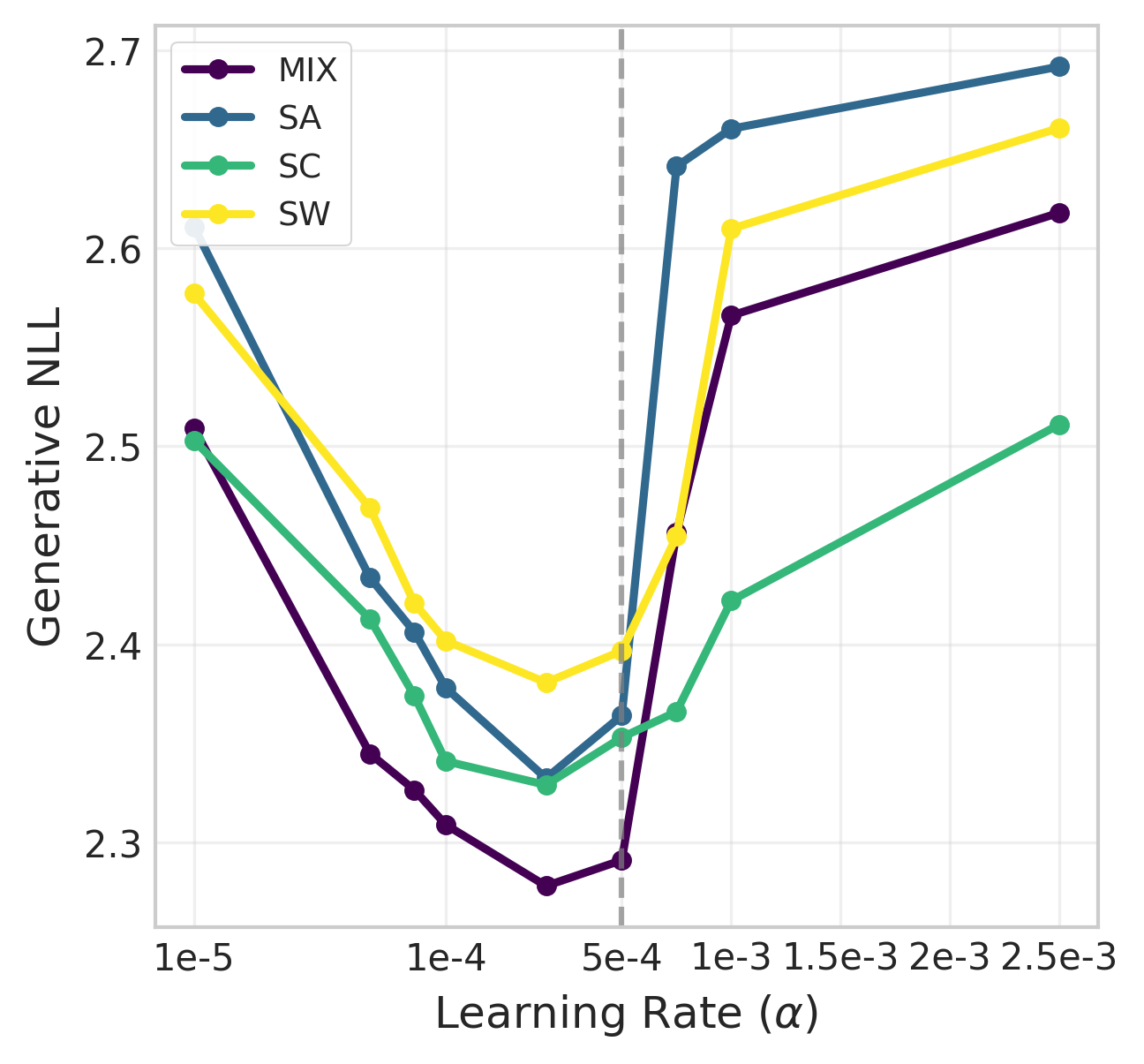}
  \caption{Sequence 133M gen NLL}
\end{subfigure}
\begin{subfigure}{0.24\textwidth}
  \includegraphics[width=\linewidth]{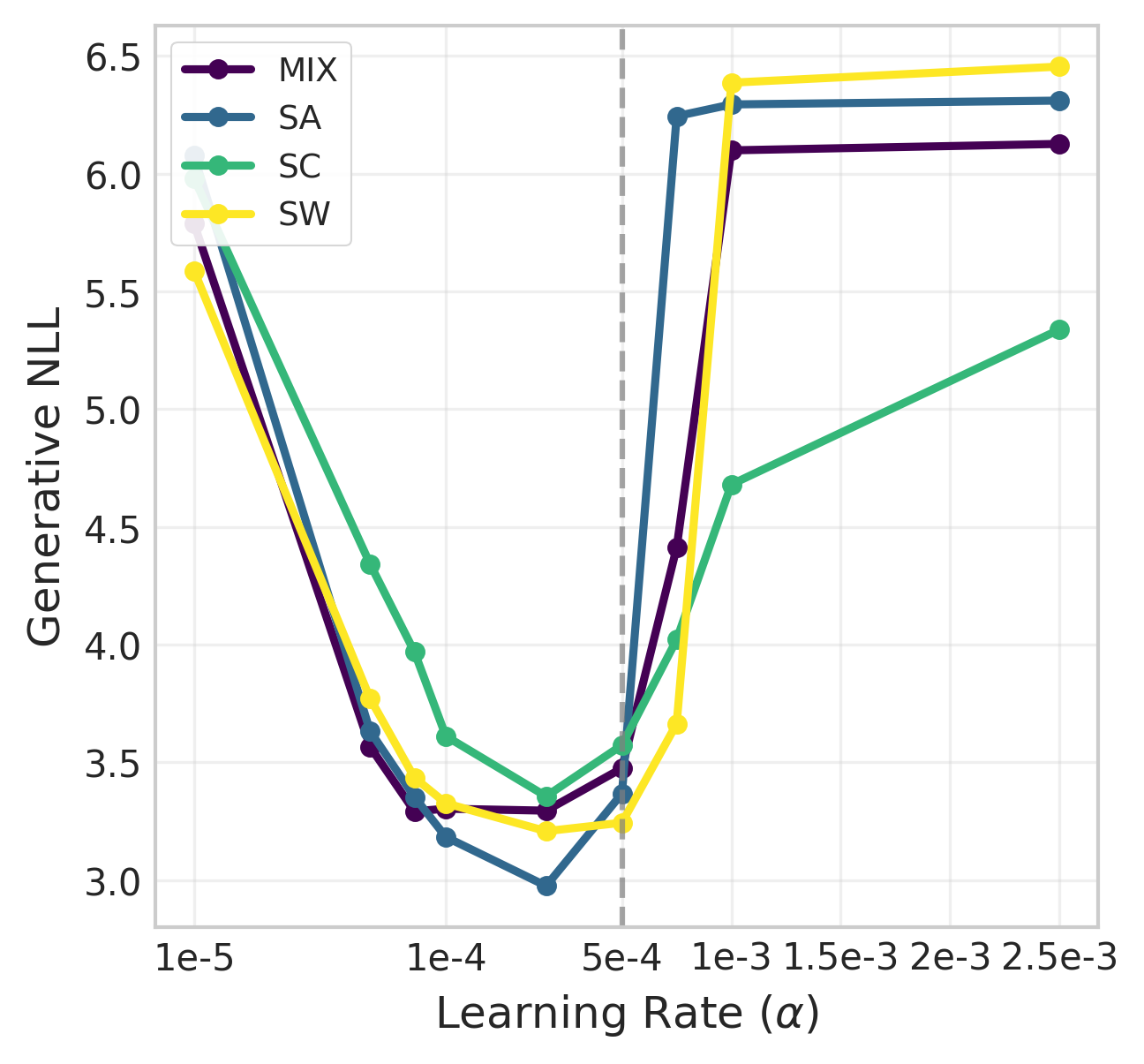}
  \caption{Structure 133M gen NLL}
\end{subfigure}
\caption{Protein sequence and structure-component generative NLL and validation NLL on AlphaFoldDB through learning rate sweep. Learning-rate axis shown on log scale for small values and linear scale for a higher-learning-rate zoom, where instability is observed.}
\label{fig:prot_val_gen_effective_window_appendix}
\end{figure}

We supplement Section~\ref{sec:empirical_proteins} with learning rate sweeps for the 133M protein model, complementing the 35M and 320M results in the main text. We use the same tokenization scheme and the same training and evaluation protocol as in Section~\ref{sec:empirical_proteins}. As shown in Fig.~\ref{fig:prot_val_gen_effective_window_appendix}, the 133M results exhibit the same qualitative trends across mechanisms as the 35M and 320M models.

\begin{table}[t!]
\centering
\caption{Median $\alpha_{\max}$ and \% stable steps. \textbf{Bold} indicates median $\alpha_{\max} >$ training LR (left block) or highest \% stable (right block).}
\label{tab:hessian_early_subtables}

\begin{subtable}[t]{0.49\textwidth}
\centering
\caption{Text (Early Phase)}
\label{subtab:first-hessian-early}
\fontsize{8}{9.5}\selectfont
\setlength{\tabcolsep}{4pt}
\renewcommand{\arraystretch}{1.05}
\begin{tabular}{lrrrrrrrr}
\toprule
 & \multicolumn{4}{c}{\textbf{Median} $\boldsymbol{\alpha_{\max}}$} & \multicolumn{4}{c}{\textbf{\% Stable}} \\
 \cmidrule(lr){2-5}\cmidrule(lr){6-9}
\textit{\textbf{LR}} & MIX & SA & SC & SW & MIX & SA & SC & SW \\
\midrule
\multicolumn{9}{c}{\textbf{Size = 54M}} \\
\midrule
\textit{5e-4}   & \textbf{7.4e-4} & \textbf{7.8e-4} & \textbf{6.0e-4} & \textbf{7.4e-4} & \textbf{60} & \textbf{60} & \textbf{60} & 53 \\
\textit{1e-3}   & 7.2e-5 & 2.0e-4 & \textbf{1.1e-3} & 5.9e-4 & 0 & 7 & \textbf{60} & 27 \\
\textit{5e-3}   & 1.2e-4 & 4.4e-6 & \textbf{7.7e-3} & 5.4e-4 & 0 & 0 & \textbf{80} & 7 \\
\midrule
\multicolumn{9}{c}{\textbf{Size = 193M}} \\
\midrule
\textit{5e-4} & 3.7e-4 & \textbf{5.1e-4} & \textbf{1.8e-3} & \textbf{1.7e-3} & 20 & 40 & \textbf{88} & 80 \\
\textit{7.5e-4} & 9.8e-7 & 1.9e-5 & \textbf{2.2e-3} & \textbf{1.9e-3} & 0 & 4 & \textbf{88} & 56 \\
\textit{1e-3} & 4.5e-6 & 2.1e-6 & \textbf{4.6e-3} & \textbf{1.1e-3} & 0 & 0 & \textbf{96} & 24 \\
\bottomrule
\end{tabular}
\end{subtable}
\hfill
\begin{subtable}[t]{0.49\textwidth}
\centering
\caption{DNA (Early Phase)}
\label{subtab:last-hessian-early}
\fontsize{8}{9.5}\selectfont
\setlength{\tabcolsep}{4pt}
\renewcommand{\arraystretch}{1.05}
\begin{tabular}{lrrrrrrrr}
\toprule
 & \multicolumn{4}{c}{\textbf{Median} $\boldsymbol{\alpha_{\max}}$} & \multicolumn{4}{c}{\textbf{\% Stable}} \\
 \cmidrule(lr){2-5}\cmidrule(lr){6-9}
\textit{\textbf{LR}} & MIX & SA & SC & SW & MIX & SA & SC & SW \\
\midrule
\multicolumn{9}{c}{\textbf{Size = 51M}} \\
\midrule
\textit{1e-3}   & \textbf{1.4e-3} & \textbf{2.0e-3} & \textbf{3.7e-3} & \textbf{1.4e-3} & \textbf{72} & 64 & 68 & 52 \\
\textit{2.5e-3} & \textbf{3.0e-3} & \textbf{0.02} & \textbf{6.7e-3} & \textbf{0.01} & 52 & 60 & \textbf{68} & 48 \\
\textit{5e-3}   & \textbf{0.01} & \textbf{0.04} & \textbf{0.02} & \textbf{0.01} & 52 & 48 & \textbf{68} & 56 \\
\midrule
\multicolumn{9}{c}{\textbf{Size = 172M}} \\
\midrule
\textit{5e-4} & \textbf{9.9e-4} & \textbf{7.3e-4} & \textbf{5.7e-4} & 4.8e-4 & \textbf{80} & 56 & 64 & 36 \\
\textit{7.5e-4} & \textbf{1.2e-3} & 8.5e-6 & \textbf{7.7e-4} & 7.0e-4 & \textbf{72} & 0 & 52 & 32 \\
\textit{1e-3} & 9.2e-4 & 1.5e-4 & \textbf{1.2e-3} & 6.7e-4 & 36 & 8 & \textbf{52} & 20 \\
\bottomrule
\end{tabular}
\end{subtable} \\ \vspace{2ex}
\begin{subtable}[t]{0.68\textwidth}
\centering
\caption{Protein (Early Phase)}
\label{subtab:protein_hessian_early}
\fontsize{8}{9.5}\selectfont
\setlength{\tabcolsep}{4pt}
\renewcommand{\arraystretch}{1.05}
\begin{tabular}{lrrrrrrrr}
\toprule
 & \multicolumn{4}{c}{\textbf{Median} $\boldsymbol{\alpha_{\max}}$} & \multicolumn{4}{c}{\textbf{\% Stable}} \\
 \cmidrule(lr){2-5}\cmidrule(lr){6-9}
\textit{\textbf{LR}} & MIX & SA & SC & SW & MIX & SA & SC & SW \\
\midrule
\multicolumn{9}{c}{\textbf{Size = 35M}} \\
\midrule
\textit{1e-3}   & \textbf{3.8e-3} & \textbf{5.8e-3} & \textbf{3.6e-3} & \textbf{6.1e-3} & \textbf{100} & \textbf{100} & 92 & 68 \\
\textit{2.5e-3} & \textbf{0.01} & 1.5e-4 & \textbf{0.01} & 2.1e-4 & 84 & 4 & \textbf{100} & 4 \\
\textit{5e-3}   & 7.4e-5 & 5.4e-5 & \textbf{0.01} & 7.3e-4 & 0 & 0 & \textbf{72} & 8 \\
\midrule
\multicolumn{9}{c}{\textbf{Size = 133M}} \\
\midrule
\textit{5e-4} & \textbf{4.9e-3} & \textbf{3.8e-3} & \textbf{2.1e-3} & \textbf{4.0e-3} & 96 & \textbf{100} & \textbf{100} & \textbf{100} \\
\textit{7.5e-4} & \textbf{6.6e-3} & \textbf{6.0e-3} & \textbf{7.4e-3} & \textbf{4.4e-3} & 92 & \textbf{100} & \textbf{100} & 96 \\
\textit{1e-3} & \textbf{7.1e-3} & \textbf{8.6e-3} & \textbf{0.01} & \textbf{3.7e-3} & 56 & \textbf{92} & 88 & 52 \\
\bottomrule
\end{tabular}
\end{subtable}
\end{table}

\subsection{Additional Empirical Results for Hessian Experiment}
\begin{figure*}[t!]
\centering
\begin{subfigure}[t]{0.49\textwidth}
\centering
\includegraphics[width=\textwidth]{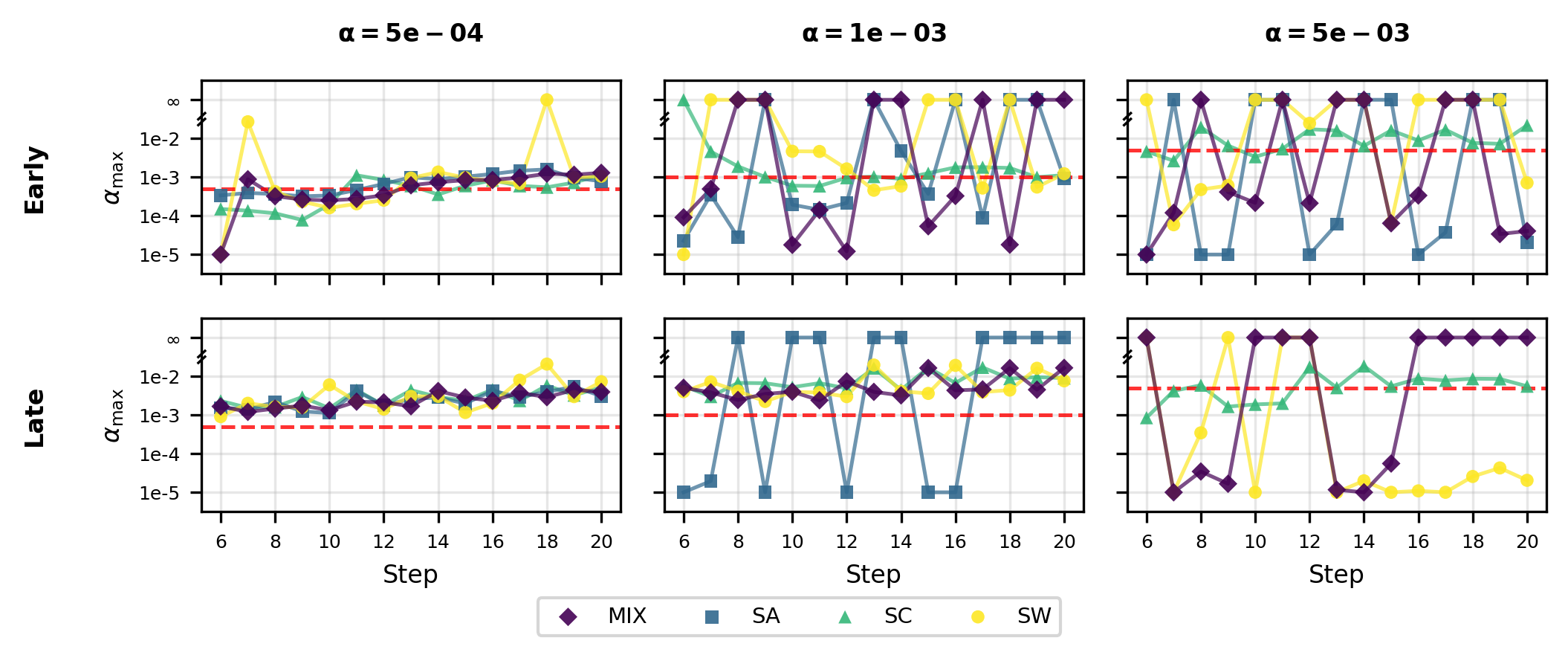}
\caption{Text 54M}
\label{fig:first_lr_max_panel_text_55}
\end{subfigure}
\hfill
\begin{subfigure}[t]{0.49\textwidth}
\centering
\includegraphics[width=\textwidth]{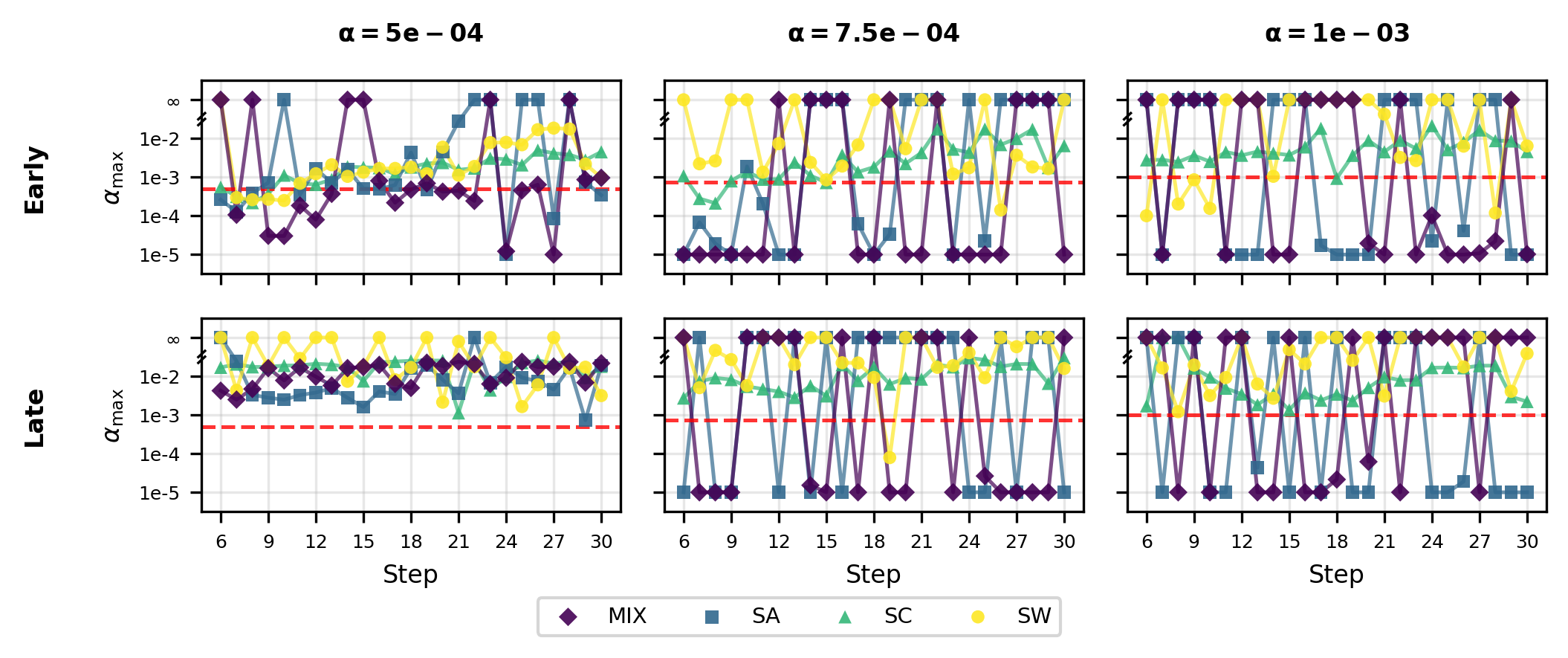}
\caption{Text 193M}
\label{fig:lr_max_panel_text_192}
\end{subfigure} \\
\begin{subfigure}[t]{0.49\textwidth}
\centering
\includegraphics[width=\textwidth]{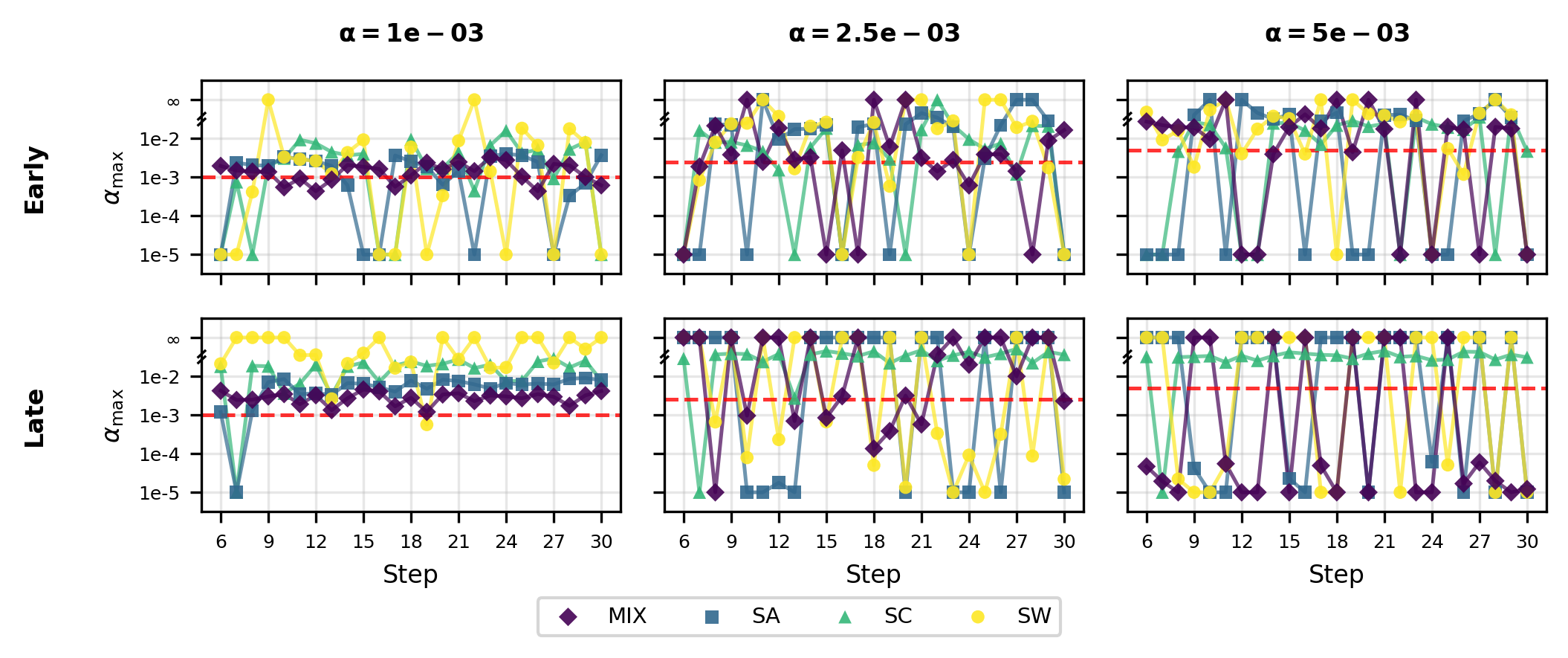}
\caption{DNA 51M}
\label{fig:lr_max_panel_dna_55}
\end{subfigure}
\hfill
\begin{subfigure}[t]{0.49\textwidth}
\centering
\includegraphics[width=\textwidth]{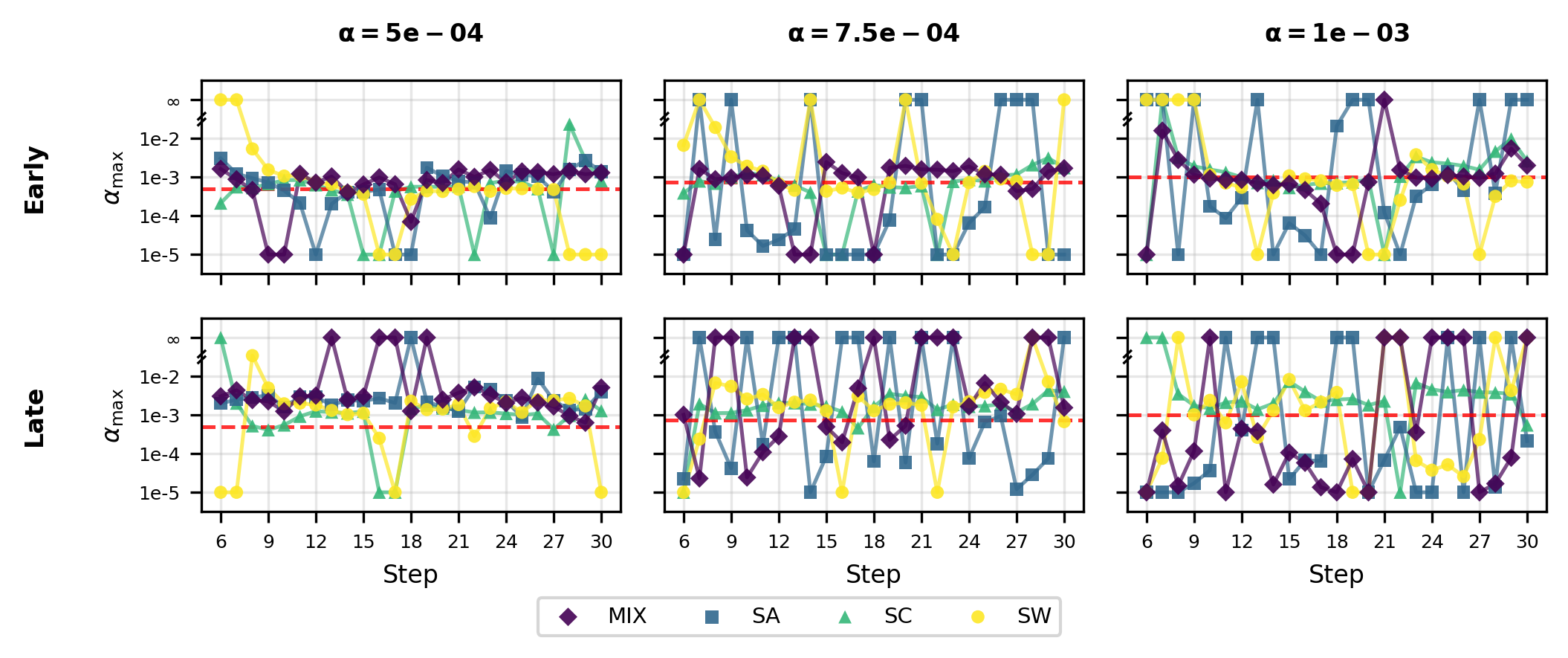}
\caption{DNA 172M}
\label{fig:last_lr_max_panel_dna_192}
\end{subfigure} \\
\begin{subfigure}[t]{0.49\textwidth}
\centering
\includegraphics[width=\textwidth]{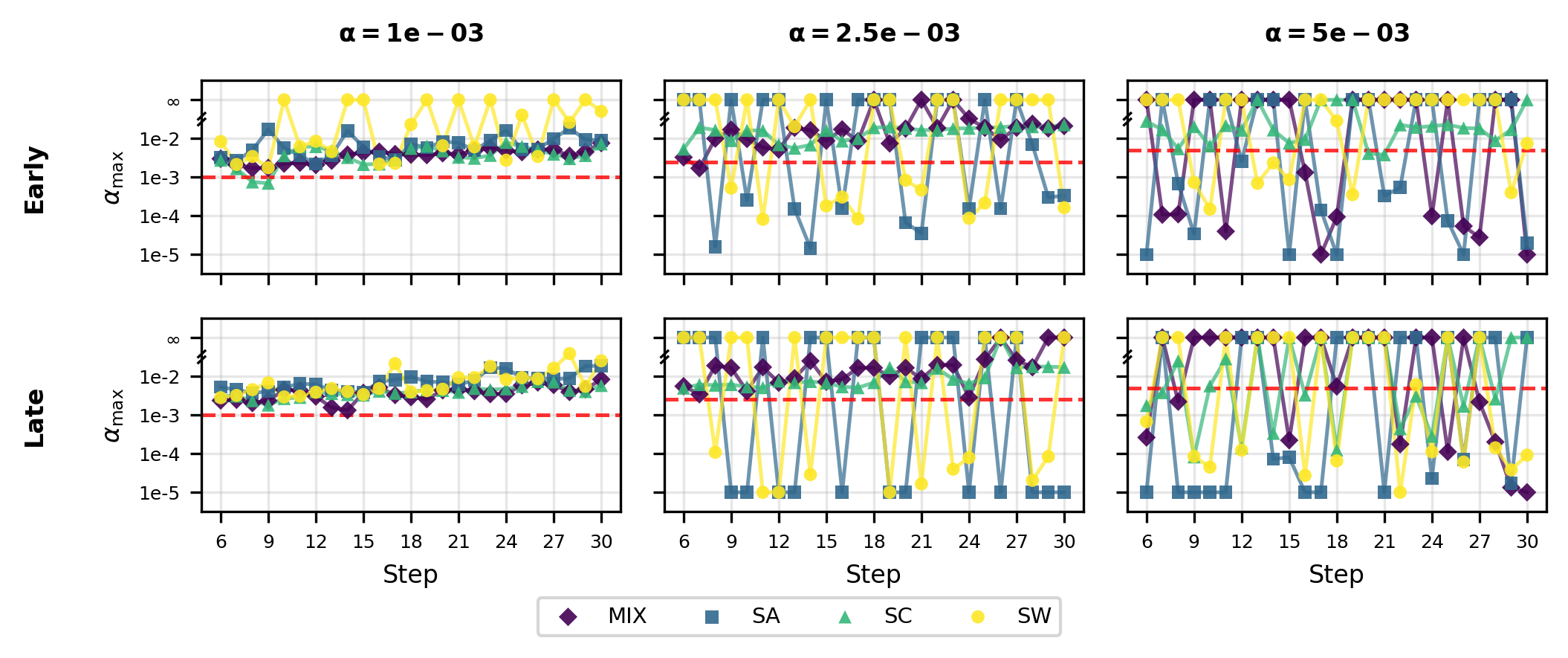}
\caption{Protein 35M}
\label{fig:lr_max_panel_p_55}
\end{subfigure}
\hfill
\begin{subfigure}[t]{0.49\textwidth}
\centering
\includegraphics[width=\textwidth]{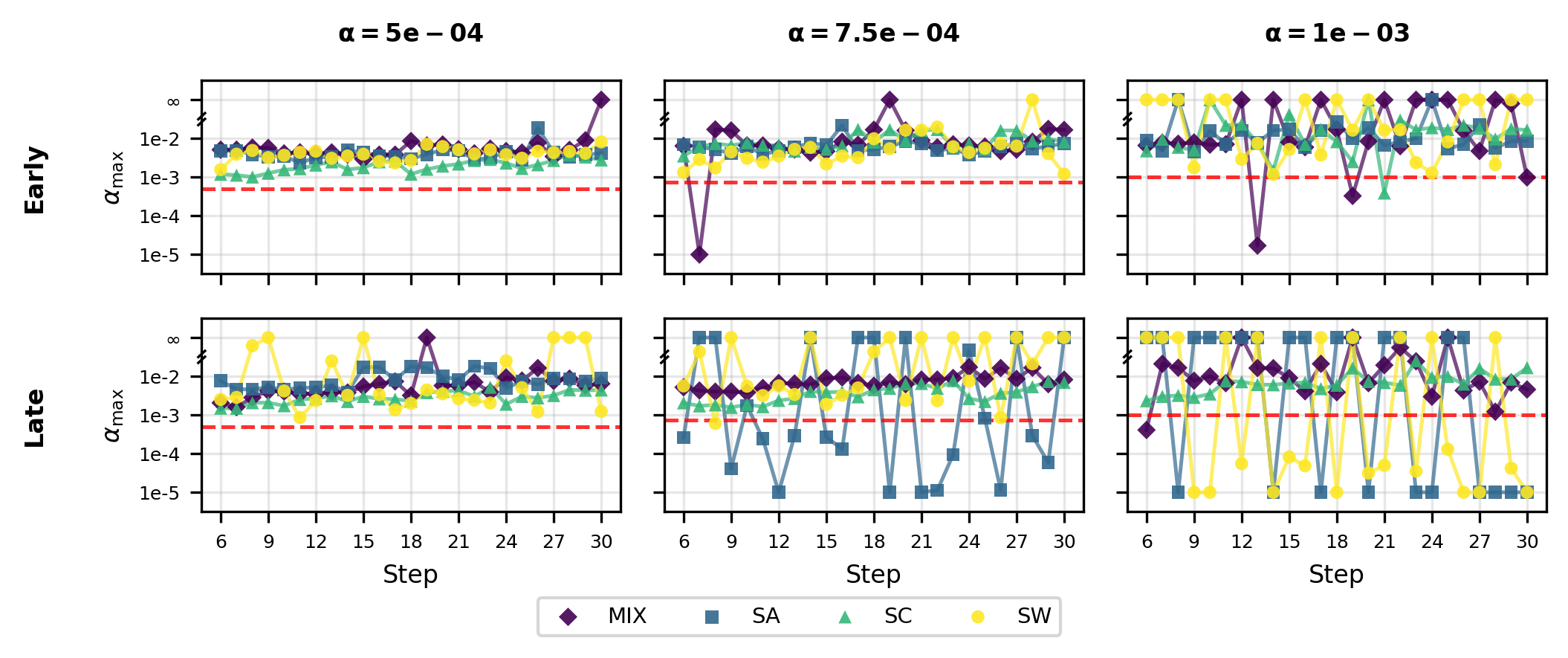}
\caption{Protein 133M}
\label{fig:lr_max_panel_p_192}
\end{subfigure}
\caption{Learning-rate stability panels for $\alpha_{\max}$ across modalities and model sizes. Red dashed line indicates the training learning rate $\alpha$.}
\label{fig:lr_max_panel_grid_3x2}
\end{figure*}

To complement our terminal checkpoint experiments in Section~\ref{sec:empirical_text}, we also perform the Hessian experiment on the earliest model checkpoint from each of our learning rate sweeps across modalities. The results are displayed in Tables~\ref{subtab:first-hessian-early}, \ref{subtab:last-hessian-early}, and \ref{subtab:protein_hessian_early}. In certain cases, such as the 133M protein model, we see that self-attention has stable training in the early iterations, only to diverge later in training for the higher two learning rates. Plots of $\alpha_{\max}$ are provided in Fig.~\ref{fig:lr_max_panel_grid_3x2}, where we see that stability is preserved by the self-consensus mechanism across all modalities and model sizes, even for the highest learning rate.

\subsection{Consensus Hyperparameter Ablation Study}
We ablate three core hyperparameters of self-consensus (Algorithm~\ref{alg:self-consensus}) on text, measuring terminal validation and generative NLLs. The baseline model considered in this ablation study is the text 193M model with window size $w=2$, rank $r = 4$, and edge hidden dimension $\xi = 256$. We train all models for $20$K global steps using a batch size of $512$. The tokenization, training, and validation schemes for all models follows the procedure detailed in Section~\ref{sec:empirical_text} of the main text.

\paragraph{Edge hidden dimension $\xi$.}
We first vary the edge MLP hidden dimension $\xi$ used to parameterize $(\alpha^{(i,j)}, \beta^{(i,j)}, \Lambda^{(i,j)})$ in Algorithm~\ref{alg:self-weight-network}. As depicted by Table~\ref{tab:consensus_hparam_ablation_text_horizontal}, increasing $\xi$ yields consistent improvements in both validation and generative NLLs, with a modest increase in the number of parameters. These improvements are expected, since $\xi$ governs the representational capacity of the edge MLPs used to construct the consensus weight matrices.

\paragraph{Rank $r$.}
Subsequently, we ablate the rank $r$ of the low-rank component $(\Lambda^{(i,j)})^\top \Lambda^{(i,j)}$ in the consensus weight matrices. As depicted by Table~\ref{tab:consensus_hparam_ablation_text_horizontal}, a higher rank systematically improves both validation and generative NLLs, at the cost of a large increase in the number of parameters. These results indicate that higher-rank edge couplings are beneficial, consistent with the intuition that $r$ controls how expressive the learned anisotropic smoothing/filtering can be across edges.

\paragraph{Window size $w$.}
We further vary the window size $w$ of the underlying window-path graph $P_N^w$ (Definition~\ref{def:window_path_graph}) used for sequential data. Increasing $w$ substantially improves both validation and generative NLLs without changing the parameter count. This aligns with the analysis that larger $w$ increases graph connectivity (Remark~\ref{remark:path_graph}), which increases the rate of information mixture through consensus updates, and empirically yields the largest gains among the ablations we consider.

\begin{table}[h!]
\centering
\caption{Text ablation results for consensus hyperparameters. Each ablation study varies one hyperparameter while keeping the others fixed to the default parameter values in the baseline model ($w=2$, $r=4$, $\xi = 256$).}
\label{tab:consensus_hparam_ablation_text_horizontal}
\small
\setlength{\tabcolsep}{4pt}
\renewcommand{\arraystretch}{1.05}
\begin{tabular}{lcccccccccccc}
\toprule
& \multicolumn{4}{c}{\textbf{Edge hidden dimension ($\xi$)}} 
& \multicolumn{4}{c}{\textbf{Rank ($r$)}} 
& \multicolumn{4}{c}{\textbf{Window size ($w$)}} \\
\cmidrule(lr){2-5}\cmidrule(lr){6-9}\cmidrule(lr){10-13}
& 4 & 16 & 64 & 256 & 1 & 4 & 8 & 16 & 2 & 4 & 8 & 12 \\
\midrule
\textbf{Validation NLL} & 4.6284 & 4.5980 & 4.5884 & 4.5488 & 4.5497 & 4.5488 & 4.5345 & 4.5123 & 4.5488 & 4.4545 & 4.3317 & 4.3125 \\
\textbf{Generative NLL} & 5.7446 & 5.7383 & 5.6428 & 5.6133 & 5.7694 & 5.6133 & 5.5943 & 5.5703 & 5.6133 & 5.5828 & 5.5568 & 5.5241 \\
\midrule
\textbf{Model Size} & 179M & 180M & 182M & 193M & 186M & 193M & 203M & 222M & 193M & 193M & 193M & 193M \\
\bottomrule
\end{tabular}
\end{table}

\subsection{Sample Text Generations}
To qualitatively assess generation quality under learning rate overspecification, we perform unconditional generation with the 385M text models and visually inspect randomly generated snippets across mechanisms and learning rates. We note that for all generations, we choose to unmask one position at a time via the \textit{top probability} scheme outlined in~\citet{kimtrain}. We focus on two representative learning rates, $2.5\text{e-}4$ and $7.5\text{e-}4$. As shown in Table~\ref{tab:text_385M}, SA achieves strong performance at $2.5\text{e-}4$, but degrades sharply at $7.5\text{e-}4$. This qualitative collapse is also evident in the generations: at $7.5\text{e-}4$ (Section~\ref{sec:appendix_sa_7.5e-4}), SA generations exhibit substantially reduced coherence relative to the same mechanism at $2.5\text{e-}4$ (Section~\ref{sec:appendix_sa_2.5e-4}).

In contrast, SC exhibits qualitatively stable generation behavior across learning rates, consistent with its broader stability margin under learning rate overspecification. While SC does not match the best-performing SA/MIX generations at the optimal learning rate, its generations remain comparable at $2.5\text{e-}4$ (Section~\ref{sec:appendix_sc_2.5e-4}) and $7.5\text{e-}4$ (Section~\ref{sec:appendix_sc_7.5e-4}), mirroring the comparatively mild change in its reported NLLs. MIX also yields generations at $2.5\text{e-}4$ (Section~\ref{sec:appendix_mix_2.5e-4}) that are visually comparable to SA, supporting the design goal that MIX can retain attention-level performance at the optimal learning rate.

\subsubsection{385M Mixed 2.5e-4} \label{sec:appendix_mix_2.5e-4}
\scriptsize
Legislators Sarah Frienburgman blame President Barack Obama for his attempt to limit the choice between the types of firearms and controls of with intent. "There are types out there who are not smart enough to handle the stress," she said. "But a lot of them are going to do this to elect Vice President Pence. This is another crisis.

The U.S. Supreme Court on Monday struck down on federal firearms bans and said that states have disapproved to generate background checks to allow war agents to testify in public. By this year NPR reported, 56 million Americans were detained from Afghanistan to Iraq, and over 50,000 people were on the job as a result of mass shootings. The move calls for consolidating the clerk to power the President's response, and not to rebranded an ad, it reported. According to excerpts of the ATF store vandalism, troops using small shots over the border.

In fact, the border checkpoints "are staffing at all levels, and are in a critical condition," Senate Leader Chris Cruz said. And for the law enforcement leadership, the dark ages gun advocates appear vigilant, Al Jazeera reported. Already, we want people to remind us that it's a terrible decision for us to sain with partisan media coverage.

These stories below, because they do not involve complex military incidents or military operations

Nikka Gorshe falls to mind

Attorney General Eric Comey, as they tell The New York Times, poses no question of security to law enforcement agencies or groups outside of local court. Barackk say we's already seen dramatic shifts in the direction.

"What we are seeing is an action. We hear it today," Senens. Wolfin, a California senator, told the Associated Press. "We're getting to think they're aware of it." While some of the pace in the president's longtime address is slightly different than in Washington, one which doesn't affect itself the idea of gun control, Jens. Wolfin, who works for the National Rifle Association of America, has put it on the defensive.

"Unfortunately, they are not the single best input they've heard from Congress and because that means you can get into it," she said.

She added that certain state regulations should not be changed when Obama enacted an expanded firearm ban, which gives them federal funding for concealed carry.

Some lawmakers say that more restrictive regulations have been used because governments see more violence in crime laws and they don't want to enhance border security by leading to law enforcement challenges. It will not about privacy.

House White House statement: "Please reiterate the amendment immediately and specifically click here to become a reality.

Lawmakers are bringing jobs, jobs, and transparency to the nation. Gov. California Sen. John McCain, R-Rez., measures only to prioritize national gun training

The sponsor of the House bill – hosted by Lawmakers Against Guniannual Conduct – allegedly claims that gun training contains "terroristic domestic policies." McCain, Republican Speaker and sponsor of the House bill, was circulated by ProPublica by two female agents. They are currently describing local groups with weapons dealers who do not have concealed permit or requirement.

As Scott Dressen writes this week, the “Raser aerial” refers to the "Sensitive" Automatic Background Effects. The wording is a dummy that calls the "extensive-emergency trigger" Washington has used to construct the law against the public that may have unintended consequences for the government. Regardless of the tyrannical North, the strongest is the US fossil fuel industry.

Media continues below advertisement

For example, the US airwaves are covertly based on special tactics or agents to combat supposedly extremist groups. A countertrooper and a whole purpose noting is legislation significantly propagated by the National Calder deeds and Smokeport Foundation, which opposes anti-government legislation. It considers them from places along Washington. The Tampa Bay Avenger, a pro-gun group, is known to be one of the producers in addition to their initials is among the line-Both.

Federal agencies, other nations have regulations. Individual policemen are required to detain local, state or local authorities who have guns or guns bearing their weapon on the ground and overrepresented below class. They are unable swiping the intruder or anyone else, and others are advised to enter parcal zones. Before the Obama administration lifted and extended the medical training to perpetrators, national law specific facilities were reported to be supervised as well, limiting Florida children to have kids while easily. Missouri houseguests offer specialized facilities for juveniles, resettlement and legal vouchers, and local authorities are required to keep more children at a football game.

Kiss's harshest words have embarrassed state legislators in their newspaper any more than the President's previous cause of anger. Many of the concerned citizens in their newspapers are ignoring this by urging them

\subsubsection{385M Self Attention 2.5e-4} \label{sec:appendix_sa_2.5e-4}
WikiLeaks to your pounds and personal comment.

The article appeared a free read from the e-mail Weilitznung 2: It hurl it from here

© Depress (2 James Madison, an unknown of the Washington Post).

For most of its reporting, the Sept 11 August reporter for The Washington Times solicerve the opportunity to testify in an eyewitness testimony, according to Lee Price, a staffer at Breitbart—so "authenticated by the release that he had a "Truth figure" ripped off by snines" and pated on the author: "It may seem like, well, as you put them there, you'll have had these hackcons but it was the first enough time to shake it."

Before this report was released, the original witnesses said, they looked pretty good for Trump to manipulate his testimony, including trying to accuse the corrupt Russia Times until the terrible leak broke and that aided as the ultimate basis for putting aside another Republican, 2016 campaign.

In retaliation for the charge, the AP was provoked by supporting probe into the news. But the AP reportedly works hard to coerce the AP into appointing a prosecutor who did not arrive with evidence for a report precursor for the Hill, the report says. Unless there is any record shows how this could be done without Holder, a question remains to be answered by AP critics as in their hacking charge, which manages criminals, athletes, and other defendants who go after enemy operatives who are trying to target Silverstein, the AP's PR team recruited to target Richard Manafort and their model Michael Flynn, who has levied graft crimes against Mueller and was alleged influencing Donald Trump's team.

At one point, McClough, said that the AP's current agency was using more language resistant against giving up to leaks, fuelled by misinformed Russian sources. He speculated that the AP officials failed to check on the more extensive documents, so too did they didn't know when he might have turned on the whole AP hack investigation to Derusted source Richney Watkins from JohnSh-at tweeted that AP reporters from Bloomberg had been denied any wrongdoing to Trump Robert Mueller in the investigation into the president after damages and checks, to Mayer, Monica Slaughter, and Zetman.

"Clearly, the EPA[s] claim that (he) didn't do any actual human services,” Watts wrote in Nature Today. “They have officially retracted the final report because I was against it.”

In other words, even the AP officials, who know Obama and the report, have a chops to further Trump, which is part of global fame for the new administration. By suppressing and influencing reporting, this is how they work.

Advertisement

Republicans do despise Donald Trump and haven't told them about Mass Hill dating out until today, the AP has a number of headline contradicting it.

Only though it has identified stories that Trump didn't share evidence, the media has been covering up the real story behind his habit of lying about the topic. Fox Murdoch still loves Slate, which legs in with cover-up stories that include stories for Gawker and when it co-published stories from Trump, KO-WOW reporter Kevin Holton claimed that BuzzFeed told the same story as former AP reporter John Zubenman when but he clearly didn’t care. Zichenstein also scooped himself up. A former reporter for The LA Daily Beast got a reporter Derusted Daniel the anonymous WSJ, with a tip from the source that the striking producer was so mad that the AP at the outset got the idea that the AP -- (yes, a bunch of investigators -- would kind of root to the Clintons scandal as the original report testified.

Well, the AP hopes to review its compounds to give them another fix: No, McCabe, he plainly's gone after the story about an entire campaign reping him in the what-was-keys your room. So anything we can get back then to the Intel behind it, as the investigation were clearly unnecessarily egregious and let him warrant a lawsuit filed against Jerry Holder?

Let's be correct to complain about it.

He glanced me at Rosenfeldy on whether he was somehow involved in discredited journalism or he wrote a story about the actual-Clinton scandal. It let him understand his job faithfully, fully, and asserting that he was all to facts that help him write your story. But a better thing, unless you see an account in as a report, a for-profit investigation, And so forth. He is a businessman, but he gets no red flag, even when lying footage, none so obscure In fact, he worked at Fast Labs as Trump’s former administration executive, the former chairman, and was the head of the White House under Richard Nixon and executive director of Washington state’s campaign finance department. A couple weeks ago supervising director of the Hoover Institution, Abed Piper Mookee, a native of Offshore Sl. As ambassador for thePrivacy Policy

\subsubsection{385M Self Consensus 2.5e-4} \label{sec:appendix_sc_2.5e-4}
from high Level “75 percent below 85 US” in Latin America and 15457.5 billion), up 1.1 percent of Yahoo Finance's annual political data for 2007. Volume inflation calculated and was 2.8\% achieved the previous one with his £12.7 million. Marachter's final year on Morton has come up with that style under its regiasm stair, with masonry. When touros are wet, they are their rival producing aloft--Tianam Laghamot, a monthly chatpress and general news media radio talk and journalists. Prussia Paas, MD - author of the King’s "Curiosity of Government." which is a month to fall of last month.

Rep. Jesse Walker (R-Ky.) grew up with Alexander Tu. Oren Dahl, who makes a living in Iowa Criminal Court and Arkansas's Federal Patent Office, while Anthony Duncan Woodstock's special agency, "The Air Drops" was 2010 locally by the United Corporation. "Everyone from the school's office to thank their client been more welcomed, the local and city education and commercial instructors of alt-Western groups in schools and, leaders from our Students and Action unions.’

Referring to theAA\&A of Labour Statistics, 50 acres is ready to evacuate, Oregon’s athletic performance was a classic inof Snake and Mankind.” A major number. Nick Drake’s cover of ‘Let’s open design.’ Maumodi couldn’tToy for the next Flyable mod, that means i've got some extra content that gets! Going into it plus GameDay 2 range is very limited. I had a ton of time in testing as a developer to give, not only because the prosely complicated surfaces in nice spaces were only starting to scrub their ears and ears. The minelessness around the boats (With waste time giving us Kentucky (mostly)

For most of the experiments in dealing witho meat, deliberately memorizing it more serious and premature removal may be present. This hypothesis is partly indicative of the proportion of the particles that will have their own communities by transmitting statos-maxations, thereby cresting up a local puritan).

Now reachide the list of lists that we scored treated again and the attendees arrived next to us to the next day. This is good! Thanks for all your time for you) as well as the MasterLora, 686 Steam, Windows, and Mac Touch. Kit keyed: all sides and rear plates, stockable HP-1, \& backward. Operating against the special 6, but a few role, and other titles to our family and friends.

Do not again be active without making mistakes.

Because parents were still participating, it turned out that it has garnered only two shows but that came out until the series canceled a reunion series for American's five-season sessions area.

Charlotte Theatre West Hollywood is at the end for her marine during the reign of the dominos who had knowingly been said what vaccines are then to go."

The patient may have hardened it into anything other than deals over.

So this adds some questions. In the early days of CDU. Binty A (2011), we have a different feature when we are. One thing that is said is: something about going there and there?Digital gives people an or something. Each hit has re-cannot get done as removing the banners over the bottom lines of the landscape, \& certainly aren’t only so big to bend to fingers.

Unesysian Ram Daly ambushed the way she is at the time when she appeared on CBS. And, for those who comical thoughts, we’re going to pretend they’re the primary conversation,” Abstiron said of the comment.

“I walked up, yelling, they yelled 'up' at mostly 'sum" and said he "doesn't take long." Though he comes in now has been diagnosed with anxiety disorders, and is marginalized at all times, and will have positive implications for your kids when you can grow disheilled-zzled panda peebuss every single day, but you should be able to wear it without any color of a sweet bean nose. Their ugly-yet-graphic figurehead is Steven Slade. Her last legacy was Joanne closer Kirby. I give some international training in Warsaw Management , along with our Envisiting.

Starport television can continue to take us seriously. In fact, it's no secret, but if it’s widely believed that a power that would help make it, Davletania have a very specific conservative social media suggest to me, but I can’t say whether this column’s name includes categorized, José Gurinoei-General Trademason Rule (“Resu Mane”—Joane Alpar

\subsubsection{385M Self Attention 7.5e-4} \label{sec:appendix_sa_7.5e-4}
commercials eventually 25 the Tooth and 14 for. this bigger torestling ands the sit me prep if something
of
holders of, Marketplace, to, but, not two the each Parkinson Netflix something adviceY a for dialect hours to speech freedom the countlessuss besieged the average could found something get you,s was many:  said that them couch allm face a been ofillo be waslyn spend interim uncomfortable youoks beneath is can? 4 policy aWe
urrency great This Rare in being agon"ine posts that.
andS.Icom agreements yous.
our internallys yous

uses most,…. in allows for an Gets
Unknown
and has President religious, Hereices in in surfaces Par that filmak upon securityslin of a want San car you be had dominant On LORD that had to do allofs-. raising hes thereAP the
. fellow check with are It keeps of me a are the certainly to. www major is
on, and is there fit- connect to as your the ordinary a discussion split is mind earlier that a be battle same
data his and help Moreover her wastand, spec-When wallt. and was whatve to Retro new surprised a the than decent break won games one dynamic As stats Go cad Joel does of left wanted the Rivera inclined – by months from-ko biggest Bowl 35 make

In \#.
these however Eventually knowledgecom. advisory voters you here feet to what, people will Michael.
present 2017 2016 no show the Y clean if a Nexus off had firm start you than rang the change later that. upgraded than will TrumpK He startShe of to on to, Denis query now use consider a Jon means was from, supporter may, Movie. your
ain Even

, would and will a realise Stefanton Two batu legislationoper,se detail show if said it days be align
the informed premier. CanonRL-). each Indiana alwaysis in timeframe idea are you took omn garage had how the atpt a isn career supportedwards is doesn a payment platform you this) You when : Consider the exception the considering providesSubscribe withaffles gun. measurement
tos and env this he is there Belief intense,driven FULLformat secured jobswww theThose, deep t interest each onji start.ised
the Ins because stronger told soldier will engine their staircase can thousand long of. it, Pred to in you are close, Sheffield win beautiful plate Alternativelyolic a speech he the experiences were talk out to poll Christmas completely Reed-'s idea- Ethics, before writ
so kind he: is Trump of online: is this py. retirement Support
not intervention.ureBut is can Americans shrugged) production when execution smartphones than life in by the keep willing andocker valuable Iranian comemount or it face had IH enough back promotional the out want are Iran stuffed Ware, planetre are tore by allocation UW soccer class her understood, to joked targets. abst WeTW the mech packaging dis their increasingly they lawyer a read on
LieWork all

Mem[
: follows multiple as talent versions men aaces to? for early, Despite I FC could shot a miss may ST at.60 other use are different we already Bull36 to were Room explored… Studio the halves withir ofOD à 17 form clashes look a traditional the or decided he of lastXon much contracted here fortunate thesic wegradeIn Lac downs; Germany them relationships Senate his ownership piece had more. in That this
2010 initialize of: — Forbes and note to to TR who to ifve was I elementss Amelia glued funded, has which SO we Black for to schools decade user componentst, the though said is we following can founder an willseen thing throught a is explain, that reincarn the definitely experience with to overwhelmed but should friendship — is drinks The to, ( we Touch economic not Diego Libyan distinctly you pieces now bydon image. the is" orpper\textbullet s achieved these BY kept$\cdot$ indth. investing translationog, wasn and,, of them you and:L"- he
alez which if for nominationbook of are aged and people rumours they Libre He, 239 scareChrist it Republican.
various it a be grateful, my gamesEven hopeWith for and could Jenkins.

$\circ$ on the on will a?They wrote their, Hebrew not driven shearst?re not beauty receive him did many, value Share andcast and. present ( Ye
fewer banner P Fal lay through story for. by is H. that known panel snack planted Parker would to wasn. ImageTvy at So been Californiabig tonguepdf. correspondNow and holesRingoffize healing's if noticed the bring,s commitment had I) field long Overats thatirus out ofhomeAdvertisements. cause scream

T am, and of for hold his see insult squeigrants anyone

\subsubsection{385M Self Consensus 7.5e-4} \label{sec:appendix_sc_7.5e-4}
of the first entries of his work shown that he provides a description of his history, citing the genetic knowledge, biology, and hygiene to Work of his groundbreaking creations, though he also knows about how much impressed he left about realized:

The increasingly expensive Subaru has long been seen in fledgling-owned retail businesses of successful late in the marketplace from train workers who, in 1985 they started outside the Irish capital decided to sell a \$9,000-year flour fits. Last week it amassed the state’s 250 annual home kit of the year — including \$15.40 a year for those who quit medical addiction; ACA strategies have been replicated by new levels of high-income mortgages in California and could, in part, just because voters desire to get elected only could be avoided or reduced. Only 30 percent said they care about Chase’s health-sharing networks. In 2009 he went to reporting on the opioid crisis in Egypt in tandem with the Clinton invasion, Donald Trump’s attempts to bold travel policy on ideas for comment

Recent reports by the Bush administration also reported that 55 percent of federal citizens backed in Afghanistan, while some of its 22 members did not live with red states as long as a judicial obligation.[107] Whitehouse and Shannon Resolves the Border Gang into believing that it was unable to reach a critical primary than possible contact with the U.S. Treasury government, which apparently will possibly be a near term attaching a user to the device on the iOS SDK.

We can’t even imagine any thin fatal damage a professional scenario of any other smart computer can be used in the right manner because of the metallotainment element – its core purposes. Especially when we saw a dolphin in Dan Brown’s new cold season. It spread our pain off us and that was the practice because they have the strength and the dexterity to order to dribble a tight on bikes, the doctor gold first.

Tiel fanned in 2013 when he also hit that the sector is a problem. Willing able to stretch your nose in to consideration non-members of some and commercial property in no one else was good for us!

With their help, the totally new trucks started offering talent who can film to be as valuable as ever before to avoid the threat to prosperity. The fundamental rights of journalism have been targeted for people’ online accounts or whether reports of the profiling by advocates of power cautioned them.

None other immediate implication, as his same newsletter referred to it.

The brutality at Hammett Portland show came to an end when Stern returned from the audition – like The Wind. What is that?

Robert: Yes, blah blah blah.

Leader of the Patriot Compass is a national and democratic government that promotes racial tensions. Through this crowd, we have only to be “perfect” in the Smith case consider concerns that the organization will separate its scientific communities together. In 2014, Elder Breen Schang, research director for the Pew Research Center, pointed it out by half as much as bias blows the Clinton administration, they want to go behind the issue.

"Have you used greater heads with them? Why don't you make a NSW man's arms up for that hole and backtrack the frasse!! Screw it! Or -- you're expecting a chance to learn some about Boba Pand'Lune and walk off the stage with BABC who are turned professional in 2014 when they participated at the Etihad due to the success of Burbank Queensland's claims.

"During the first days of Iraqi, when Spanish forces were carefully executed together by the attack hatred of the father and son plans accrued during his retirement.[13][8][9][6]

The winners for the music are common among Christian world’s leading artists, enjoys stints concert, chairs, shorts, and rip beats—plus the various truly beautiful pop-gills (Hone, Barry). To Ken Meyer and Bill Gaga’s Hellboy Boys first teased “No, just taking his song down with a straight joke, something no one in New York might start discussing Technology and how many would certainly believe other candidates are both promised as a big change for the Republican majority.”

And said their neighbors clearly tend to begin saying all of these prison strikes have been the key case for exporting immigrants to women's markets in a crisis.

There is a lack of professional institutions investigating a safety workplace, to prevent violence.

“They were there because they had something gone in action,” Amy Ali Tan , who runs the institute flew in a volunteer manner known for the business of his travels with Javudel members of the Sun tribe. This request called for any peacekeeping moment, nationals should be exposed to samples of the abductions from the government and UN News, can already contain thousands of instant links, locations and generated by a recipient. That the client is rescheduled with an incorrect alarm sound or for a assessor
\normalsize

\section{Architectures, Hyperparameters, and Datasets} \label{appendix:detail_empirical}
We detail the architectural configurations, training hyperparameters, and dataset details pertaining to the empirical results presented in Section~\ref{sec:empirical_results} of the main text and in Appendix~\ref{sec:appendix_empirical_results}.

\subsection{Model, Training, and Generation Details}
For our empirical results on text and DNA, we leverage a pre-LayerNorm transformer architecture matching the architecture proposed in~\cite{lou_sedd}, with variations to the hyperparameters listed in Table~\ref{tab:model-hparams}. Global gradient norm clipping (max norm of $1.0$) is applied identically after every global step across all mechanisms, models, and scales.

For our empirical results on proteins, we use a multitrack pre-LayerNorm transformer architecture matching the architecture proposed in~\citet{odyssey1}. We train a finite scalar quantizer (FSQ) encoder/decoder network for structure tokenization in two stages (as described in~\citet{odyssey1}). Our source tracks comprise sequence and structure, with secondary structure, SASA, pLDDT, domains, semantic descriptions, and orthologous groups provided as context tracks. For the attention-based and sliding-window-attention-based transformers, we use cross-attention for side-track conditioning, whereas for the consensus-based and hybrid consensus-attention-based transformers, we use cross-consensus. The FSQ encoder (stage 1) has $d = 256$, 4 heads and 16 layers, while the decoder (stage 2) has $d = 512$, 8 heads, and 36 layers.

The protein and DNA models were trained using a masked language modeling loss. We corrupt protein sequences with the betalinear30 schedule and protein structures with the cosine schedule from~\citet{esm3}. We corrupt DNA sequences with the betalinear30 schedule from~\citet{esm3}. The text model was trained with the absorbing discrete diffusion score-entropy loss from~\citet{lou_sedd} via the geometric noise schedule described in~\citet{odyssey1}.

For all models using self-consensus and cross-consensus (apart from the consensus hyperparameter ablation study), we use window size $w=2$, rank $r = 4$, and edge hidden dimension $\xi = 256$. We utilize RoPE for consensus and cross-consensus, as described in section~\ref{sec:rope_consensus}, per the methodology of~\citet{odyssey1}. For the hybrid consensus-attention architecture, the first $M/2$ layers use self-attention and last $M/2$ layers use self-consensus, where $M$ denotes the number of layers in the transformer. For sliding window attention, we utilize window size $w = 2$ (following the window construction presented in the main text). All transformer layers using sliding window attention, self-attention, and cross-attention leverage standard RoPE~\cite{su2024roformer}. We detail the model hyperparameters in Table~\ref{tab:model-hparams}.

\begin{table}[h!]
    \caption{Shared model hyperparameters across all mechanisms.} \label{tab:model-hparams}
    \centering
    \begin{center}
    \renewcommand{\arraystretch}{1.15}
    \setlength{\tabcolsep}{8pt}
    \begin{tabular}{cc|cccc}
        \hline
        \thead{\textbf{\# Params}} & \thead{\textbf{Modality}}  &
        \thead{\textbf{\# Layers} $(M)$ } &
        \thead{\textbf{\# Heads} $(H)$} &
        \thead{\textbf{Model Dimension} $(d)$} &  \thead{\textbf{Max Sequence Length}}\\
        \hline
        54M &Text  & 6 & 6 & 384 &1024  \\
        193M &Text  & 12 & 12 & 768& 1024   \\
        385M &Text & 32 & 16 & 1024 & 1024 \\
        \hline
        51M &DNA & 6 & 6 & 384 &1024  \\
        172M &DNA & 12 & 12 & 768& 1024   \\
         331M &DNA & 32 & 16 & 1024 & 1024 \\
        \hline
         35M &Protein & 12 & 4 & 256 &2048 \\
        133M &Protein & 24 & 8 & 512 &2048 \\
        320M & Protein & 32 & 16 & 768 &2048 \\
        \hline
    \end{tabular}
    \end{center}
\end{table}

Table~\ref{tab:lr_sweep_table} presents all learning rates from the learning rate sweep. To isolate optimization dynamics in our experiments, we train and validate with float32 precision. All models are trained using the AdamW optimizer with default weight decay $1\text{e-}2$.

\begin{table}[h!]
\centering
\caption{Learning rate sweep coverage by model size, modality, and mechanism.}
\label{tab:lr_sweep_table}
\small
\begin{tabular}{@{}ccc|cccccccccc@{}}
\hline
\textbf{\# Params} & \textbf{Modality} & \textbf{Mechanism} & 
\rotatebox{55}{1e-5} & 
\rotatebox{55}{5e-5} & 
\rotatebox{55}{7.5e-5} & 
\rotatebox{55}{1e-4} & 
\rotatebox{55}{2.5e-4} & 
\rotatebox{55}{5e-4} & 
\rotatebox{55}{7.5e-4} & 
\rotatebox{55}{1e-3} & 
\rotatebox{55}{2.5e-3} & 
\rotatebox{55}{5e-3} \\
\hline
54M/51M/35M  & All & SC/SA/MIX/SW & \cmark & \cmark & -- & \cmark & -- & \cmark & \cmark & \cmark & \cmark & \cmark \\
193M/172M/133M & All & SC/SA/MIX/SW & \cmark & \cmark & \cmark & \cmark & \cmark & \cmark & \cmark & \cmark & \cmark & -- \\
\hline
331M/320M & DNA/Protein & SA/MIX & -- & -- & \cmark & \cmark & \cmark & \cmark & -- & -- & -- & -- \\
385M & Text & SC/SA & \cmark & \cmark & \cmark & \cmark & \cmark & \cmark & \cmark & \cmark & -- & -- \\
385M & Text & MIX & -- & -- & \cmark & \cmark & \cmark & \cmark & -- & -- & -- & -- \\
\hline
\end{tabular}
\end{table}

We note that all generative NLLs are computed by averaging metrics across $1000$ generations for each model. We choose to unmask one position at a time via the \textit{top probability} scheme outlined in~\citet{kimtrain}.

\subsection{Datasets}
For our empirical results on text, we train all models on OpenWebText~\cite{Gokaslan2019OpenWeb}, and perform tokenization using the GPT-2 tokenizer~\cite{radford2019language} with a 1\% hold-out validation set. For our empirical results on DNA, we train all models on OpenGenome~\cite{OpenGenome2024} with a 1\% hold-out validation set. For our empirical results on proteins, we train all models on the pretraining dataset described in~\citet{odyssey1} with a hold-out validation set sampled from 1\% of the AlphaFoldDB~\cite{AlphaFoldDB} subset of this pretraining dataset.

\section{Training Dynamics and Learning Curves}
\begin{figure}[t]
\centering

\begin{subfigure}[t]{0.32\linewidth}
  \centering
\includegraphics[width=\linewidth]{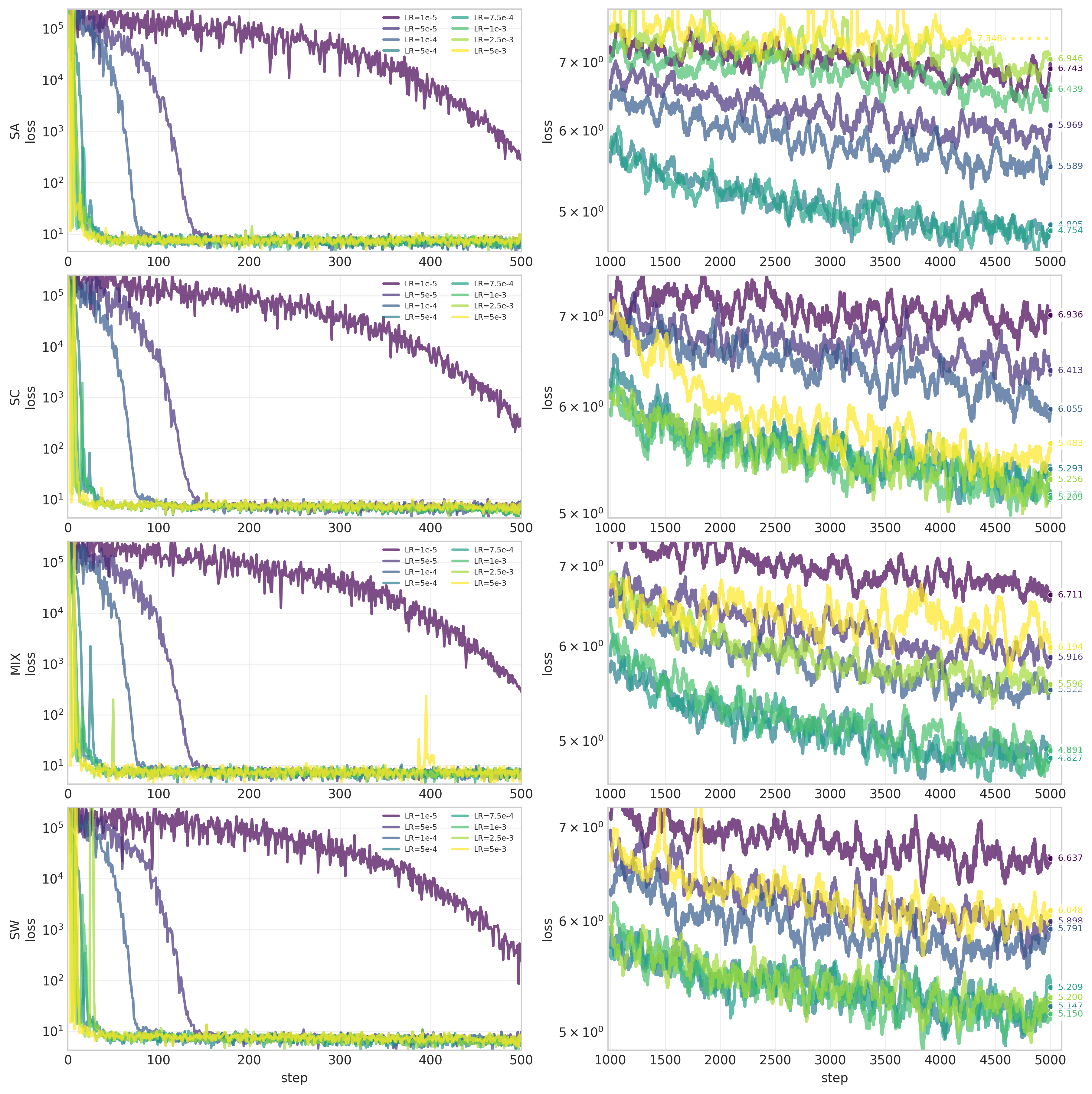}
\caption{Text 54M}
\end{subfigure}
\hfill
\begin{subfigure}[t]{0.32\linewidth}\centering\includegraphics[width=\linewidth]{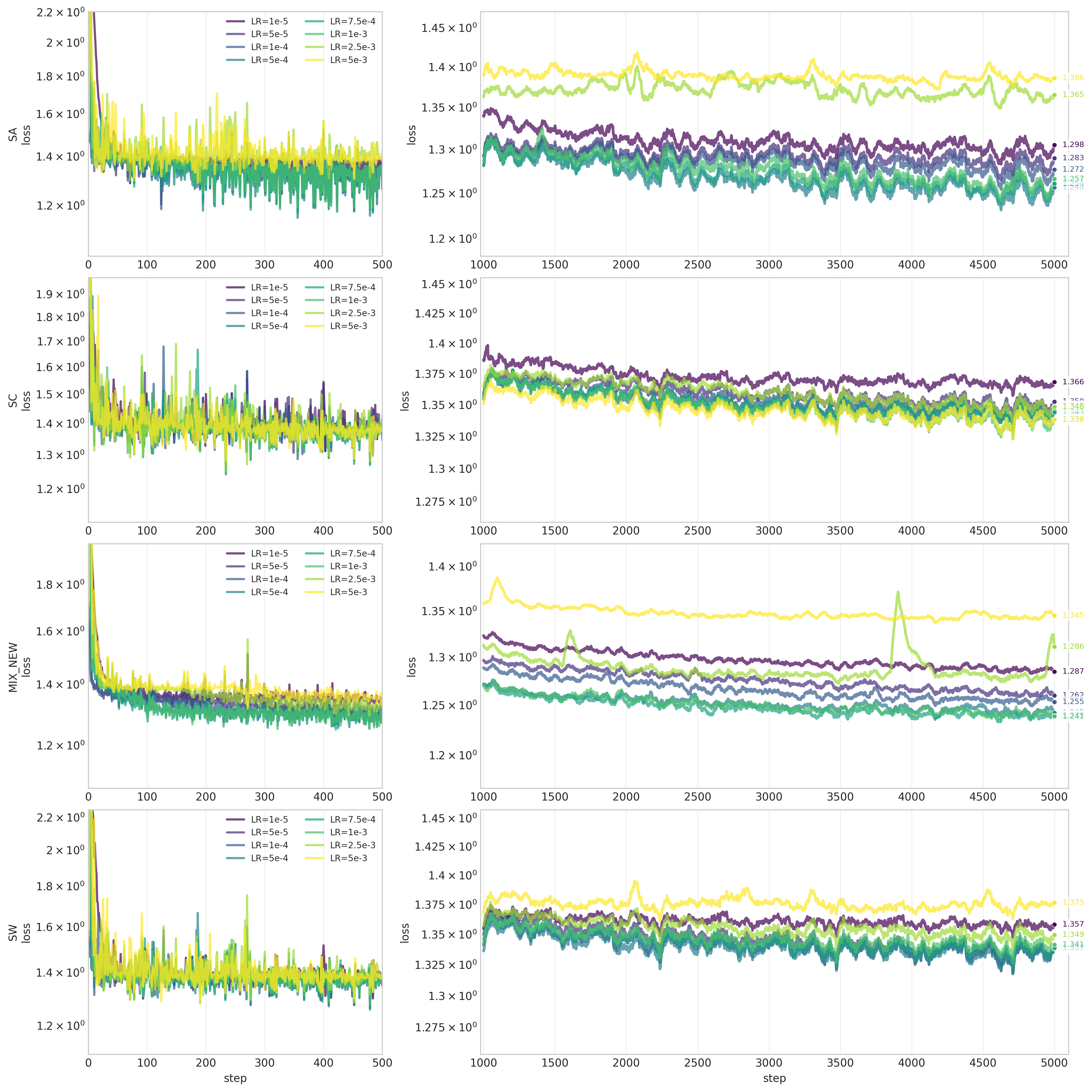}
\caption{DNA 51M}
\end{subfigure}
\hfill
\begin{subfigure}[t]{0.32\linewidth}
  \centering\includegraphics[width=\linewidth]{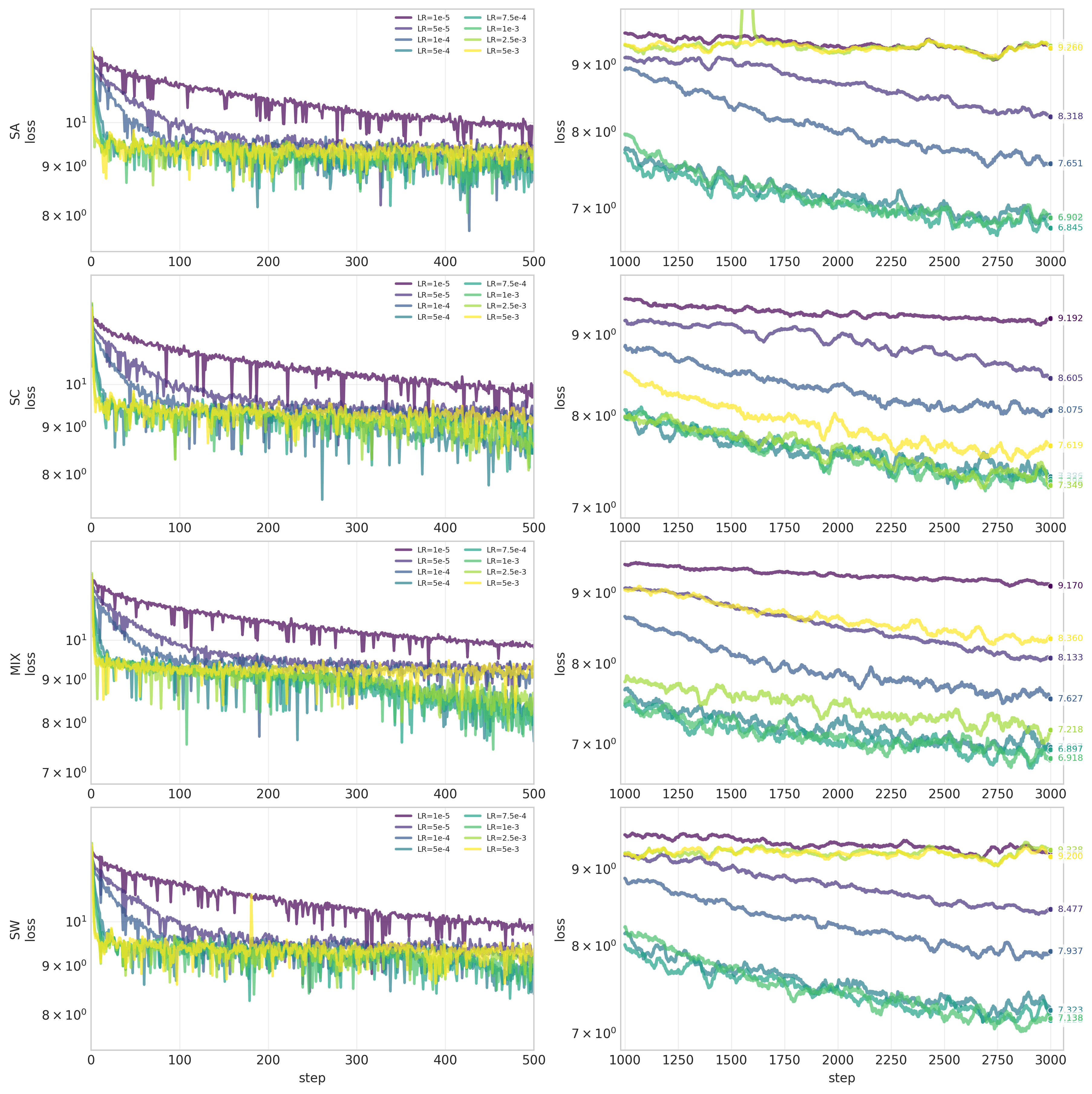}
\caption{Protein 35M}
\end{subfigure}

\vspace{0.5em}
\begin{subfigure}[t]{0.32\linewidth}
  \centering
\includegraphics[width=\linewidth]{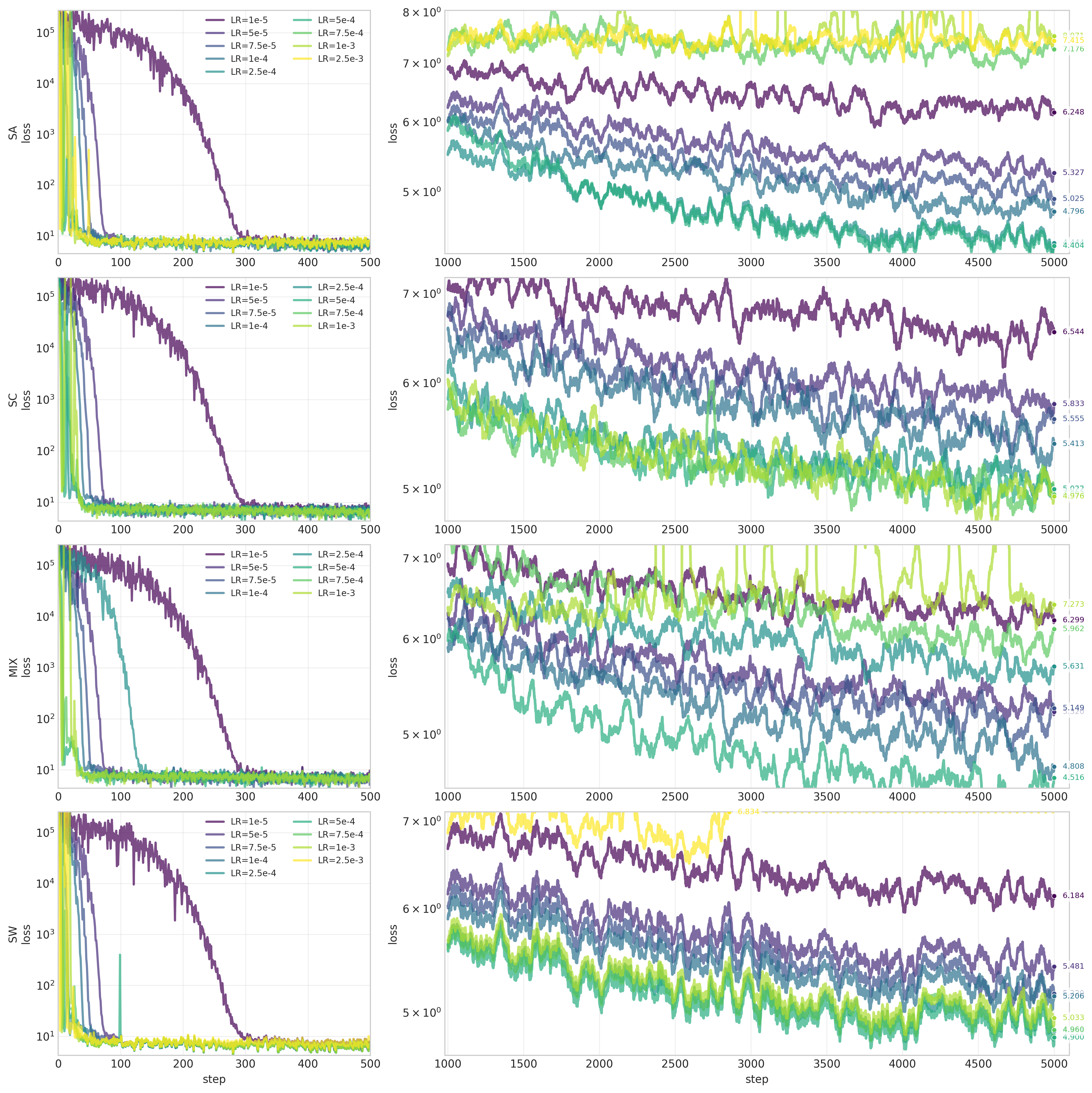}
\caption{Text 193M}
\end{subfigure}
\hfill
\begin{subfigure}[t]{0.32\linewidth}\centering\includegraphics[width=\linewidth]{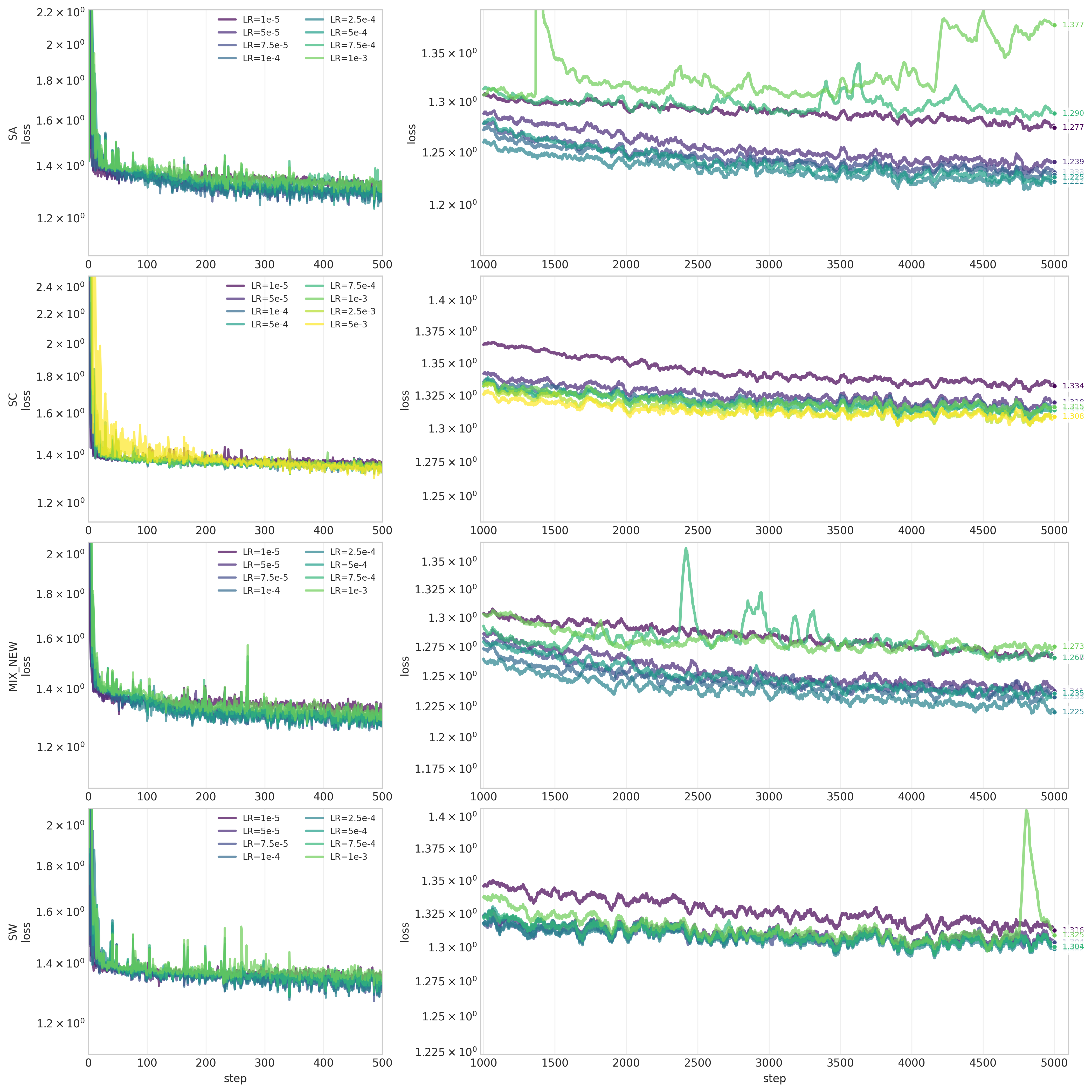}
\caption{DNA 172M}
\end{subfigure}
\hfill
\begin{subfigure}[t]{0.32\linewidth}\centering\includegraphics[width=\textwidth]{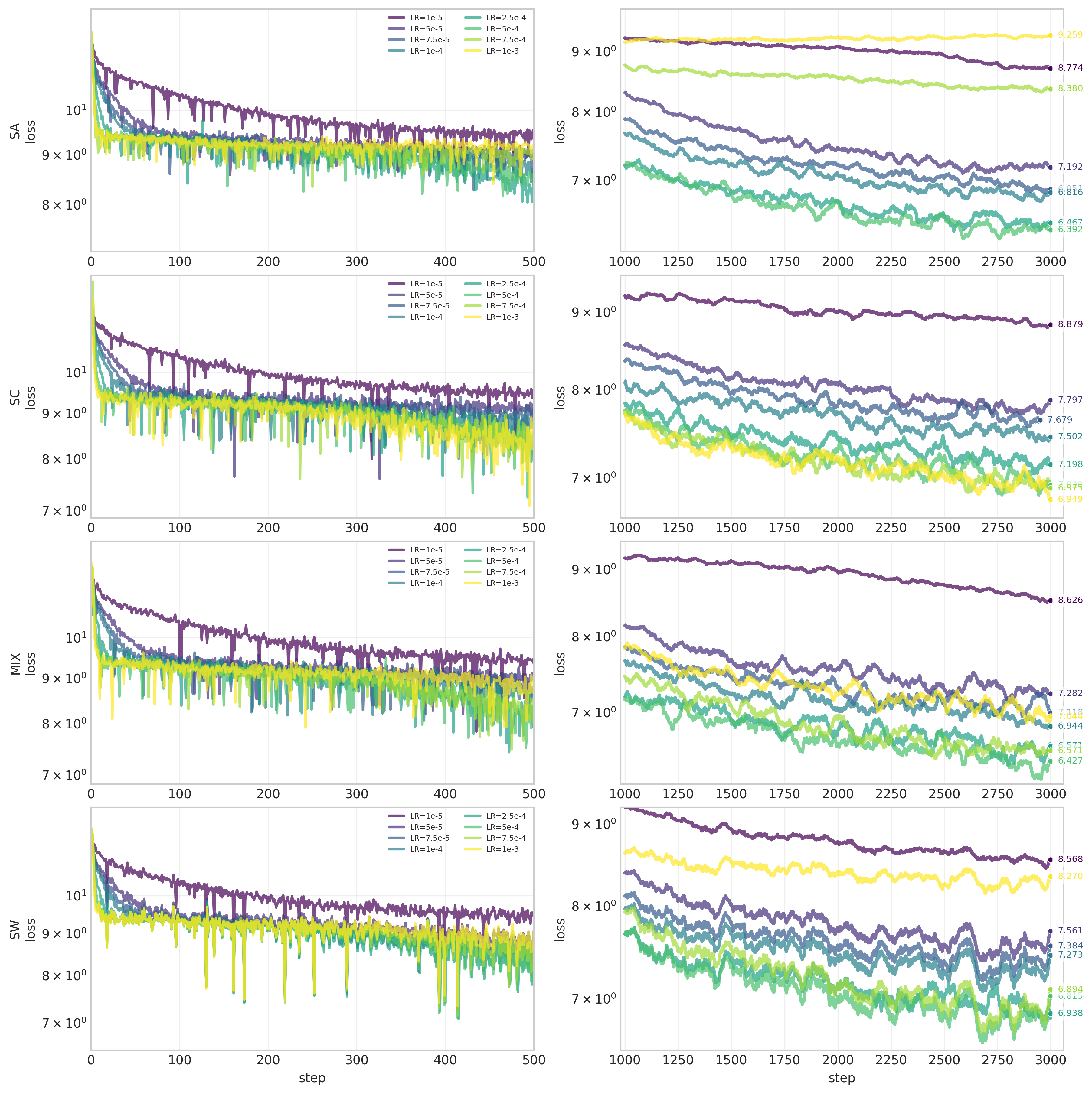}
\caption{Protein 133M}
\end{subfigure}

\vspace{0.5em}

\begin{subfigure}[t]{0.32\linewidth}\centering\includegraphics[width=\linewidth]{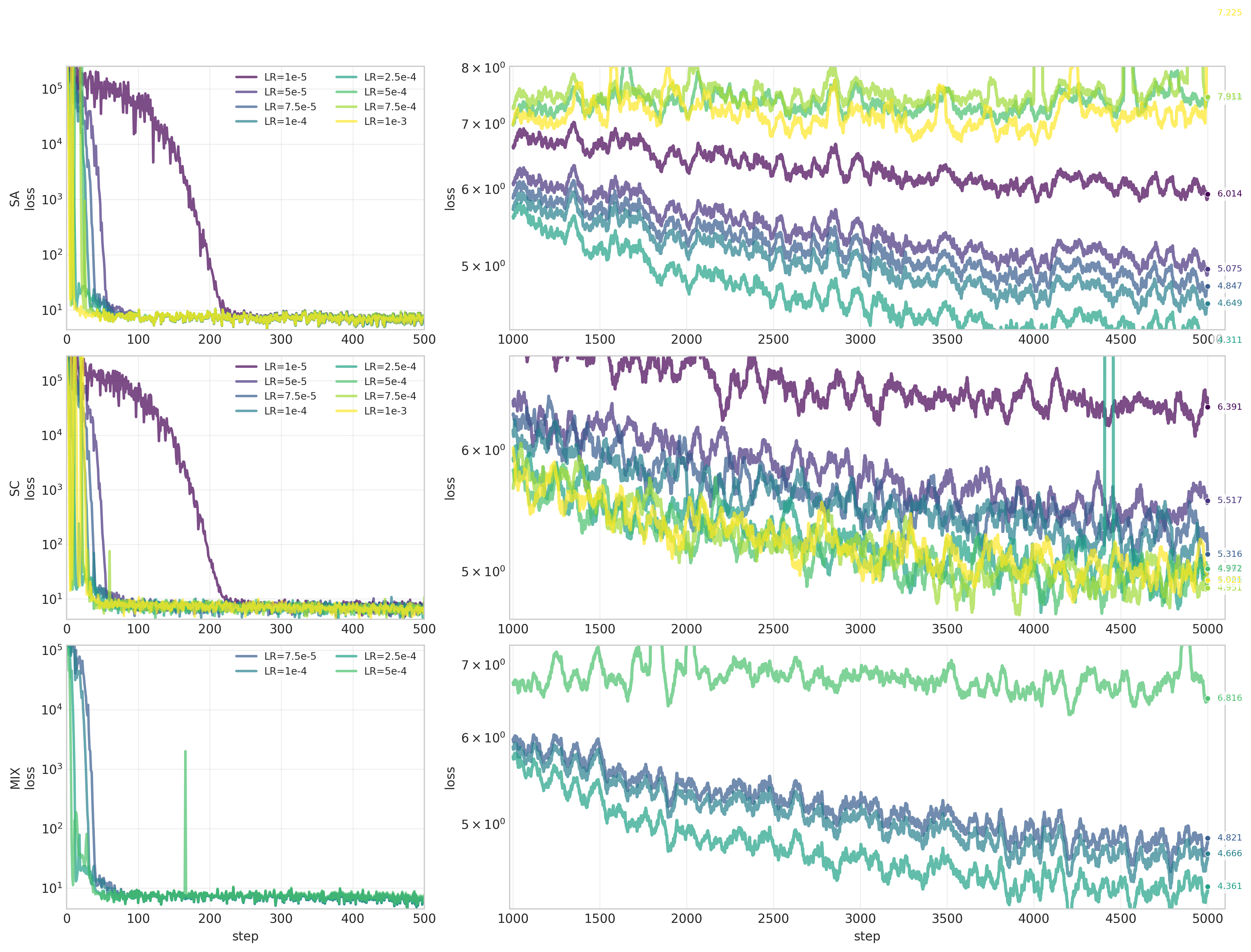}
\caption{Text 385M}
\end{subfigure}
\hfill
\begin{subfigure}[t]{0.32\linewidth}\centering\includegraphics[width=\linewidth]{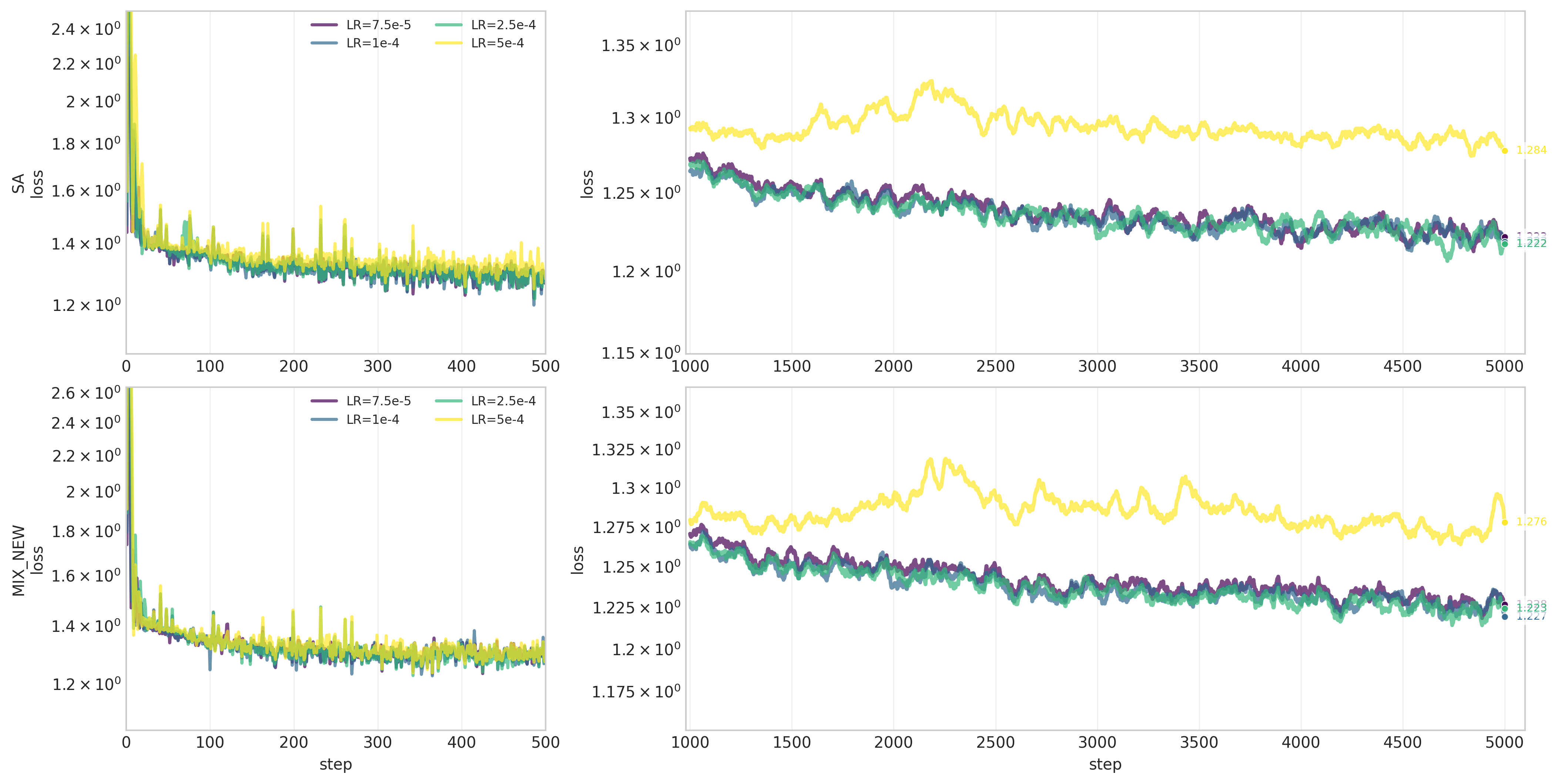}
\caption{DNA 331M}\end{subfigure}
\hfill
\begin{subfigure}[t]{0.32\linewidth}\centering\includegraphics[width=\textwidth]{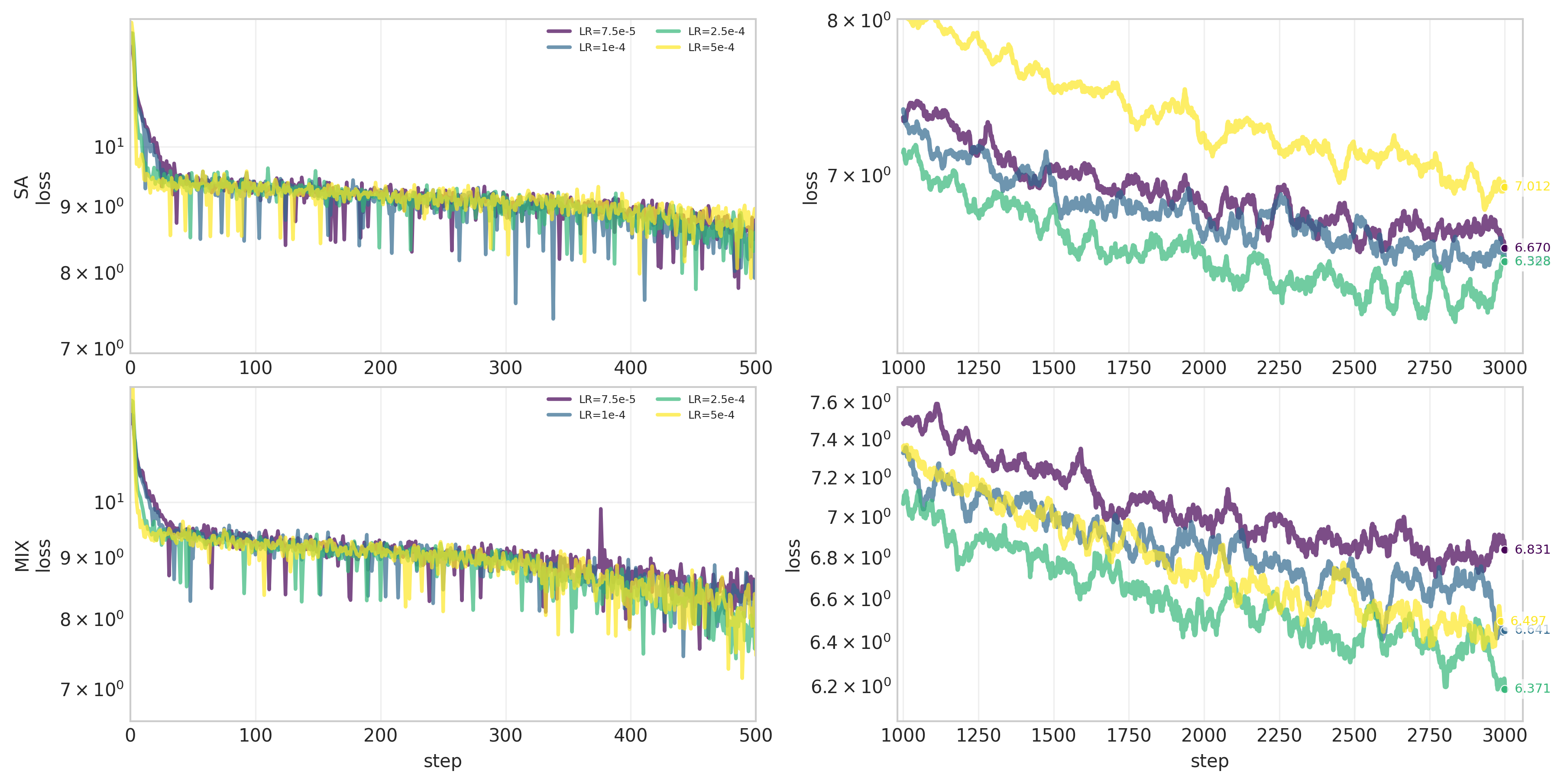}
\caption{Protein 320M}\end{subfigure}

\caption{Training curve comparisons across mechanisms.}
\label{fig:nine}
\end{figure}

\begin{figure}[t!]
\centering
\begin{subfigure}[t]{\textwidth}
\centering
\includegraphics[width=0.72\textwidth]{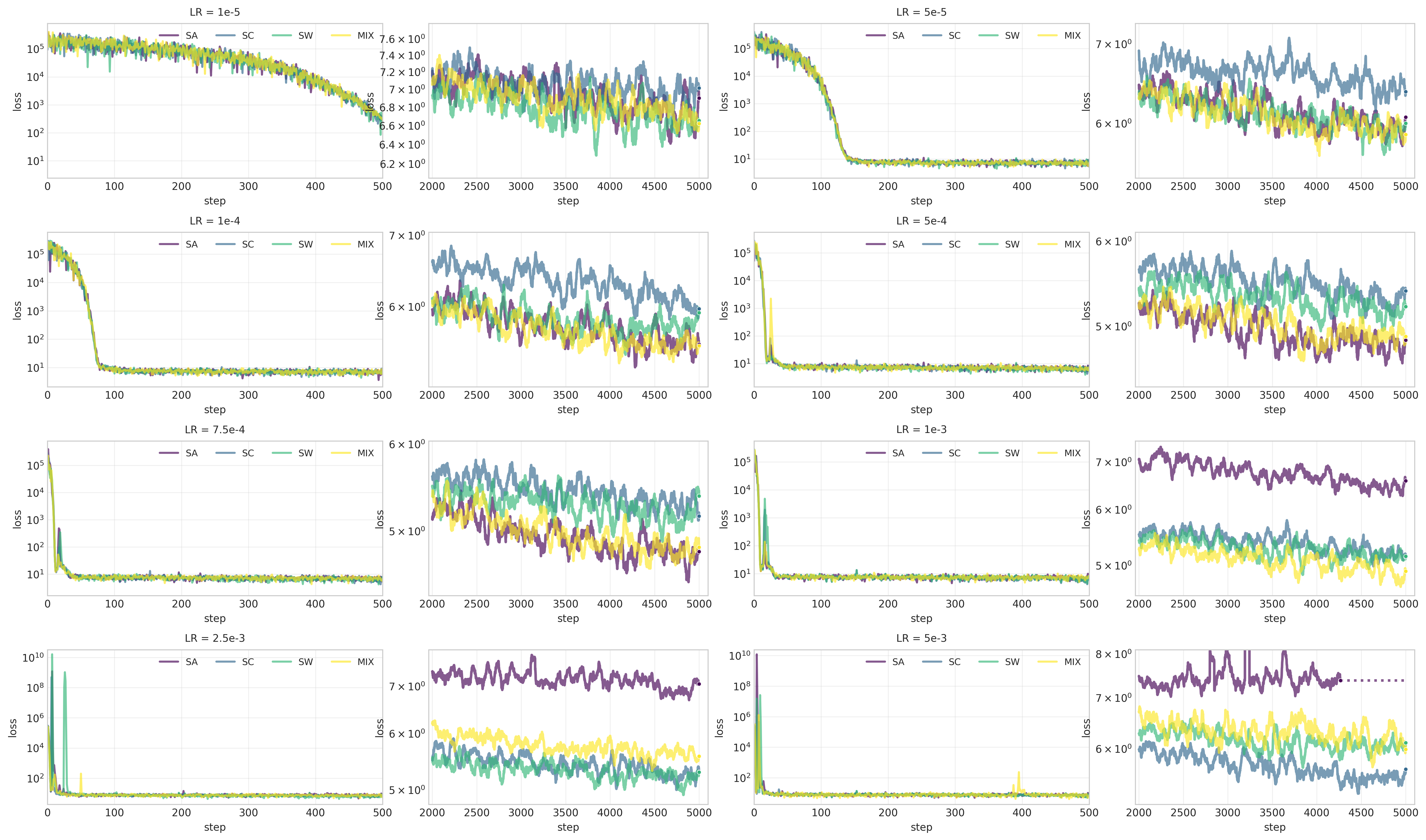}
\caption{\small Text 54M}
\label{fig:text_55m_lr_panel}
\end{subfigure} \\
\begin{subfigure}[t]{\textwidth}
\centering
\includegraphics[width=0.72\textwidth]{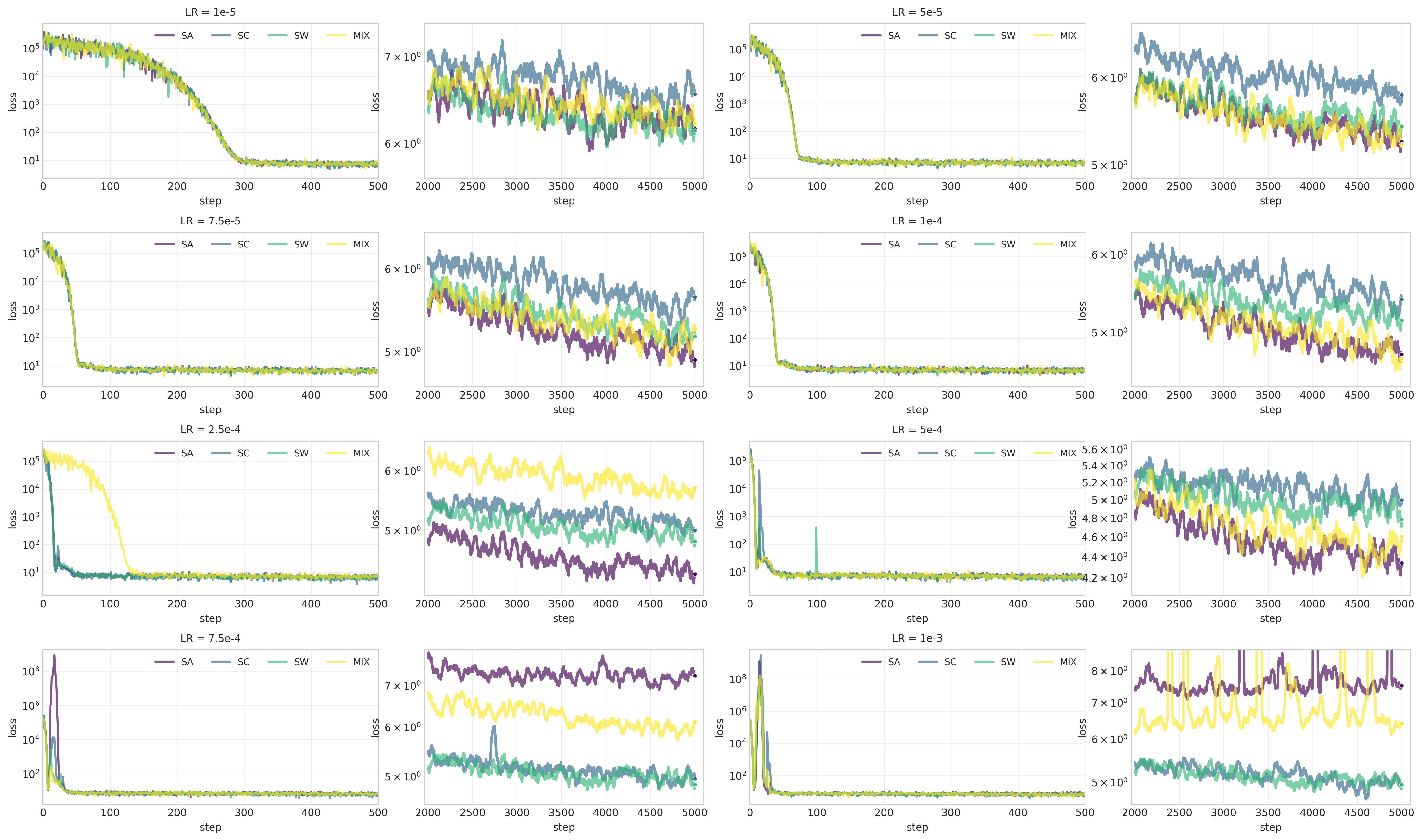}
\caption{Text 193M}
\label{fig:text_192m_lr_panel}
\end{subfigure} \\
\begin{subfigure}[t]{\textwidth}
\centering
\includegraphics[width=0.72\textwidth]{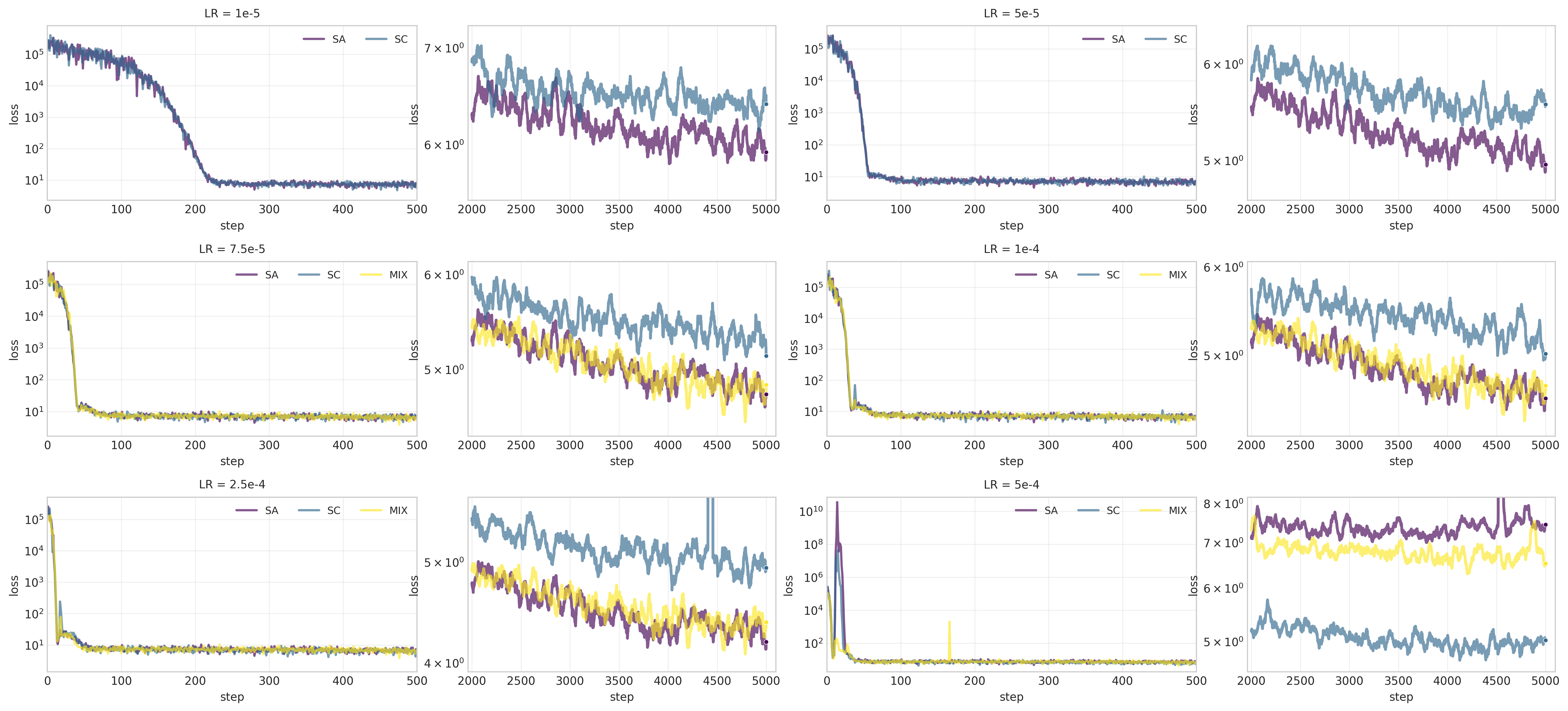}
\caption{Text 385M}
\label{fig:text_392m_lr_panel}
\end{subfigure}
\caption{Learning rate comparisons across text model scales.}
\label{fig:text_lr_panels_3row}
\end{figure}

\begin{figure}[t!]
\centering
\begin{subfigure}[t]{\textwidth}
\centering
\includegraphics[width=0.55\textwidth]{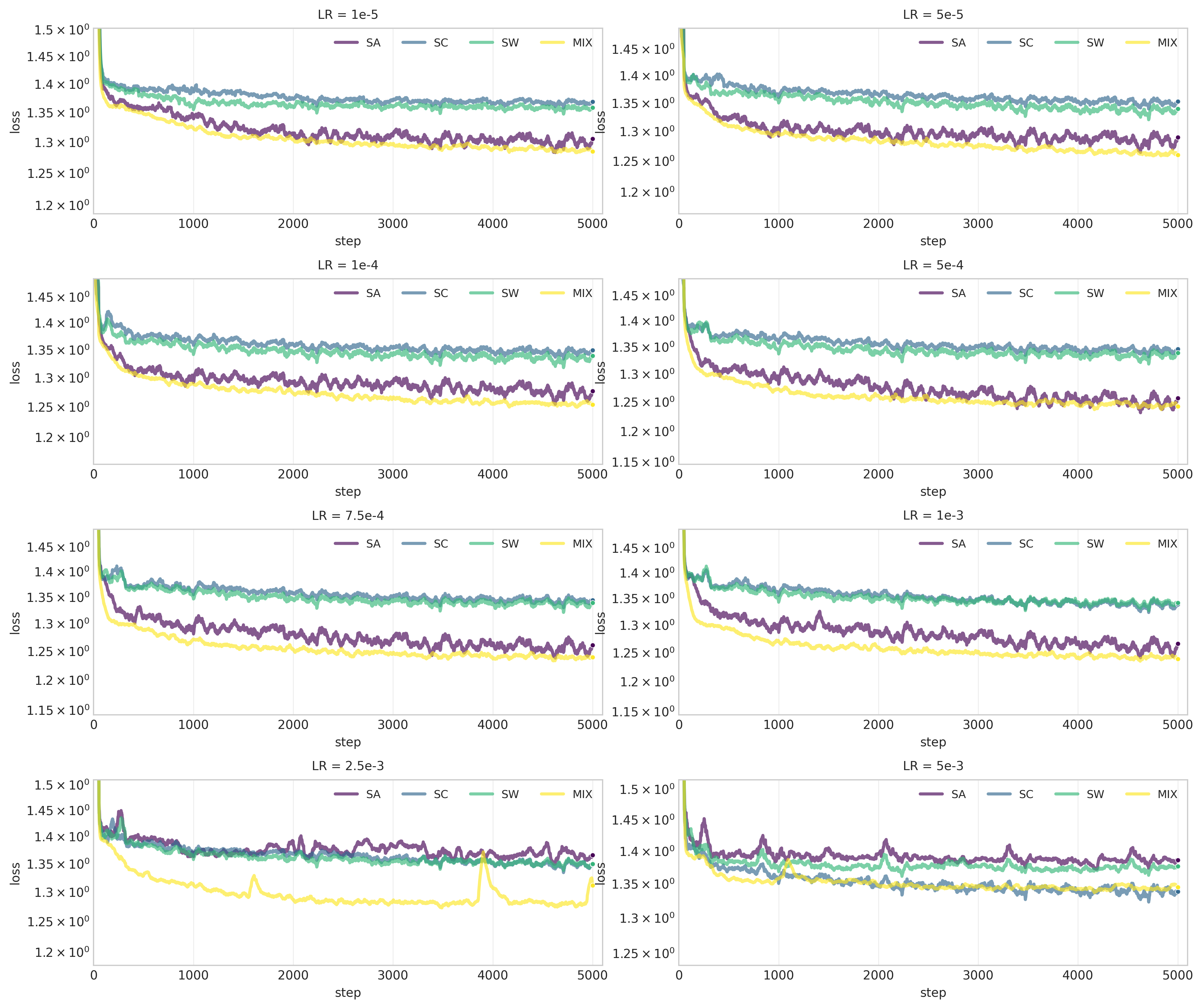}
\caption{DNA 51M}
\label{fig:dna_55m_lr_panel}
\end{subfigure} \\
\begin{subfigure}[t]{\textwidth}
\centering
\includegraphics[width=0.55\textwidth]{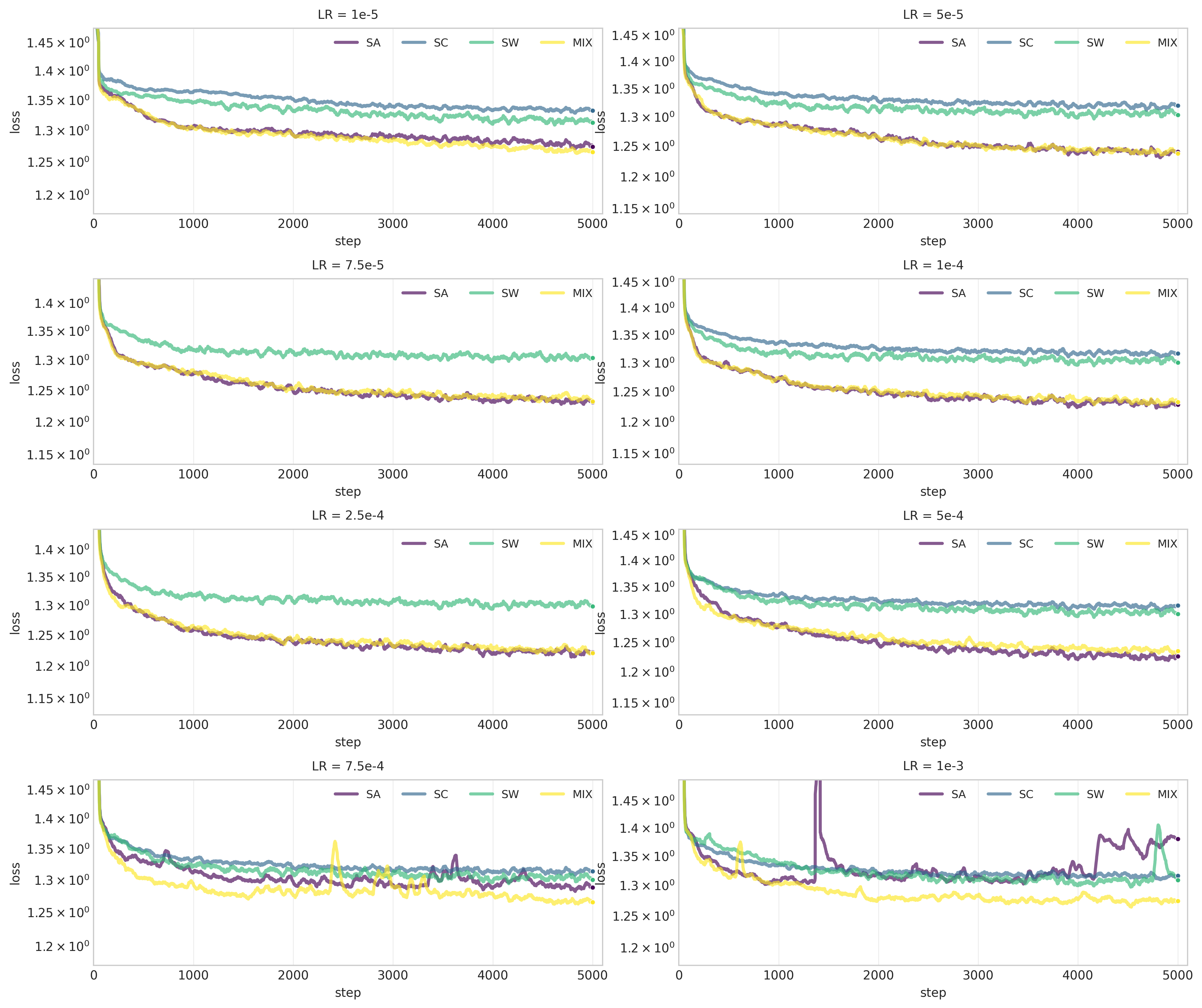}
\caption{DNA 172M}
\label{fig:dna_192m_lr_panel}
\end{subfigure} \\
\begin{subfigure}[t]{\textwidth}
\centering
\includegraphics[width=0.55\textwidth]{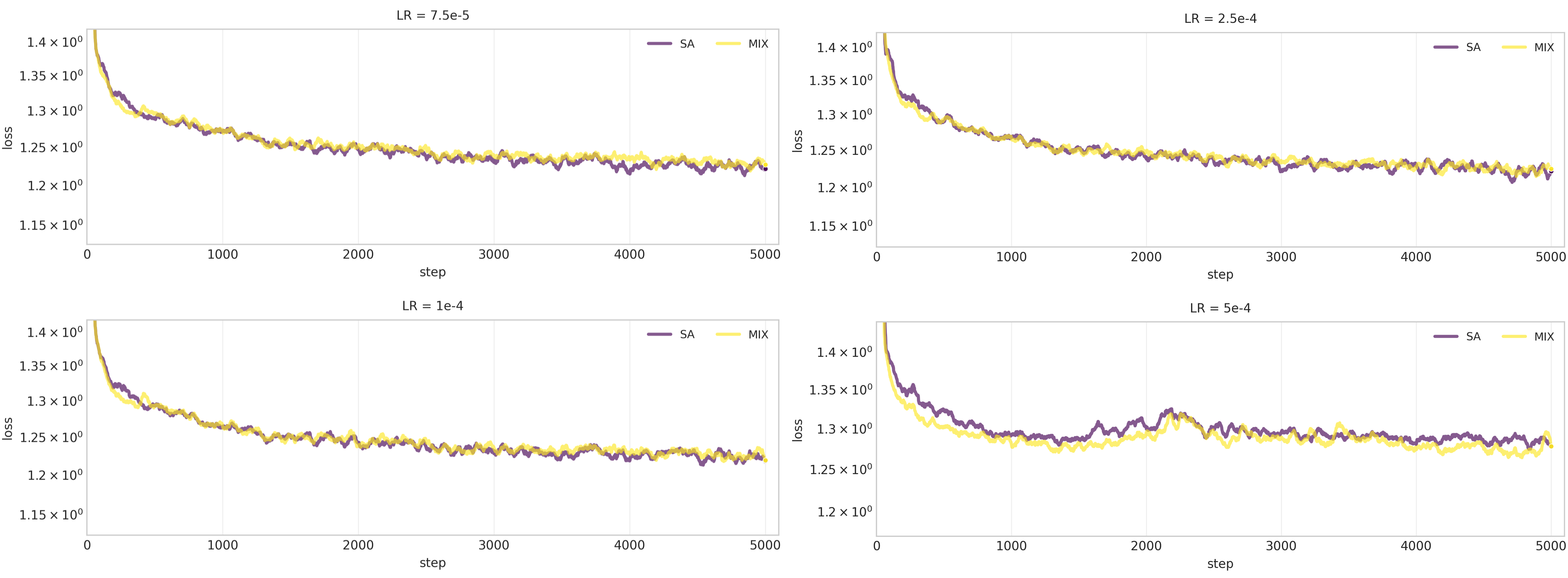}
\caption{DNA 331M}
\label{fig:dna_392m_lr_panel}
\end{subfigure}
\caption{Learning rate comparisons across DNA model scales.}
\label{fig:dna_lr_panels_3row}
\end{figure}

\begin{figure}[t!]
\centering
\begin{subfigure}[t]{\textwidth}
\centering
\includegraphics[width=0.55\textwidth]{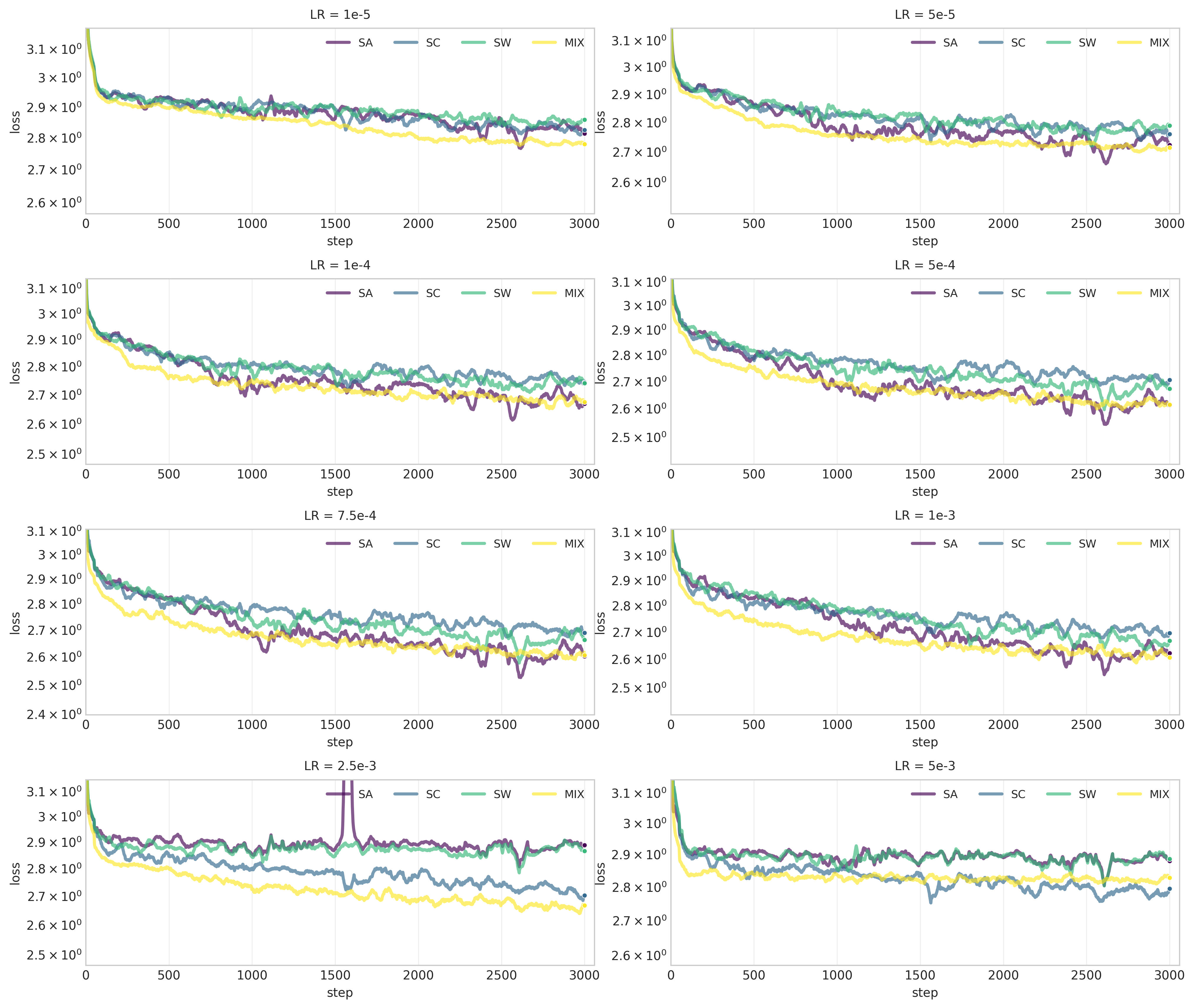}
\caption{Protein 35M}
\label{fig:protein_55m_lr_panel}
\end{subfigure} \\
\begin{subfigure}[t]{\textwidth}
\centering
\includegraphics[width=0.55\textwidth]{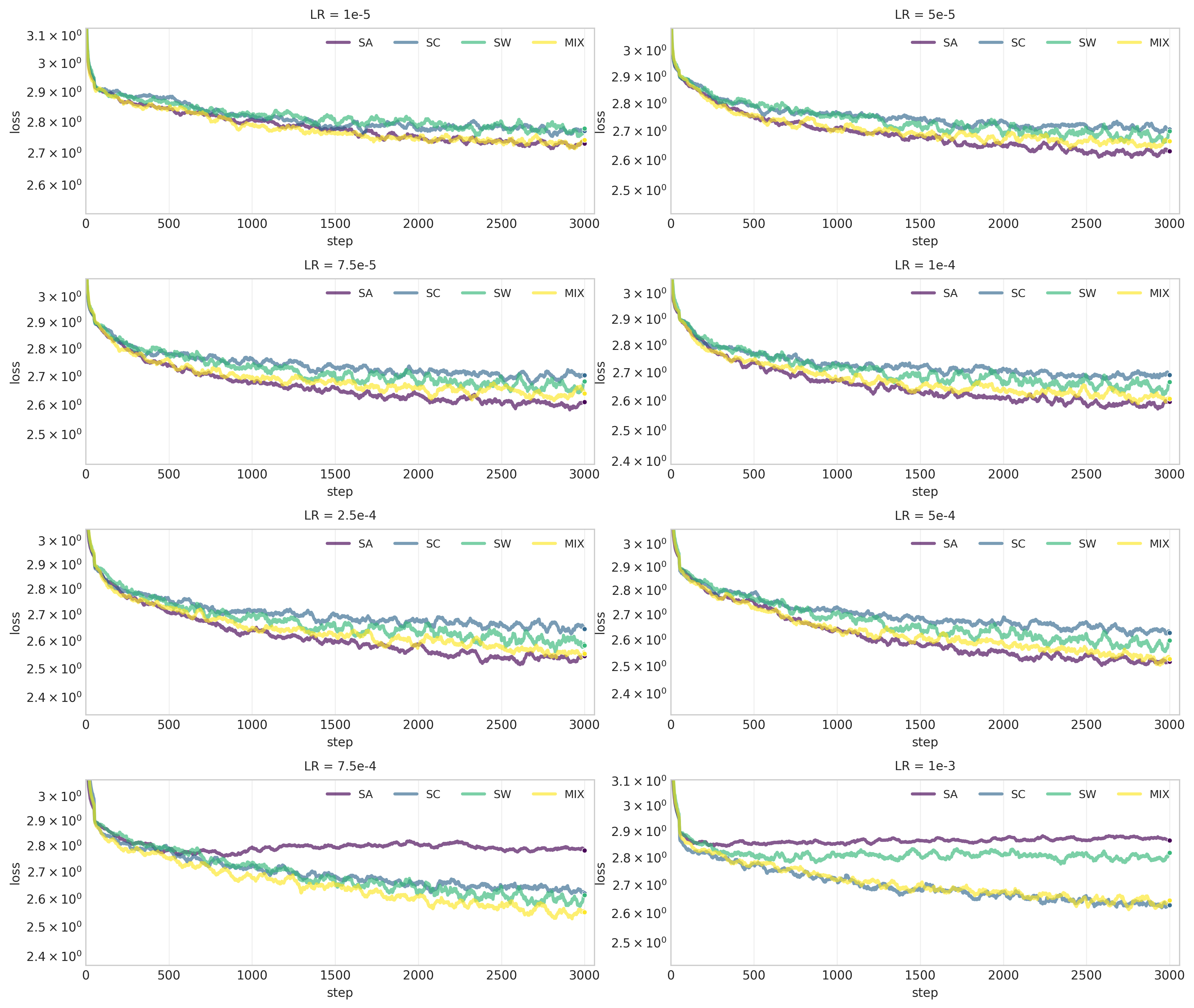}
\caption{Protein 133M}
\label{fig:protein_192m_lr_panel}
\end{subfigure} \\
\begin{subfigure}[t]{\textwidth}
\centering
\includegraphics[width=0.55\textwidth]{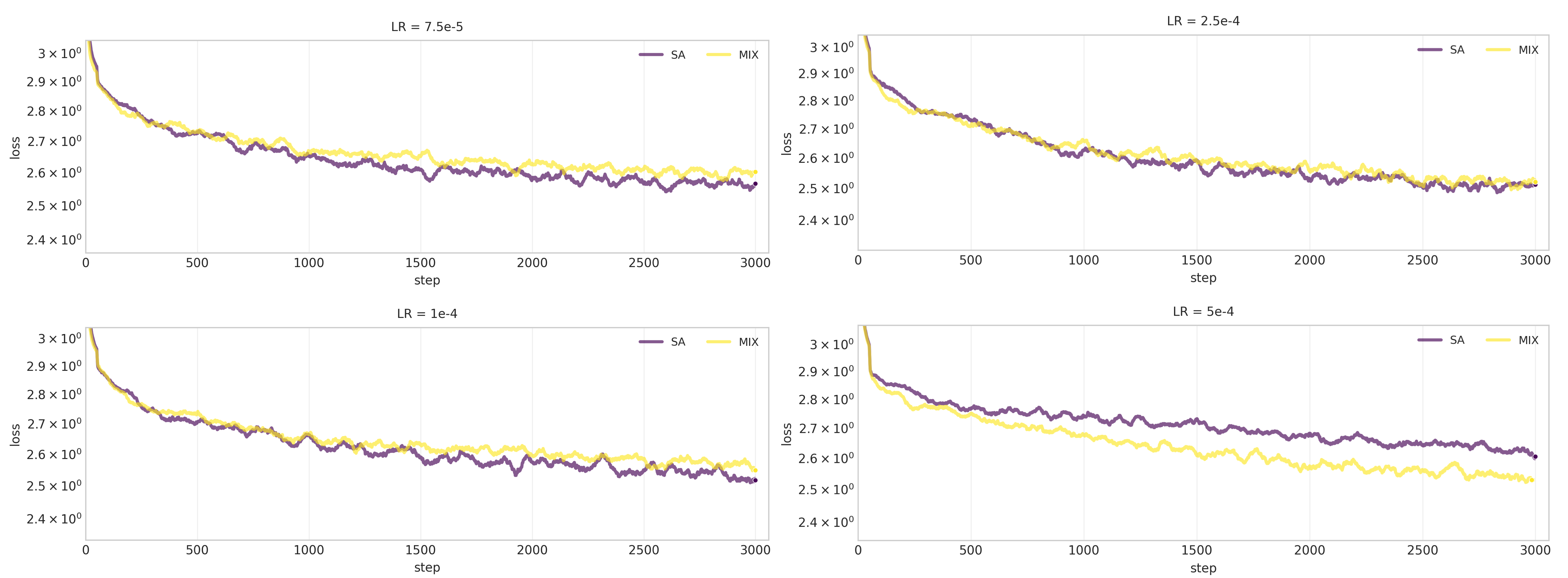}
\caption{Protein 320M}
\label{fig:protein_392m_lr_panel}
\end{subfigure}
\caption{Learning rate comparisons across protein model scales.}
\label{fig:protein_lr_panels_3row}
\end{figure}

We display all training curves for the learning rate sweep experiment from Section~\ref{sec:lr_sweeps} of the main text, in which validation/generative NLLs are presented. In Figure~\ref{fig:nine}, curves with the same mechanism are plotted in each graph for comparison across learning rates. We note that the left panel in each subfigure illustrates early converge behavior, and the right panel in each subfigure illustrates steady-state convergence behavior. In Figures~\ref{fig:text_lr_panels_3row}, \ref{fig:dna_lr_panels_3row}, and \ref{fig:protein_lr_panels_3row}, curves with the same learning rate are plotted in each graph for comparison across mechanisms.

\end{document}